\documentclass[10pt,journal,compsoc]{IEEEtran}

\usepackage[nocompress]{cite}
\usepackage{array}
\usepackage{times}
\usepackage{epsfig}
\usepackage{graphicx}
\usepackage{amsmath}
\usepackage{amssymb}
\usepackage{caption}
\usepackage{subcaption}
\usepackage{tabularx}
\usepackage{threeparttable}
\usepackage{multirow}
\usepackage{comment}
\usepackage{makecell}
\usepackage{enumitem}
\usepackage{enumerate}
\usepackage{MnSymbol}
\usepackage{lscape}
\usepackage{rotating}
\usepackage{algorithm}
\usepackage{algorithmic}

\usepackage[pagebackref=false,breaklinks=true,colorlinks,bookmarks=false]{hyperref}
\usepackage{tikz}
\usepackage{siunitx}
\usepackage{stfloats}
\usepackage{url}
\usepackage[usestackEOL]{stackengine}
\usepackage{makebox}
\begin{document}

\title{Empirical Assessment of End-to-End Iris Recognition System Capacity}

\author{Priyanka~Das,~\IEEEmembership{Student Member~IEEE,}
        Richard~Plesh,~\IEEEmembership{Student Member~IEEE,}
        Veeru~Talreja,~\IEEEmembership{Member~IEEE,}
        Natalia~Schmid,~\IEEEmembership{Member~IEEE,} 
        Matthew~Valenti,~\IEEEmembership{Member~IEEE Fellow,} 
        Joseph~Skufca, 
        and~Stephanie~Schuckers,~\IEEEmembership{Senior Member~IEEE}%
        
\IEEEcompsocitemizethanks{\IEEEcompsocthanksitem P. Das, R. Plesh, and S. Schuckers are with the Department
of Electrical and Computer Engineering, Clarkson University, Potsdam,
NY, 13676.\protect\\
E-mail: prdas@clarkson.edu
\IEEEcompsocthanksitem J. Skufca is with the Department of Mathematics, Clakson University, Potsdam, NY, 13676.
\IEEEcompsocthanksitem V. Talreja, N. Schmid and M. Valenti are with Computer Science and Electrical Engineering, West Virginia University, Morgantown, WV, 26505.}%
}

\markboth{IEEE Transactions on Biometrics, Behavior, and Identity Science}%
{Das \MakeLowercase{\textit{et al.}}: Empirical Assessment of End-to-End Iris
Recognition System Capacity }

\IEEEtitleabstractindextext{%
\begin{abstract}
Iris is an established modality in biometric recognition applications including consumer electronics, e-commerce, border security, forensics, and de-duplication of identity at a national scale. In light of the expanding usage of biometric recognition, identity clash (when templates from two different people match) is an imperative factor of consideration for a system's deployment. This study explores system capacity estimation by empirically estimating the constrained capacity of an end-to-end iris recognition system (NIR systems with Daugman-based  feature extraction) operating at an acceptable error rate i.e. the number of subjects a system can resolve before encountering an error. We study the impact of six system parameters on an iris recognition system's constrained capacity- number of enrolled identities, image quality, template dimension, random feature elimination, filter resolution, and system operating point. In our assessment, we analyzed 13.2 million comparisons from 5158 unique identities for each of 24 different system configurations. This work provides a framework to better understand iris recognition system capacity as a function of biometric system configurations beyond the operating point, for large-scale applications. 
\end{abstract}

}

\maketitle

\begin{figure}[h]
\centering
    \includegraphics[width=3in,keepaspectratio]{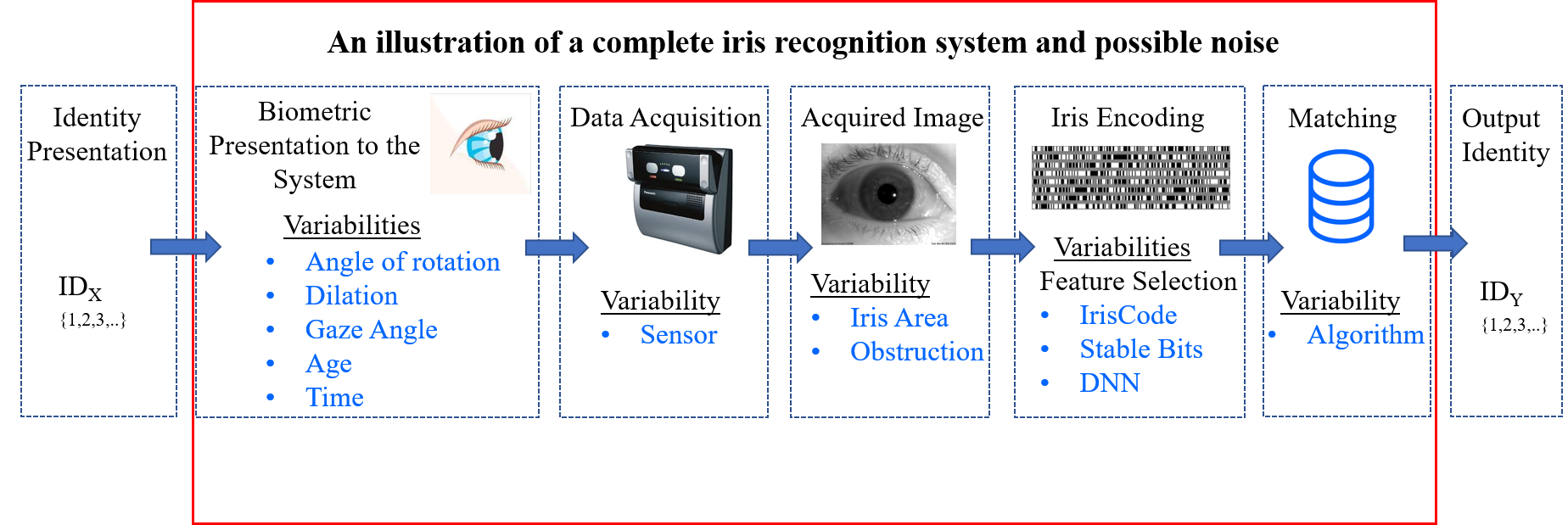}
    \caption{\footnotesize Overview of an end-to-end iris recognition system where a variety of factors induce variability and impact capacity at different stages}
    \label{fig:IrisRecognitionChannel}
 \end{figure}

\IEEEraisesectionheading{\section{Introduction}\label{sec:introduction}}
\IEEEPARstart{B}{iometric}   recognition technology is being used in widespread applications for verification and identification in both commercial and government platforms. With the widening horizon of applications, the technology is growing in its implementation from small cohorts (e.g. access control)  to  large scale ( e.g. criminal identification \cite{FBI}) to the national level (e.g. de-duplication of identity \cite{Aadhaar}). The capacity of a biometric system i.e. \textbf{the number of identities the biometric system can accommodate before it encounters an identity clash} \cite{gong2017capacity}, is a quintessential factor especially in large-scale or national level applications that deal with 1:N or N:N matching. Ideally, biometric characteristics captured from different identities should have separable features due to their inherent property of uniqueness. Uniqueness is an indispensable property of biometrics that allows biometrics to define identity. Biometric uniqueness is characterized as \textit{no two people should have the same identifier} \cite{clarke1994human}.  In automated biometric recognition systems, we encounter cases of false accept errors where two biometric samples from different identities match. Insufficiently, the distinctiveness between identities is sometimes explained in terms of error rates. This puts some doubt on the uniqueness of biometrics. However, a case of false accept does not necessarily mean the biometric characteristics of the two different individuals are the same or similar. The decision is impacted by multiple factors in a biometric recognition channel- noise induced by how a biometric is presented to the system, variability in sensors in terms of camera pixel distortions, variability in the enrollment and the probe sample captured at different time points, the method for feature extraction, the choice of features being used for matching, and the matching algorithm used. An illustration of an iris recognition pipeline and factors that might impact a decision is shown in Figure~\ref{fig:IrisRecognitionChannel}.

The inherent random variation and complexity in the iris pattern are considered the basis for uniqueness in the iris. In an automated operational iris recognition scenario where false accept is a reality, we cannot logically argue based on only the ``inherent'' unique characteristics. Biometric Doddington zoo menagerie \cite{doddington1998sheep} \cite{yager2008biometric} \cite{paone2011consistency} was introduced to explain non-uniformity in biometric recognition performance. Among many classes, ``wolves'' have been denoted for individuals who are capable of easily impersonating other classes leading to false accepts, and ``lambs" has been denoted for individuals that are easy to imitate, also contributing to false accepts. However, this explainable concept does not quantify the uniqueness of an operating system. Attempts have been made to explain uniqueness in different modalities including iris. Daugman has discussed ``identity collision" in iris in terms of entropy and established the high entropy per bit of feature encoding in the IrisCode as the basis for high resistance to clash in IrisCodes  \cite{daugman2015information}.

 \textit{Given the challenge of establishing uniqueness, are there other approaches which can be used to quantify the ``uniqueness" of an iris recognition system?} In this paper we investigate the uniqueness of an iris recognition system 
 for images captured under NIR illumination following Daugman-based IrisCode feature templates in terms of ``constrained capacity" i.e. \textit{estimating the maximum number of users a system can identify at an acceptable error rate} \cite{gong2017capacity}. We evaluate iris recognition system capacity in terms of parameters of \begin{itemize}[leftmargin=*, noitemsep]
    \item Quality of biometric samples
    \item Filter resolution
    \item Template dimension
    \item Random feature reduction 
    \item Number of identities in the system
    \item System operating point
\end{itemize}

In this assessment of constrained capacity, we take a data-driven approach. We assessed 24 different system configurations, each with approximately 13.2 million comparisons based on N:N matching, from approximately 5.1k unique identities. To analyze the impact of feature dimension, random feature reduction, and filter resolution on capacity, we consider traditional iris features, the IrisCode inspired by Daugman's approach \cite{daugman1993high}, implemented in OSIRIS \cite{othman2016osiris}. It is believed that most commercial deployments of iris recognition systems utilize an IrisCode-type algorithm \cite{daugman2015information}. The main contributions of this paper on iris recognition capacity are that it has:
\begin{itemize}[leftmargin=*, noitemsep]
    \item Empirically established the constrained capacity of NIR illuminated Daugman-based iris recognition system, studying the impact of template dimension, filter resolution, random feature reduction, image quality, and system operating points
    \item Established the relationship between errors encountered by a system in terms of identity clash, the number of identities in a system, and the number of features in an iris template 
\end{itemize}

The rest of the paper is organized as follows- Section~\ref{sec:SOA} summarizes the state-of-the-art research in the scope of identity clash in iris recognition; Section~\ref{sec:Methodology} details our approach towards assessing the constrained capacity of iris recognition system; Section~\ref{sec:Results} reports on our analysis, findings and conclusion and Section~\ref{sec:DC} provides an insightful discussion on our conclusions.

\section{State of the Art}\label{sec:SOA}
 \textit{How many identities can a biometric system resolve?} It is a long persistent query that has been approached by researchers with many different techniques, for different modalities like face \cite{gong2017capacity} \cite{balazia2021unique}  \cite{de2013entropy}, fingerprint \cite{yankov2019fingerprint}, iris \cite{bolle2004iris} \cite{yoon2005individuality} \cite{daugman2006probing}\cite{daugman2015information}, including designing models to estimate capacity \cite{bolle2004iris}, adapting concepts from information theory \cite{adler2006towards}  \cite{schmid2008empirical} \cite{daugman2015information} \cite{de2013entropy}, score-based uniqueness measure \cite{balazia2021unique}, and empirical computation of capacity \cite{daugman2006probing}. 

We found the earliest reference to the study of iris capacity in 2004 under the study of concepts like ``individuality" or ``uniqueness". Bolle et al. \cite{bolle2004iris}, in their modelling approach for iris individuality, refers to the concept as ``given a biometric sample, determine the probability of finding an arbitrary biometric sample from the target population sufficiently similar to it" i.e. the lower bound on the false accepts. The work modelled iris individuality as the probability of False Accept Rate (FAR) and False Reject Rate (FRR) in terms of bit flips in the 256-byte IrisCode and compared the performance with the empirical performance concluding that their designed FAR model follows the empirical performance and is not affected by probability of bit flip. However, the modelled FRR does not corroborate with the empirical performance; theoretically, the performance degrades rapidly with an increased probability of bit flip, unlike their empirical observation. In 2005, Yoon et al. \cite{yoon2005individuality} explored the individuality of iris biometrics in an identification scenario, by transforming the many-class problem to a binary problem of intra-subject and inter-subject distinctiveness, as a factor of features, distance measures and classifiers, by ``showing the distinctiveness of the individual classes with a very small error rate in discrimination". The study concluded that considering a distance measure of histogram distances to compute intra-class and inter-class separability with multi-level 2D wavelets as features provides the best methodology to determine the individuality of iris biometrics out of the eleven methodologies tested in the study. In 2006, Daugman published a report \cite{daugman2006probing} on the analysis of approximately 200 billion imposter comparisons of the 256 bytes IrisCode from 632500 unique irides in an operational dataset obtained by special access from the UAE Ministry. In its assessment of the uniqueness of the IrisCode, the report concludes that for non-mated pair of irides, 35\% to 65\% of bits in the 2048 bits IrisCode do not match; in other words, at least 35\% of the iris bits being compared, after masking the 2048 bits IrisCode, from irides of two different persons, match. One gap is that, the study does not report on genuine comparisons. Considering multiple relative rotations of the IrisCodes during matching to account for the angle of rotation of the iris with respect to the camera during capture, the agreeing bits further increase to approximately 45\%. The report points to the impact of correlation from two sources - internal correlation present in IrisCodes due to the iris structure, and correlation introduced in the IrisCode during Gabor filtering, on the effective independent bit comparisons. In 2009 and subsequently in 2012 National Institute of Standards and Technology (NIST) conducted large scale (1.2 billion impostor comparisons from 8400 individuals and 1.2 trillion imposter comparisons respectively) evaluation of iris recognition systems from leading iris recognition industrial providers in IREX-I \cite{grother2009performance} and IREX-III \cite{grother2012irex}. However, the templates from these systems are proprietary black boxes which are ``non-standard, non-interoperable and not suitable for cross-agency exchange.'' The report provided an extensive assessment of quality factors (dilation,  occlusion, centre displacement, quality score), impact of image compression, template size (range: 257 bytes to 45080 bytes), computation time and accuracy trade-off, and their impact on recognition accuracy. The most relevant observations that relate to our study are: \begin{itemize}[leftmargin=*, noitemsep]
    \item Approximate size of standard iris image record (not templates; cropped, masked versions of the originally captured image) is approximately thirty kilobytes for large-scale identification (1:N) applications and is much lower for verification (1:1) application \cite{grother2009performance}. \textit{Thus, iris images with less distortion and high information content provide better performance in large-scale applications.}
    
    \item Removing poor quality images improves false non-match rate \cite{grother2009performance}. \textit{Thus, consideration of data quality in large-scale applications is an important factor.}
    
    \item False match rates are impacted by compression \cite{grother2009performance}
    
    \item  False Positive Identification Rate (FPIR) has a linear dependency on population size and threshold \cite{grother2012irex}; \textit{Thus, the threshold should be adjusted in the operational scenario based on population size}
    \item \textit{False positive cases are attributed to defective images, biological similarity and quality factors} \cite{grother2012irex}
\end{itemize}    

In 2015, Daugman \cite{daugman2015information} adapted information theory concepts including the Hidden Markov Model (HMM) to emulate the IrisCode, to compute the per-bit entropy of the IrisCode, and to further explain the anatomical and filter-induced correlations. An analytical methodology of the capacity of IrisCode is discussed which provides a quantitative understanding of the strong resistance of IrisCode against false matches. The report concludes that the high entropy of the IrisCodes per bit (0.469 bits of entropy per encoded bit), even in presence of biological and induced correlations, is the backbone of the high capacity of the IrisCode. This is supported by HMM predictions and NIST evaluations \cite{grother2009performance} \cite{grother2012irex} - accepting 36\% disagreement in bits between two IrisCode (i.e 64\% agreement of bits) as a match, leads to one case of the false match out of 24,000 imposter comparisons. However, the assessments of Daugman consider 256 bytes whereas NIST evaluations do not have information on the template dimension. 

Our work extends the state-of-the-art work by Daugman\cite{daugman2006probing} and NIST\cite{grother2009performance} \cite{grother2012irex} on empirical assessment of system capacity. While Daugman's assessment was specific to the capacity of the feature template, the IrisCode, we studied \textbf{ Daugman-style iris recognition systems from an end-to-end perspective, considering six different parameters in the iris recognition channel} addressing- 

\begin{itemize}[leftmargin=*, noitemsep]

\item How does identity count in a system impact capacity?
\item How does quality impact system capacity?
\item How does filter resolution impact system capacity?
\item Does a higher template dimension increase discriminable information in terms of system capacity?
\item Does template generation methodology impact system capacity?
\end{itemize}

NIST in their reports \cite{grother2009performance} \cite{grother2012irex} on large-scale N:N assessment with commercial ``black box" systems, discuss the importance of some of the parameters like quality and template size. We report a systematic study of different system parameters on constrained system capacity with publicly available datasets (refer Table~\ref{table:Dataset}) and open-source software (OSIRIS) with the scope of scientific reproducibility and continuation to address global challenges.

\section{Methodology}\label{sec:Methodology}

\subsection{Constrained Capacity}
Ideally, system capacity is the number of identities a system can correctly identify without any error. However, practically  biometric systems are prone to errors and function at an operating point which is a trade-off between an acceptable false accept rate and a false reject rate. Thus, the system capacity is computed at an acceptable error rate, and defined here as ``constrained capacity''. 

\textbf{Hypothesis:} There is an upper bound on the number of identities, $M$ a system can resolve and the number of features required, $n,$ to resolve the identities at an acceptable error rate.

We study the impact of $n,$ i.e optimum information content, on the capacity of the system. In our approach to optimize information content to achieve highly constrained system capacity, we study six different parameters: identity count in the system, feature dimension, filter resolution, random feature reduction, image quality and operating point, as further detailed below:

\begin{itemize}[leftmargin=*, noitemsep]
    \item \textbf{Quality:} ISO quality images (ISOQ) vs All quality images (ALLQ)
    \item \textbf{Filter Resolution:} Multi-resolution template vs Single-resolution template
    \item \textbf{Template Dimension:} Structured unwrapping at lower dimension - 26k bits (D2) vs 48k bits (D1)
    \item \textbf{Random feature reduction:}  Template features: 100\%,  75\%, 50\% ,25\%, 20\%, 15\% ,10\%   
    \item \textbf{Operating Point:} 0.1\% FAR vs 0.01\% FAR vs 0.001\% FAR
\end{itemize}

All biometric systems operate at a pre-defined threshold that indicates the system's expected FAR and FRR. Different thresholds may be chosen based on the system configurations i.e. the quality of images from which the templates are extracted, at what resolution the iris templates are extracted, the number of bits forming the template and how the feature points are selected, as detailed in Section~\ref{sec:EA}. We chose 24 different thresholds for the 24 different system configurations at fixed operating points (OP). We acknowledge the impact of the system configurations on the FRR and the necessity to choose system configurations at realistic OP. The selected OPs and the corresponding FRR are tabulated in Table~\ref{table:FAR_FRR}. Summarizing the key observations from Table~\ref{table:FAR_FRR}:
\begin{itemize}[noitemsep, leftmargin =*]
    \item Systems employing ISOQ dataset consistently renders low FRR across all comparing system configurations. The impact of quality on system errors is substantial.
    \item Systems employing single-resolution templates exhibit lower FRR than systems employing multi-resolution templates
    \item Systems employing low dimensional template (D2) outperforms high dimensional templates (D1) by 1\% - 8 \% FRR.
\end{itemize}

\begin{table}[!ht]
\footnotesize
\centering
\caption{\footnotesize False Reject Rate (FRR) at chosen operating points for 24 system configurations (S1 - S24) based on a different combination of feature dimensions (D1, D2), operating point (FAR), filter resolution (Single and Multi-Resolution), and quality (ALLQ, ISOQ)}
\label{table:FAR_FRR}

\begin{tabular}{|c|c|c|c|c|c|}
\hline
&  & \multicolumn{2}{c|}{\textbf{Multi Resolution}} & \multicolumn{2}{c|}{\textbf{Single Resolution }} \\
\cline{3-6}
\textbf{Feature} &\textbf{FAR} & \textbf{ALLQ} & \textbf{ISOQ} & \textbf{ALLQ} & \textbf{ISOQ}\\
\cline{3-6}
 \textbf{Dimension} & \textbf{(\%)} &\multicolumn{4}{c|}{\textbf{FRR (\%) at different Operating Point}} \\
\cline{2-6}
 \hline
 \hline
& 0.1 & S1: 16.6 & S2: 3.05 & S3: 11.19 & S4: 2.61 \\
\cline{2-6}
\textbf{D1} & 0.01 & S5: 24.9 & S6: 4.48 & S7: 16.59 & S8: 3.25  \\
\cline{2-6}
& 0.001 & S9: 33.9 & S10: 6.47 & S11: 22.01 & S12: 4.25 \\
\hline
\hline

& 0.1 & S13: 13.51 & S14: 2.59 & S15: 5.97 & S16: 0.71 \\
\cline{2-6}
\textbf{D2} & 0.01 & S17: 20.38 & S18: 5.30 & S19: 9.5 &  S20: 1.18 \\
\cline{2-6}
& 0.001 & S21: 27.99 & S22: 9.56 & S23: 14.48 & S24: 1.76 \\
\hline
\end{tabular}
\end{table}

\begin{figure*}[h]

\centering
    \includegraphics[width=18cm,keepaspectratio]{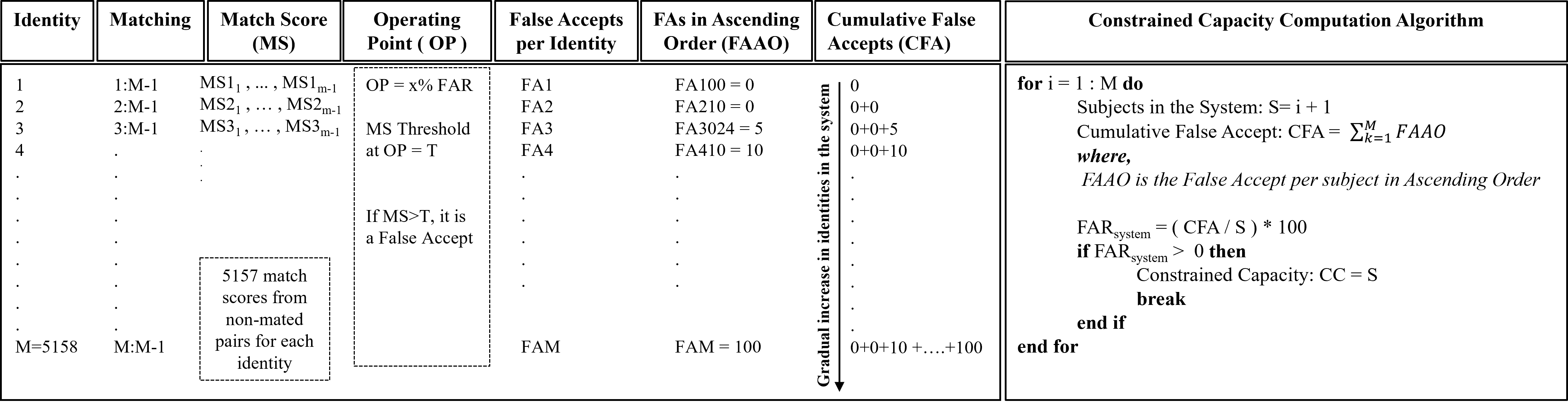}
    \caption{\footnotesize An illustrative representation of Constrained Capacity Computation Algorithm: The database contains `M' unique identities. Considering a single sample per identity, each identity is matched with M-1 other identities computing M-1 match scores per identity. Based on the predefined threshold at x\% FAR (OP), each identity (i) encounters $FA_i$ false accepts. Identities are arranged in ascending order of the $FA_i$, represented here as FAAO. The number of identities in the system is increased based on FAAO. The number of identities the system could accommodate before encountering any identity clash is the constrained capacity.}
    \label{fig:CapacityComputation}
\vspace{-4mm}
 \end{figure*}

 For each system structure, we compute the identity clash (false accepts) of each unique identity i.e. the number of times an iris template matches with another iris template of a different identity, at each OP. We compute the constrained capacity of the system at the specific OP for each system structure as we increase, $M,$ the number of identities in the system, by gradually adding one identity at a time in the ascending order of identities with no false accepts with any other subjects. Then the identities are sorted in terms of the number of false accepts per identity. Constrained capacity is the number of subjects the system could resolve before encountering the first case of false accept. The computation of the constrained capacity is illustrated in Figure~\ref{fig:CapacityComputation}. 

\begin{figure}[h]
\centering
    \includegraphics[width=8.5cm]{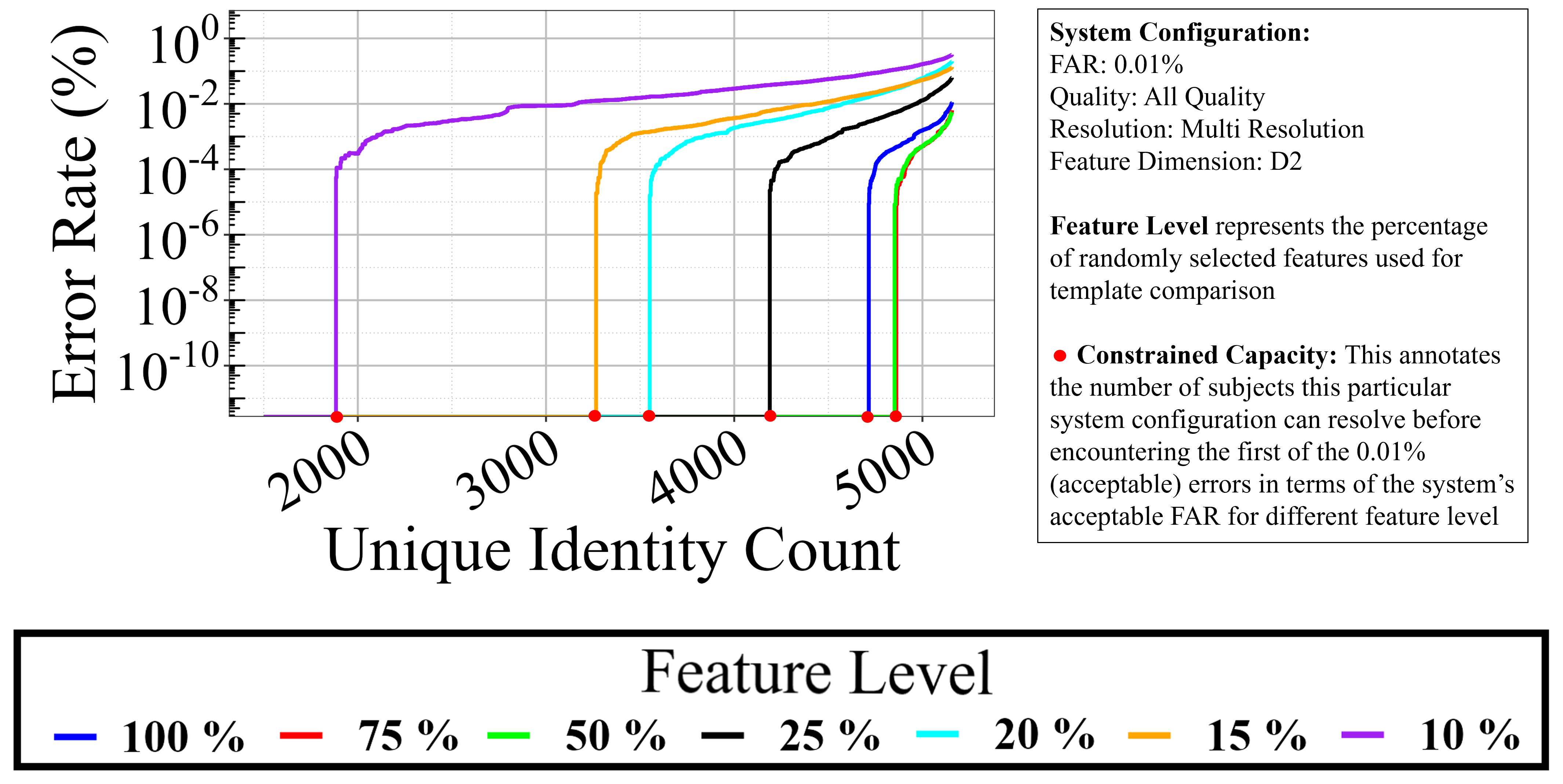}
    \caption{\footnotesize D2, Multi-Resolution, All Quality, FAR=0.001\% : Error rate as a result of the increase in the number of identities in the system for different feature levels. The number of identities the system resolves before encountering the first error is the constrained capacity. }
    \label{fig:ErrorCount_Subject}
    \vspace{-3mm}
 \end{figure}

An illustration of the relationship between the error rate as the number of identities is increased in a system is shown in Figure~\ref{fig:ErrorCount_Subject} for one of the 24 system configurations studied. The figure illustrates the variation in constrained capacity (CC) and the error rate of the system as random radial features  (columns: binary codes at a fixed angle) are eliminated from the feature template for seven different feature levels. We note that as identities are gradually increased in the system (x-axis) in the ascending order of false accepts, the cumulative false accepts (CFA) increases after resolving a certain number of identities. Constrained capacity is the point in the x-axis where the CFA is no longer zero.

Figure~\ref{fig:Error_IdentityCount_FeatureLevel} shows error rate as a function of increased identities for different feature levels for all 24 system configurations. The number of identities the system could resolve at a predefined OP (acceptable number of identity clashes) is the constrained capacity of that system.  

\subsection{Experimentation Setup}\label{sec:EA}
This section provides a detailed background of the concepts leading to constrained capacity estimation analysis - iris code generation impacting the template dimensions and resolutions, random feature selection and image quality. 

\begin{figure}[h]
\centering
    \includegraphics[width=3.5in,keepaspectratio]{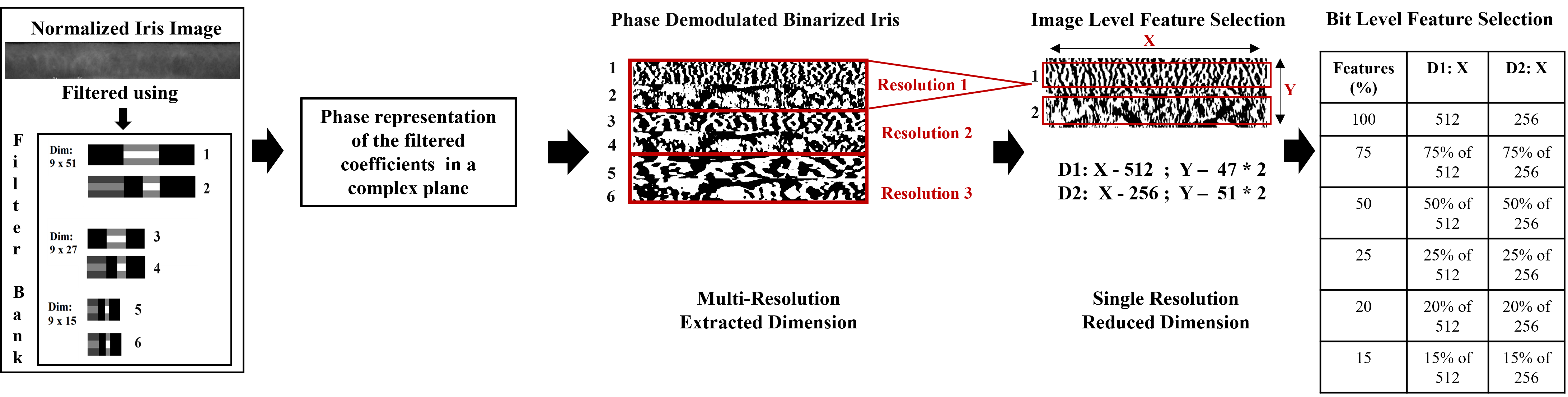}
    \caption{\footnotesize Basic steps of IrisCode generation with multiple filter resolutions, feature dimensions and random feature reduction}
    \label{fig:IrisSteps}
    \vspace{-4mm}
 \end{figure}

 \subsubsection{IrisCode Extraction}
 
 The annular iris area is extracted from the captured iris image by segmentation and then unwrapped into a rectangular representation. It is then filtered using a Gabor filter bank and a patch-wise phase representation of the filtered samples in a complex plane is performed. The quadrant, where the resulting phasor for that patch of the iris is projected on the complex plane, is identified. Both the real and imaginary parts of the phase representation are considered in the generation of the IrisCode with a 2-bit representation (00,01,11,10). The same dimensional mask is generated for each template which identifies areas of obstruction like eyelashes, eyelids, etc. The methodology follows Daugman's approach \cite{daugman1993high}; iris feature extraction methodology is graphically illustrated in Figure~\ref{fig:IrisSteps}.
 
 \subsubsection{Feature Dimension and Resolution}\label{sec:FDR}
 
 For this study, the IrisCode is generated at different saptial resolutions and dimensions to assess the impact of information content. We extracted features of two different dimensions, $\sim$26k bits and $\sim$48k bits, based on the filtering at the unwrapping stage, i.e., patch-wise translation of the raw iris image to the phasor representation, varying the patch size. Each of these dimensions is extracted at 3 different resolutions of the Gabor filter (Filter Dimensions: 9*51, 9*27, 9*15). Feature templates are developed for single-resolution and multi-resolution (i.e., the combination of features extracted at all three resolutions).

 The optimal filter design as developed by Daugman remains unpublished and proprietary. The three filters used in our analysis were designed by the developers of OSIRIS for the original template dimension of 64 * 512 (D1).  D2 is a downsampled representation of D1 by the unwrapping mechanism. For a direct comparison, ideally, the filters could be proportionally downsampled. However, we have used the same filters for the $64 *512 $(D1) dimensional template and $70 * 256 $ (D2) dimensional template. By not changing the filter design for D2, we have essentially used different filters in our analysis of the two dimensions. 
 For single-resolution template assessment, we use the first (dimension: 9 x 51) of the 3 filters in the filter bank. Thus, the feature content extracted on filtering D2  with the single-resolution filter is different from D1 given that the filtering is performed on a larger spacial area of lower spatial resolution. For multi-resolution template assessment, each consecutive filter is a downsampled representation of the prior filter. Thus, by design, the multi-res template partially extracts similar frequency bands in both D1 and D2. Thus, D1 and D2 are templates generated with different spatial resolutions and different frequency components for the same iris sample. 
 
 To reduce noise from the extracted feature templates, approximately 27\% of the area around the two edges of the iris is removed; 9\% of the area from the pupillary boundary and 18\% of the area from the limbus boundary is eliminated, which has the highest potential noise in terms of obstruction from eyelash, eyelids which may remain undetected in the iris mask. A graphical representation of the different resolution templates, the reduced dimensions are shown in Figure~\ref{fig:IrisSteps}. Table~\ref{table:IrisCode Dimension} tabulates the extracted feature dimension (D1, D2) of IrisCode at a single resolution (Col: Extracted Dimensions), the reduced dimensions of the feature template after removing potential noisy bits (Col: Reduced Dimensions), and the operational pixels/bits at single and multi-resolution (Col: Bit Count by Resolution).

\begin{table}[!ht]
\scriptsize
\centering
\caption{IrisCode Dimension Summary}
\label{table:IrisCode Dimension}
\begin{tabular}{|c|c|c|c|c|}
\hline
& \textbf{Extracted}  & \textbf{Reduced} & \multicolumn{2}{c|}{\textbf{Bit Count by Resolution}}\\
\cline{4-5}
& \textbf{Dimensions} &\textbf{ Dimensions} & \textbf{Single} & \textbf{Multi}  \\
\hline
\hline

\textbf{D1} & 64 *512 & 47 * 512 & 48128 ($\sim 5kB$) & 144386 ($\sim 18 kB$) \\
\hline
\textbf{D2} & 70 *256 & 51 * 256 & 26116 ($\sim 3kB$) & 78348 ($\sim 9kB$) \\

\hline
\end{tabular}
\vspace{-4mm}
\end{table}

\subsubsection{Random Feature Reduction}\label{sec:RFR}
We explore random feature reduction at the bit-level to assess the impact of feature content on system capacity. Feature content after bit-level random feature reduction is shown in Figure~\ref{fig:IrisSteps}. \\ 
\textbf{Bit-level Random Feature Reduction:} For each dimension, the extracted feature template represents the entire raw iris image. We randomly eliminate radial features (columns: binary codes at a fixed angle) at the bit level for multiple random feature reduction levels (0\%, 25\%, 50\%, 75\%, 80\%, 85\% and 90\% of the extracted feature templates of different dimensions). After elimination, the remaining percentage of features in the template is referred to as `Feature level' in this report. The implementation induces randomization. Random radial feature columns were selected. The same columns were eliminated from the mating pair of samples before matching. The process was repeated for every mating pair. Different feature columns were eliminated for different pairs.  

The idea is to compare bit-level random feature reduction, with the structured generation of reduced feature dimensions at the image unwrapping level, as described in Section~\ref{sec:FDR}.

\subsubsection{Quality}\label{sec:Quality}
\begin{figure}[h]
\centering
    \includegraphics[width=3.5in,keepaspectratio]{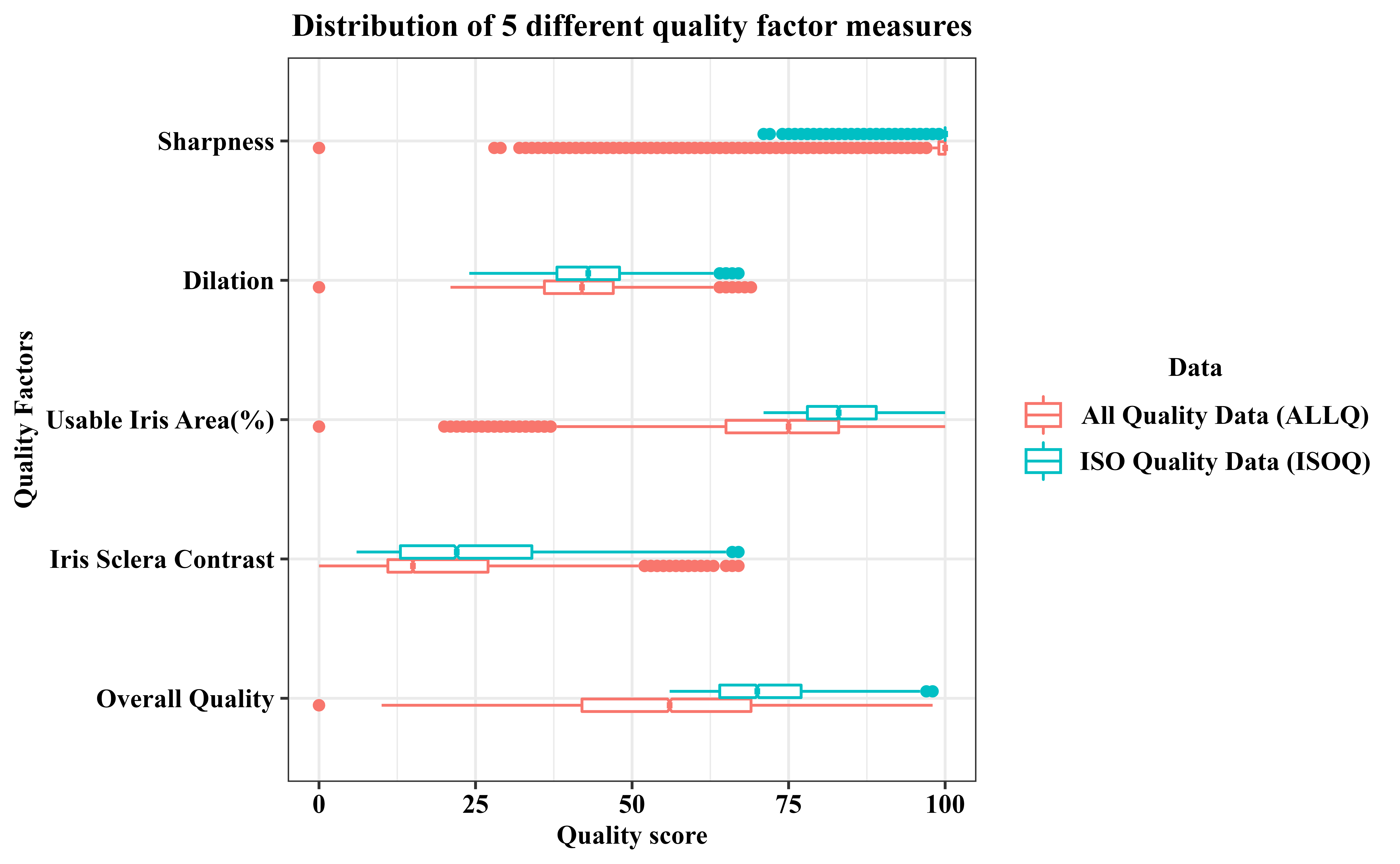}
    \caption{\footnotesize Comparative boxplot of five quality score distribution of the entire dataset vs ISO quality standard based cleaned dataset }
    \label{fig:Quality_Factor_Distribution}
\vspace{-3mm}
 \end{figure}

Quality impacts performance. We choose the best quality samples because we are trying to understand capacity when quality problems are minimized. International standards have been set to benchmark iris image quality for optimal performance \cite{iris_standard_report}. We assess the impact of quality on system capacity and its importance on large-scale operations. We assess 2 scenarios- 
\begin{itemize}[leftmargin=*, noitemsep]
    \item  \textbf{ISO Quality Data (ISOQ):} All iris samples follow ISO standard- ISO-IEC- 29794-6 \cite{iris_standard_report}. Nine quality factors encompassing anatomical metrics and illumination are considered in the selection of the samples - overall quality score, iris radius, dilation, usable iris area, iris-sclera contrast, iris-pupil contrast, grayscale utilization, iris-pupil concentricity,  margin-adequacy.

    \item \textbf{All Quality Data (ALLQ):} The best quality sample from an individual based on the overall quality irrespective of whether it follows ISO standards is selected.  The distribution of five quality measures for both ISOQ and ALLQ  is shown in Figure~\ref{fig:Quality_Factor_Distribution}.
\end{itemize}

\subsection{Dataset}
Empirically assessing the capacity of an iris recognition system requires access to a large dataset. No single publicly available dataset has a large number of subjects for this research. For our study, we created a composite dataset of 5158 unique irides putting together multiple independent, publicly available datasets from different sources as summarized in Table~\ref{table:Dataset} to form the ALLQ dataset. A subset of 2982 unique irides is filtered based on ISO quality standards to form the ISOQ dataset. Right and left irides from the same subjects are considered different identities in terms of uniqueness. For our analysis, samples were chosen based on the best overall quality score. The experiment was designed such that exactly one sample with the best quality score would be used for enrollment. For non-mated comparisons, each sample of the M-enrolled samples was matched against M-1 samples from different identities; total imposter comparisons for ALLQ systems: 5158 * (5158-1) / 2 $\simeq$ 13.2 million; total imposter comparisons for ISOQ systems: 2982 * (2982-1) / 2 $\simeq$ 4.4 million. For mated comparisons, the two next-best-quality samples from each unique iris were selected for matching against the enrolled sample; total genuine comparisons for ALLQ systems: 5158 * 3 = 15474; total genuine comparisons for ISOQ systems: 2982 * 3 = 8946.

\begin{table}[!ht]
\scriptsize
\centering
\caption{Dataset Summary}
\label{table:Dataset}
\begin{tabular}{|c|c|c|}
\hline
Database & Unique Irides & Sensor\\
\hline
\hline

ITR Clarkson & 484 & OKI IRISPASS \\
\hline
ND CrossSensor 2012 & 1353 & LG2200 \\
\hline 
CASIA Lamps & 820 & OKI IRISPASS-h\\
\hline
CASIA Twins & 400 & OKI IRISPASS-h\\
\hline
CASIA Thousand & 2000 & IKEMB-100\\
\hline
CASIA Interval & 395 & CASIA Iris Camera\\
\hline
\end{tabular}
\vspace{-4mm}
\end{table}

\subsection{Algorithms}
For quality assessment we used a commercial software VeriEye 11.0 SDK \cite{VerieyeSDK} following ISO/IEC 29794-6\cite{iris_standard_report}. The software computes iris quality factors following ISO guidelines.
For feature extraction, we adapted the open-source software, OSIRIS\cite{othman2016osiris}. OSIRIS was developed following Daugman's approach of iris feature extraction \cite{daugman1993high}, IrisCode, and allows flexibility in feature extraction in terms of dimension and resolution. This is an important element in exploring beyond the conventional standard in iris recognition. Conventionally 256 bytes (1028 bits) of IrisCode are generated and used for matching. OSIRIS allows exploration of the customized dimensions of feature extraction by considering different dimensional patches at the unwrapping stage. Additionally, the software has the capacity to generate IrisCode  at three different resolutions of the Gabor filter. Exploring feature content in the template goes to the core of our study on the impact of structured template generation at different dimensions (D1, D2) versus random feature reduction (0\% to 90\%) on the capacity of the iris recognition system. 

The extracted iris templates of different dimensions with different frequency components (D1, D2), filter resolution (single and multi-resolution), quality (ISOQ, ALLQ) and after random feature reduction (0\% to 90\%), are used for template comparison for mated and non-mated pairs of images. A corresponding mask is generated for each template. We developed a method to perform the $M:M-1$ matching for non-mated pairs ($\sim$ 13.2 million for ALLQ systems and $\sim$ 4.4 million for ISOQ systems) of images for each set of the 24 system configurations, where $M$ represents the number of unique identities. The matching methodology involves computation of the Hamming distance between two templates for different system configurations following equation \ref{eq:hd}.
\newenvironment{conditions}
  {\par\vspace{\abovedisplayskip}\noindent\small\begin{tabular}{>{$}l<{$} @{${}={}$} l}}
  {\end{tabular}\par\vspace{\belowdisplayskip}}
 
\begin{equation}\label{eq:hd}
\footnotesize
    HD =\frac{ || (IC1 \otimes IC2) \cap M1 \cap M2 ||} {|| M1 \cap M2 ||}
\end{equation}
where IC denotes IrisCode and M denotes Mask.\\
56 shifts (28 on each side) for D1 and 28 shifts (14 on each side) for D2 of the iris template are performed during Hamming distance (HD) computation to mitigate errors induced by the angle of rotation of the mating irides. Each shift corresponds to 0.7 degrees and 1.4 degrees for D1 and D2 respectively, leading to a cumulative flexibility of 19.6 degrees on each side. The minimum HD  (or best match) is considered the final score for each pair of comparing templates.

Next, we compute the constrained capacity of the system, i.e. the number of identities the biometric system can accommodate before encountering an identity clash, for each of the 24 system configurations.

\subsection{M:M Matching and Computational Aspects}

\SI{1.6e10} iris code comparisons are needed to calculate the non-match scores required for this study. Each of the 12 ALLQ system configurations requires roughly 13.3 million iris comparisons and 12 ISOQ system configurations require roughly 4.4 million iris comparisons. 
 Since our iris matching algorithm includes an iris alignment step, we need to consider all the possible non-match scores computed by different alignments of the gallery and probe irides. To compute the optimal iris alignment, we calculate the distance matrix for each shift of the probe iris template up to 19.6 degrees on each side and then identify the minimum distance of all possible shifts as the final score. This amounts to 57 and 29 probe iris code pixel shifts (and corresponding distance matrices) for the D1 and D2 templates respectively. We then repeat the computation for each system configuration (filter resolution, feature dimension, random feature reduction).
A large number of comparisons required us to develop a custom framework as we projected that sequential computation of the distances would require 62 days to complete using a high-end computational server.
Since 99.96 \% of comparisons in our analysis are in the non-match distribution, we focus our optimization effort on this set. Our final comparison engine is able to compute all computations in 1.99 days using one server, with a proportional decrease in time with each additional server tasked. Intel Xeon Cascade Lake R computational server was leased using the Chameleon testbed \cite{keahey2020lessons}.

\section{Results}\label{sec:Results}

\begin{figure}[h]
\centering
    \includegraphics[width=3.5in,keepaspectratio]{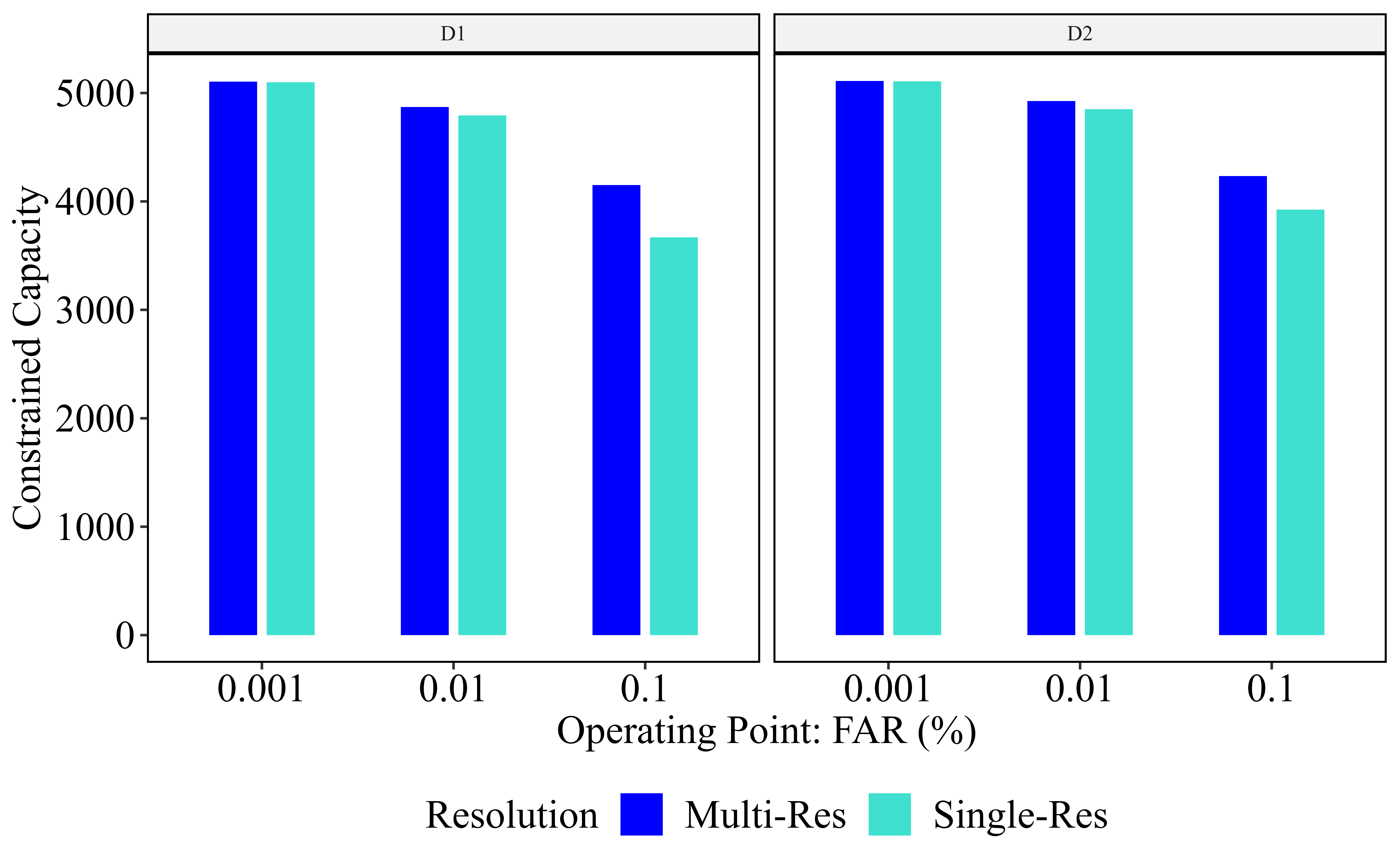}
    \caption{Constrained Capacity of ALLQ Dataset  with 50\% feature level at different parametric compositions }
    \label{fig:CC_ALLQ_FL50}
\vspace{-3mm}
 \end{figure}

This section discusses constrained system capacity as a function of multiple parameters--the number of identities in the system, image quality (ALLQ, ISOQ), filter resolution (single and multi-resolution), template dimension and filter design (D1, D2), random feature reduction (0\% to 90\%) and operating points (0.1, 0.01, 0.001\% FAR). Different combinations of these parameters define the structure of a system. Overall 24 different system configurations (refer Table~\ref{table:FAR_FRR}) are studied and reported.  Table~\ref{table:ALLQ_FA_NICF} and Table~\ref{table:ISOQ_FA_NICF} provide a detailed report on the impact of different system parameters on the imposter distribution of $\sim 13.2$ million comparisons ($ (5158 * 5157)/2 $) for ALLLQ dataset and ISOQ dataset. Figure~\ref{fig:CC_ALLQ_FL50} provides a graphical representation of constrained capacity for the ALLQ dataset at 50\% feature level, the best performing feature level, to depict the impact of three parameters- operating point, resolution and feature dimension. The ISO dataset representation is similar and thus not included in this report. The tables report the number of cases of false accepts (FA), the corresponding false accept rate (FAR), constrained capacity (CC) and percent capacity (PC) at specific operating points ( OP: 0.1\% / 0.01\% / 0.001\% FAR). The operating threshold (Hamming distance (HD)) for OP is different for each system configuration (filter resolution, template dimension and data quality). For example, a system operating with ISO quality data would operate at a different HD than a system operating with ALLQ data for the same FAR, as the imposter distribution of both systems would be different. However, for different feature dimensions (100\% to 10\% features), the HD has been fixed at 100\% features for that operating point. \par 
We report on the impact of each of the system configurations and discuss the trade-off between different parameters in a system to achieve high system capacity in realistic operational scenarios. 

 \subsection{Capacity Assessment: Impact of Identity Count on System Capacity}
\begin{sidewaysfigure*}
\centering
\scriptsize

\makebox[0.15\textwidth][c]{\textbf{MULTI RESOLUTION}}
\hfill
\makebox[0.15\textwidth][c]{\textbf{MULTI RESOLUTION}}
\hfill
\makebox[0.15\textwidth][c]{\textbf{MULTI RESOLUTION}}
\hfill
\hspace{1em}
\makebox[0.15\textwidth][c]{\textbf{SINGLE RESOLUTION}}
\hfill
\makebox[0.15\textwidth][c]{\textbf{SINGLE RESOLUTION}}
\hfill
\makebox[0.15\textwidth][c]{\textbf{SINGLE RESOLUTION}}\\

\makebox[0.15\textwidth][c]{\textbf{OP: 0.1\%}}
\hfill
\makebox[0.15\textwidth][c]{\textbf{OP: 0.01\%}}
\hfill
\makebox[0.15\textwidth][c]{\textbf{OP: 0.001\%}}
\hfill
\hspace{1em}
\makebox[0.15\textwidth][c]{\textbf{OP: 0.1\%}}
\hfill
\makebox[0.15\textwidth][c]{\textbf{OP: 0.01\%}}
\hfill
\makebox[0.15\textwidth][c]{\textbf{OP: 0.001\%}}
\hfill\\
\vspace{1em}
\subcaptionbox{ \footnotesize D1: ALLQ\label{fig:D1_AR_ALLQ_0.1}}
    {\includegraphics[width=3.5cm, height=3cm,keepaspectratio]{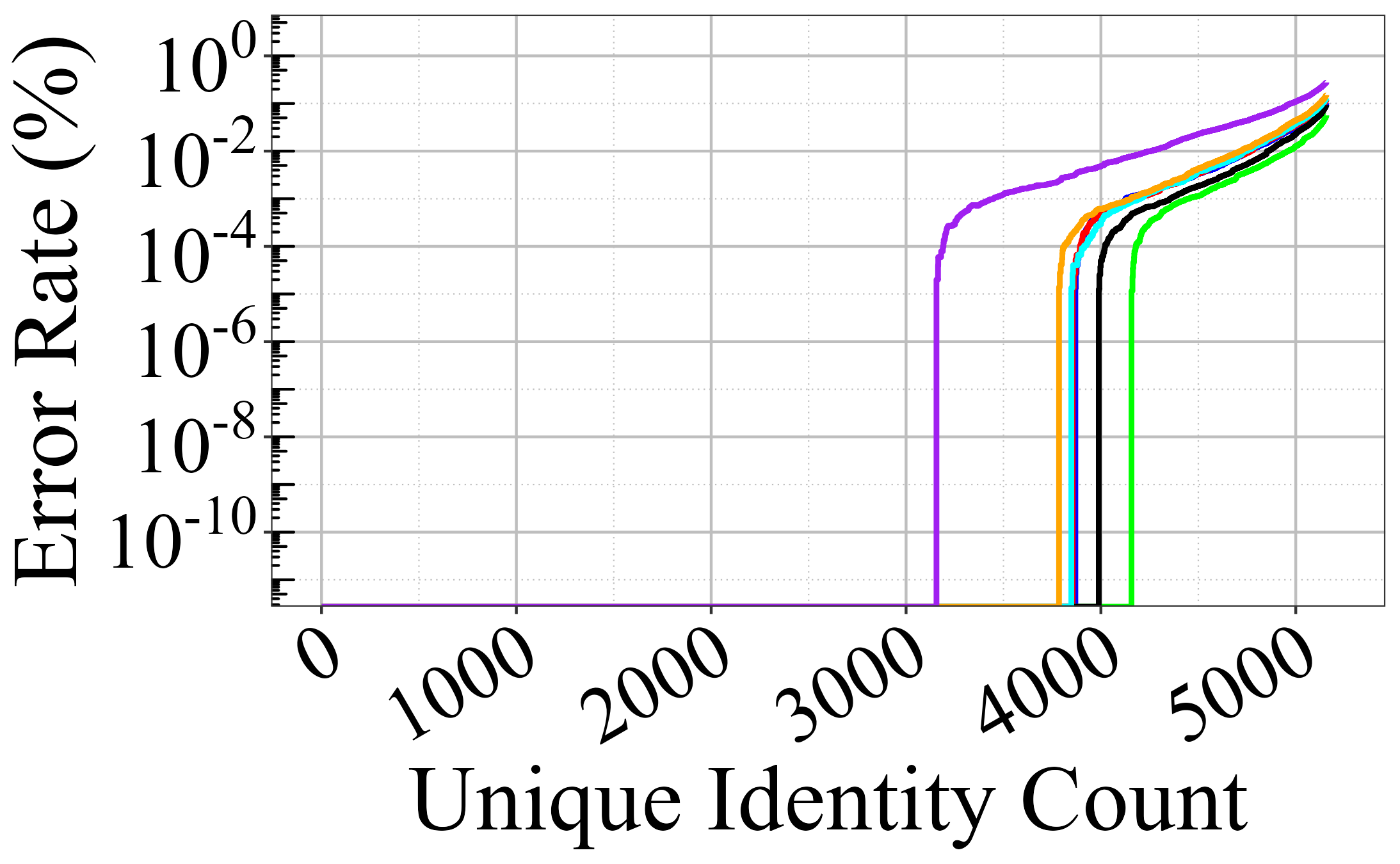}}
\hfill   
\subcaptionbox{\footnotesize D1: ALLQ\label{fig:D1_AR_ALLQ_0.01}}
    {\includegraphics[width=3.5cm, height=3cm,keepaspectratio]{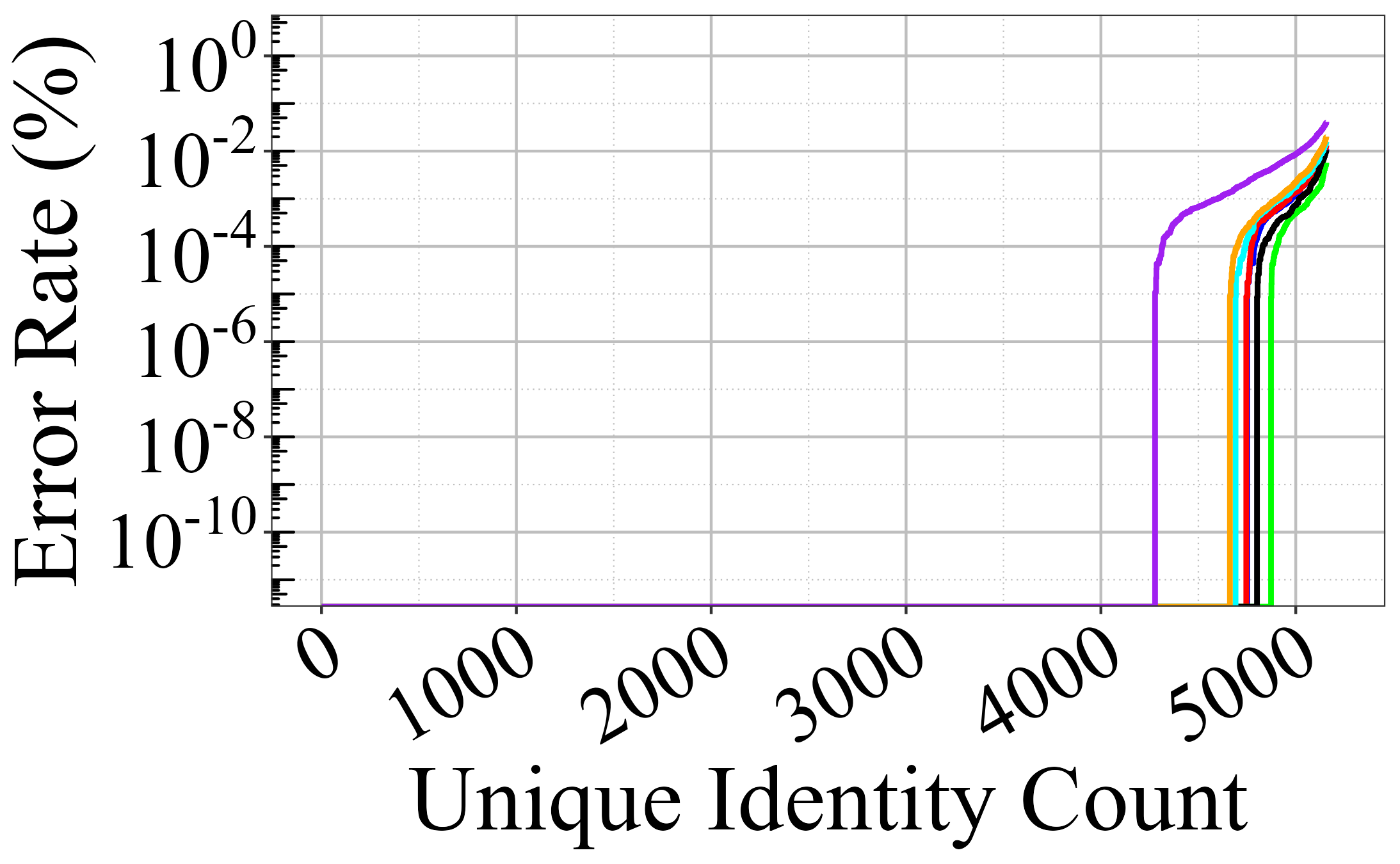}}    
\hfill  
\subcaptionbox{\footnotesize D1: ALLQ\label{fig:D1_AR_ALLQ_0.001}}
    {\includegraphics[width=3.5cm, height=3cm,keepaspectratio]{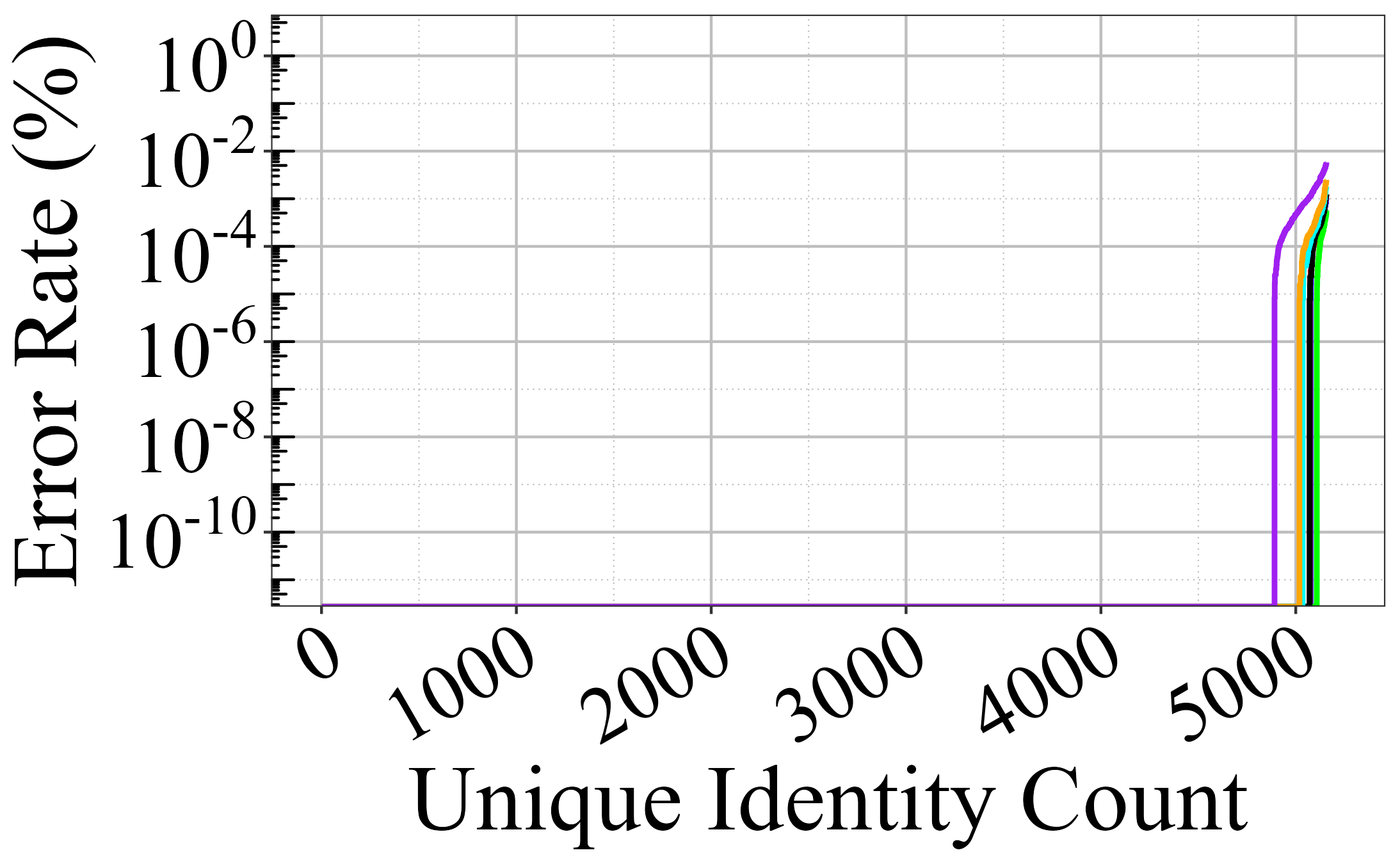}}
\hfill
\vrule
\hspace{1em}
\subcaptionbox{\footnotesize D1: ALLQ\label{fig:D1_SR_ALLQ_0.1}}
    {\includegraphics[width=3.5cm, height=3cm,keepaspectratio]{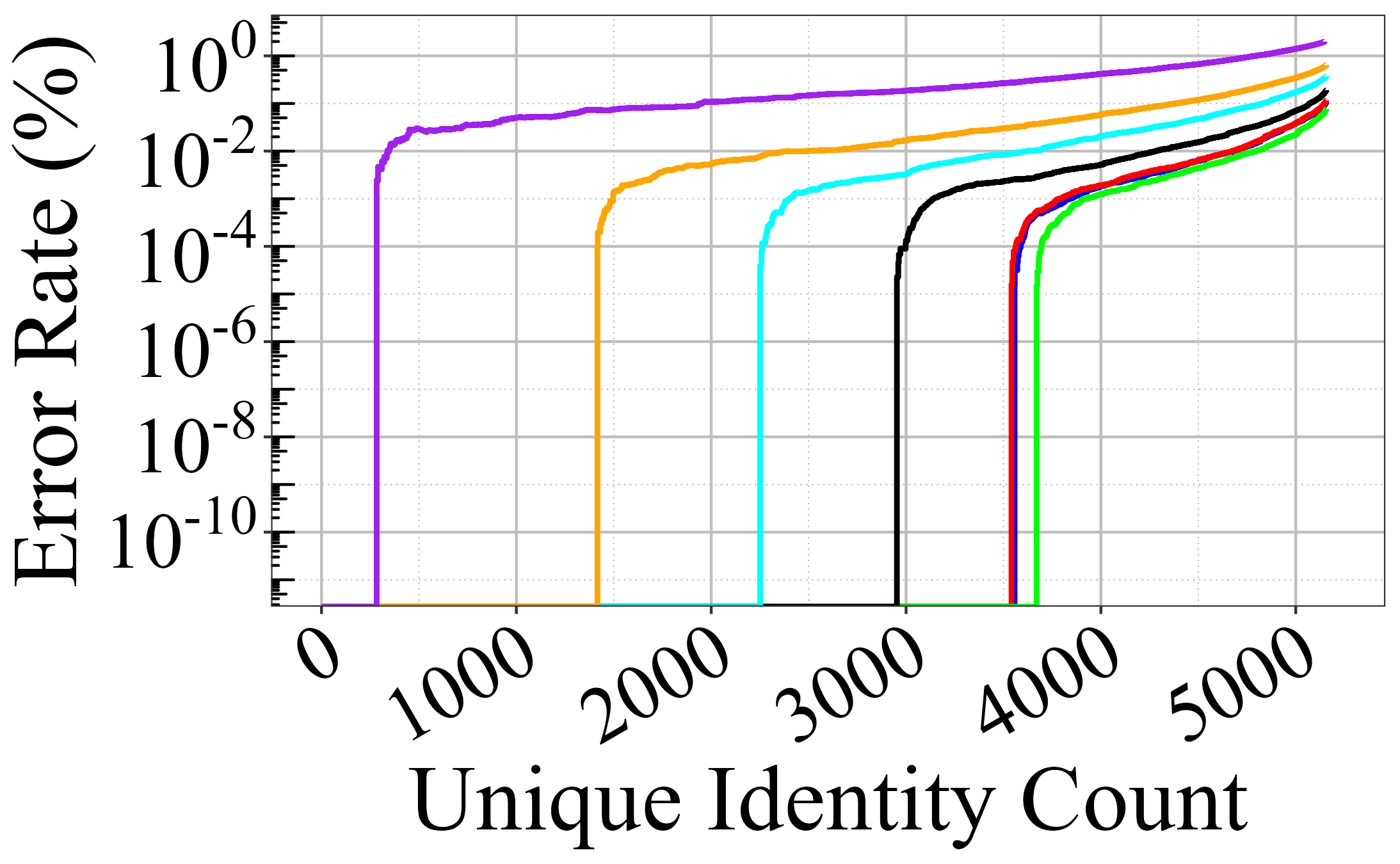}}
\hfill        
\subcaptionbox{\footnotesize D1: ALLQ\label{fig:D1_SR_ALLQ_0.01}}
    {\includegraphics[width=3.5cm, height=3cm,keepaspectratio]{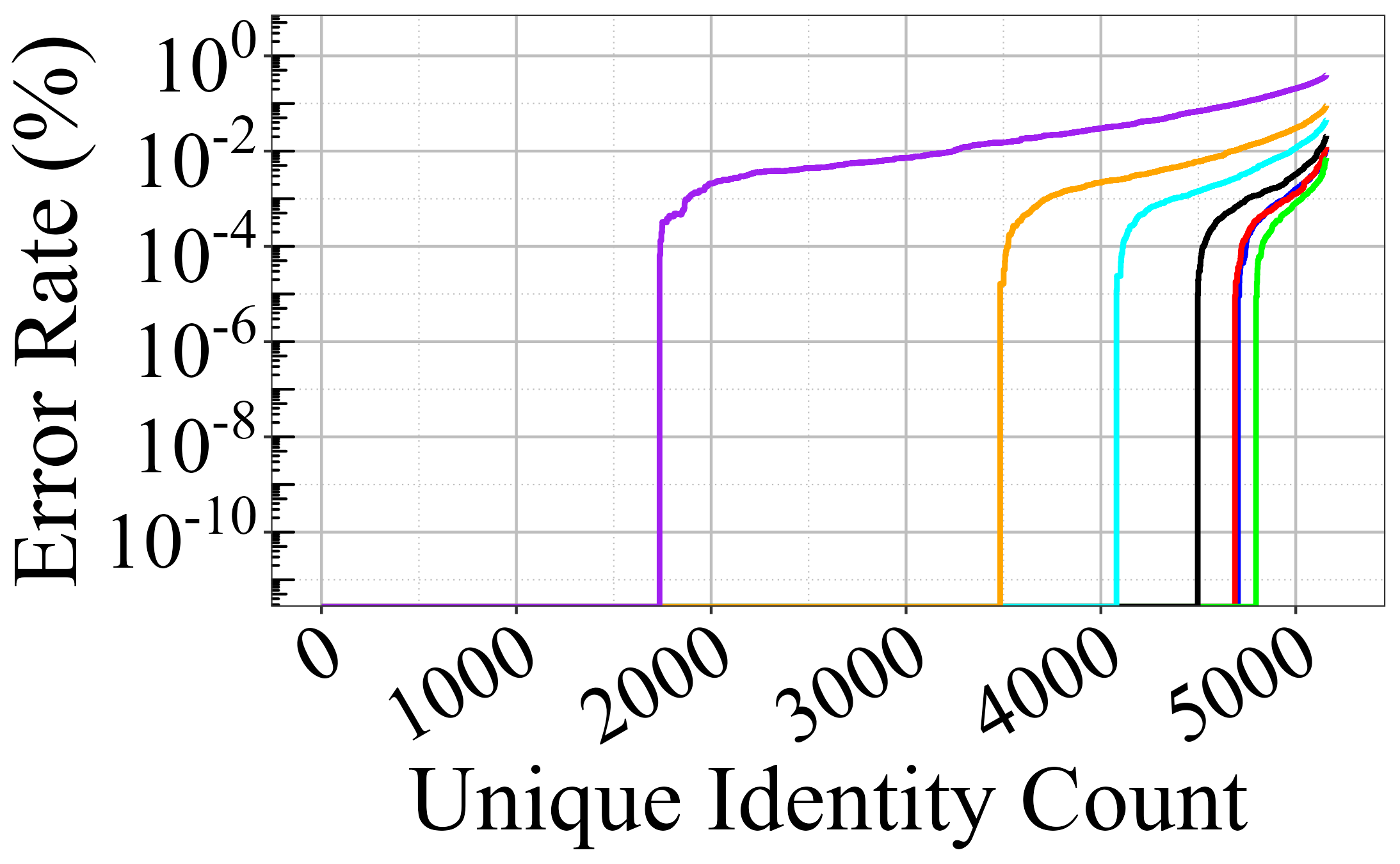}}  
\hfill      
\subcaptionbox{\footnotesize D1:ALLQ\label{fig:D1_SR_ALLQ_0.001}}
    {\includegraphics[width=3.5cm, height=3cm,keepaspectratio]{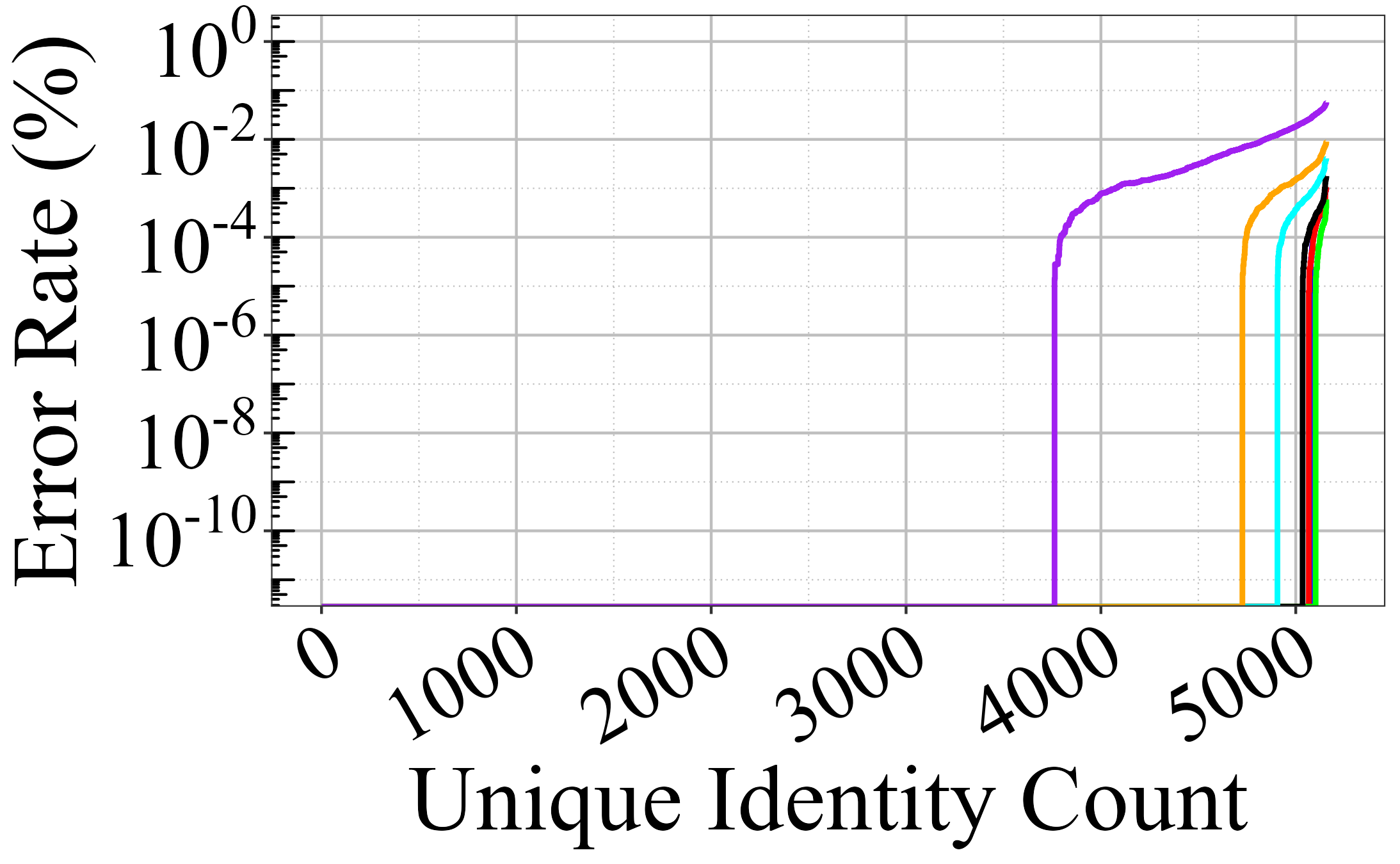}}
 \hfill
 \par\bigskip
\subcaptionbox{\footnotesize D2: ALLQ\label{fig:D2_AR_ALLQ_0.1}}
    {\includegraphics[width=3.5cm, height=3cm,keepaspectratio]{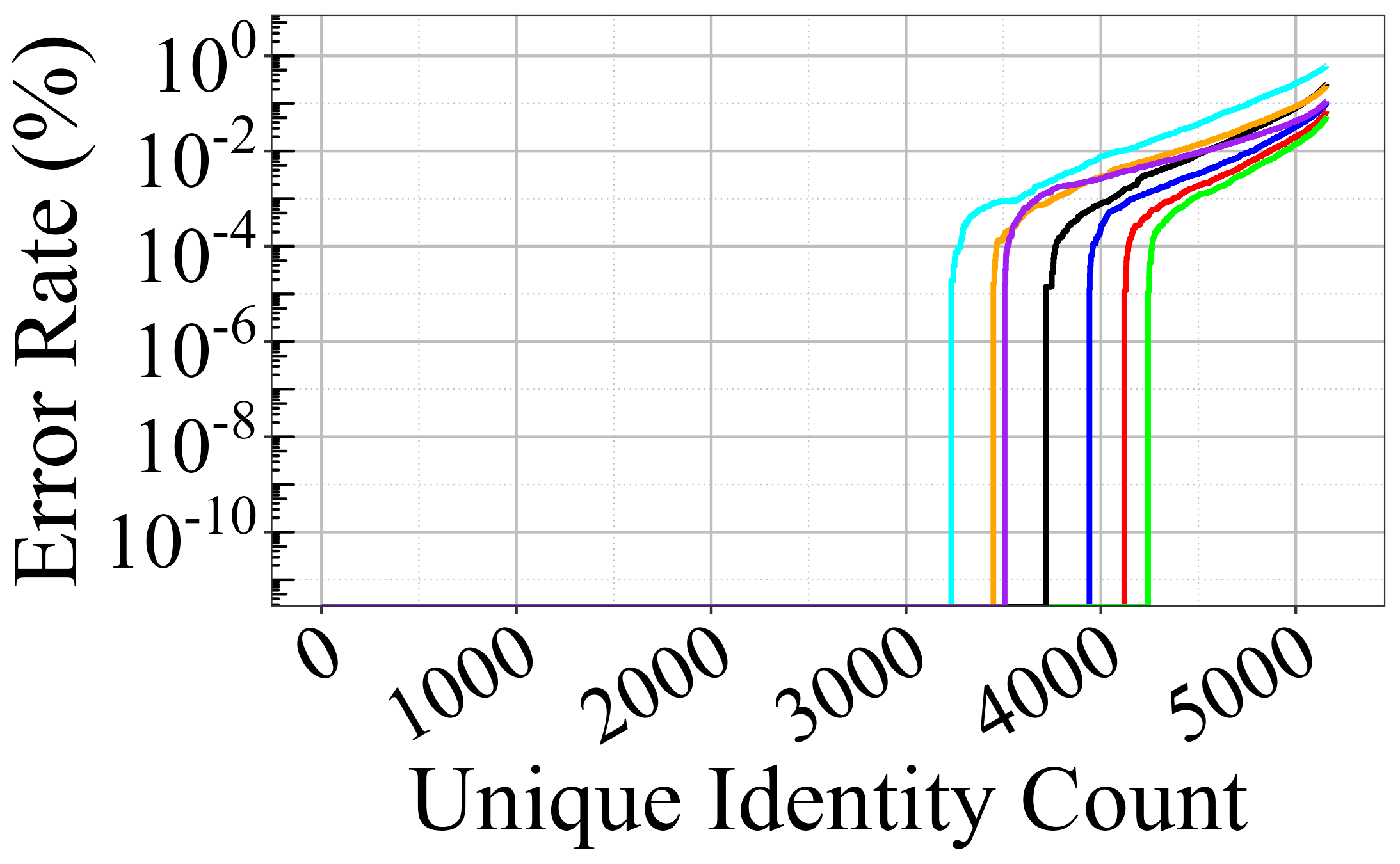}}
\hfill    
\subcaptionbox{\footnotesize D2: ALLQ\label{fig:D2_AR_ALLQ_0.01}}
    {\includegraphics[width=3.5cm, height=3cm,keepaspectratio]{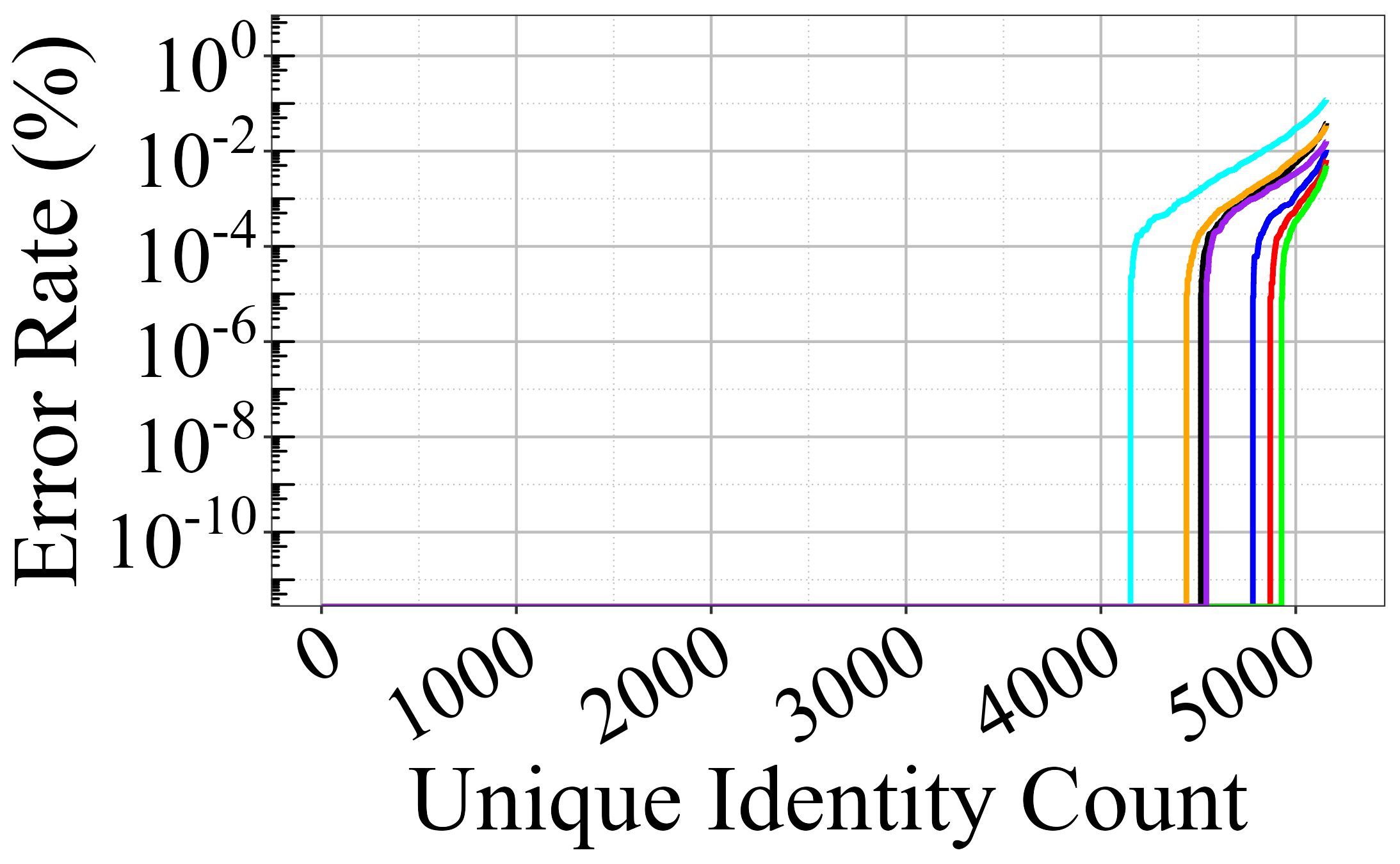}}    
\hfill      
\subcaptionbox{\footnotesize D2: ALLQ\label{fig:D2_AR_ALLQ_0.001}}
    {\includegraphics[width=3.5cm, height=3cm,keepaspectratio]{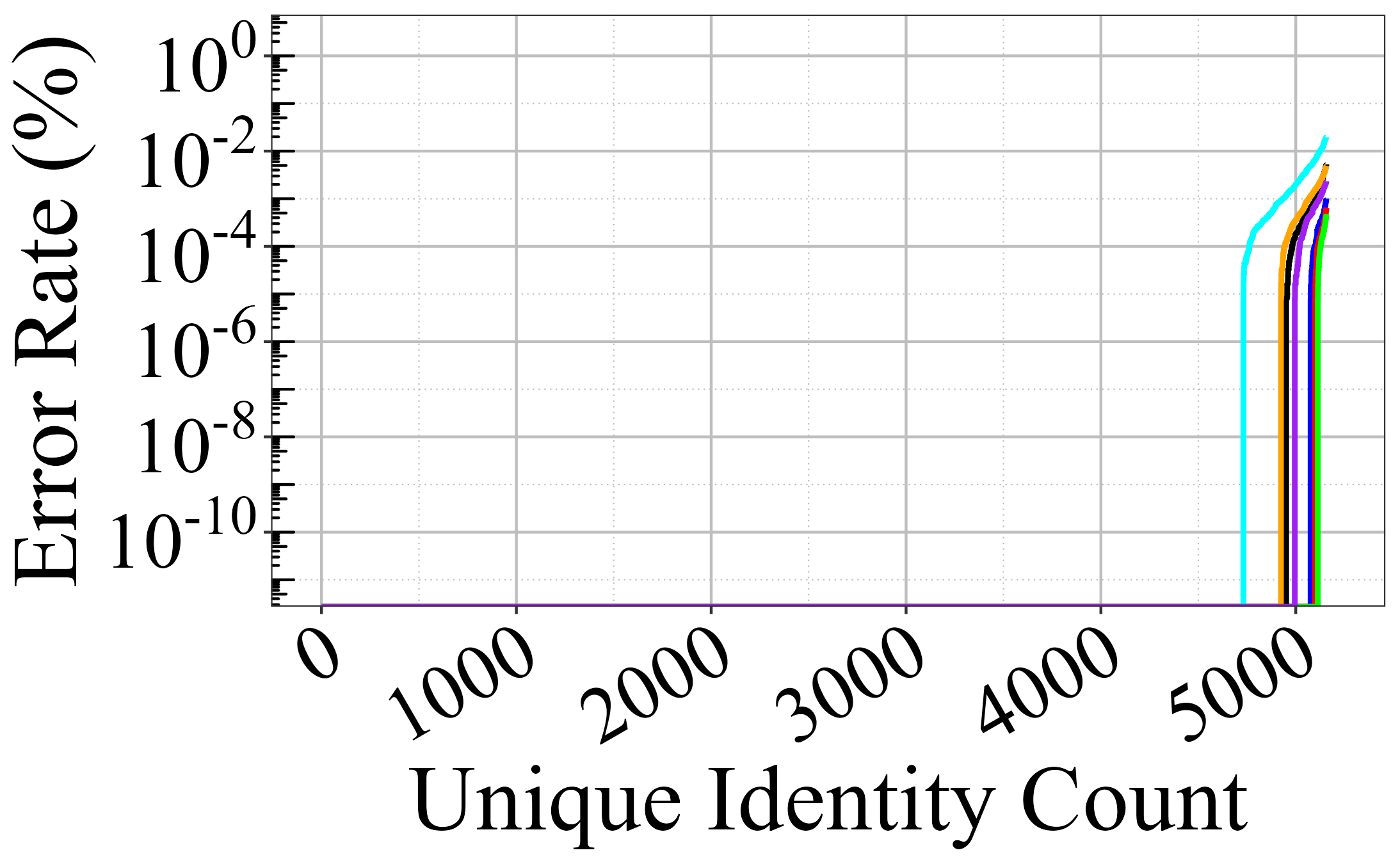}}
\hfill  
\vrule
\hspace{1em}
\subcaptionbox{\footnotesize D2: ALLQ\label{fig:D2_SR_ALLQ_0.1}}
    {\includegraphics[width=3.5cm, height=3cm,keepaspectratio]{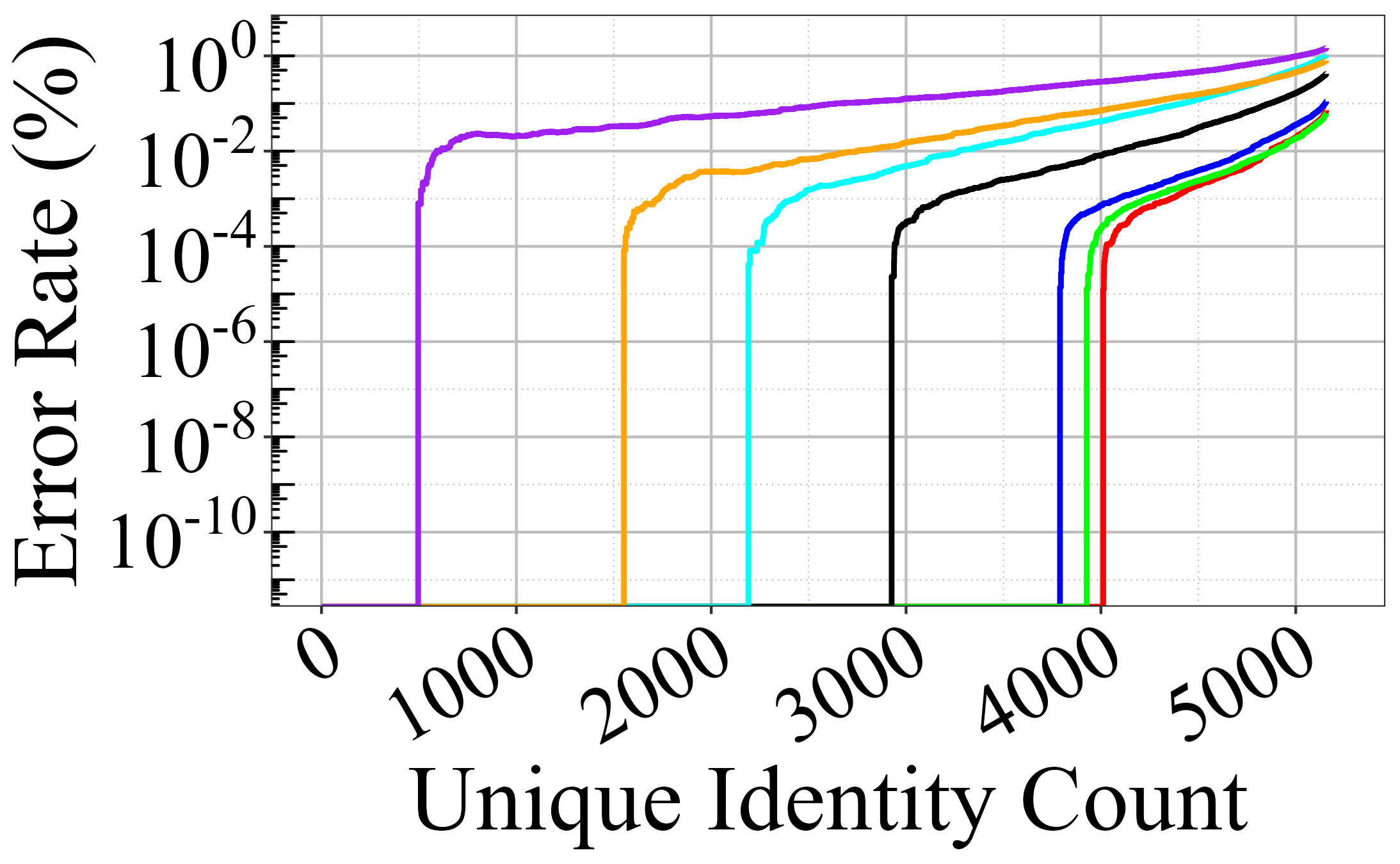}}
\hfill        
\subcaptionbox{\footnotesize D2:ALLQ\label{fig:D2_SR_ALLQ_0.01}}
    {\includegraphics[width=3.5cm, height=3cm,keepaspectratio]{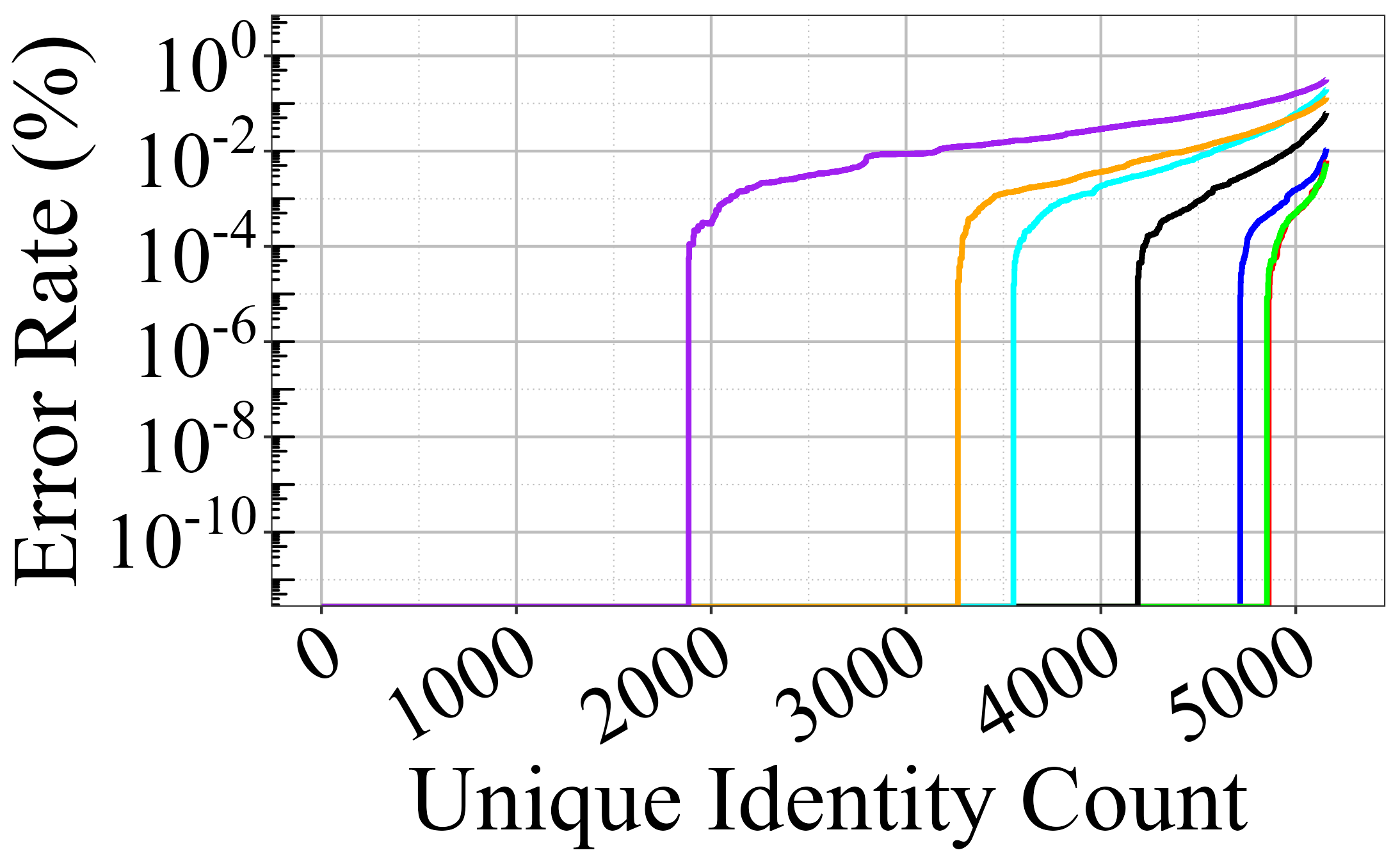}}  
\hfill      
\subcaptionbox{\footnotesize D2: ALLQ\label{fig:D2_SR_ALLQ_0.001}}
    {\includegraphics[width=3.5cm, height=3cm,keepaspectratio]{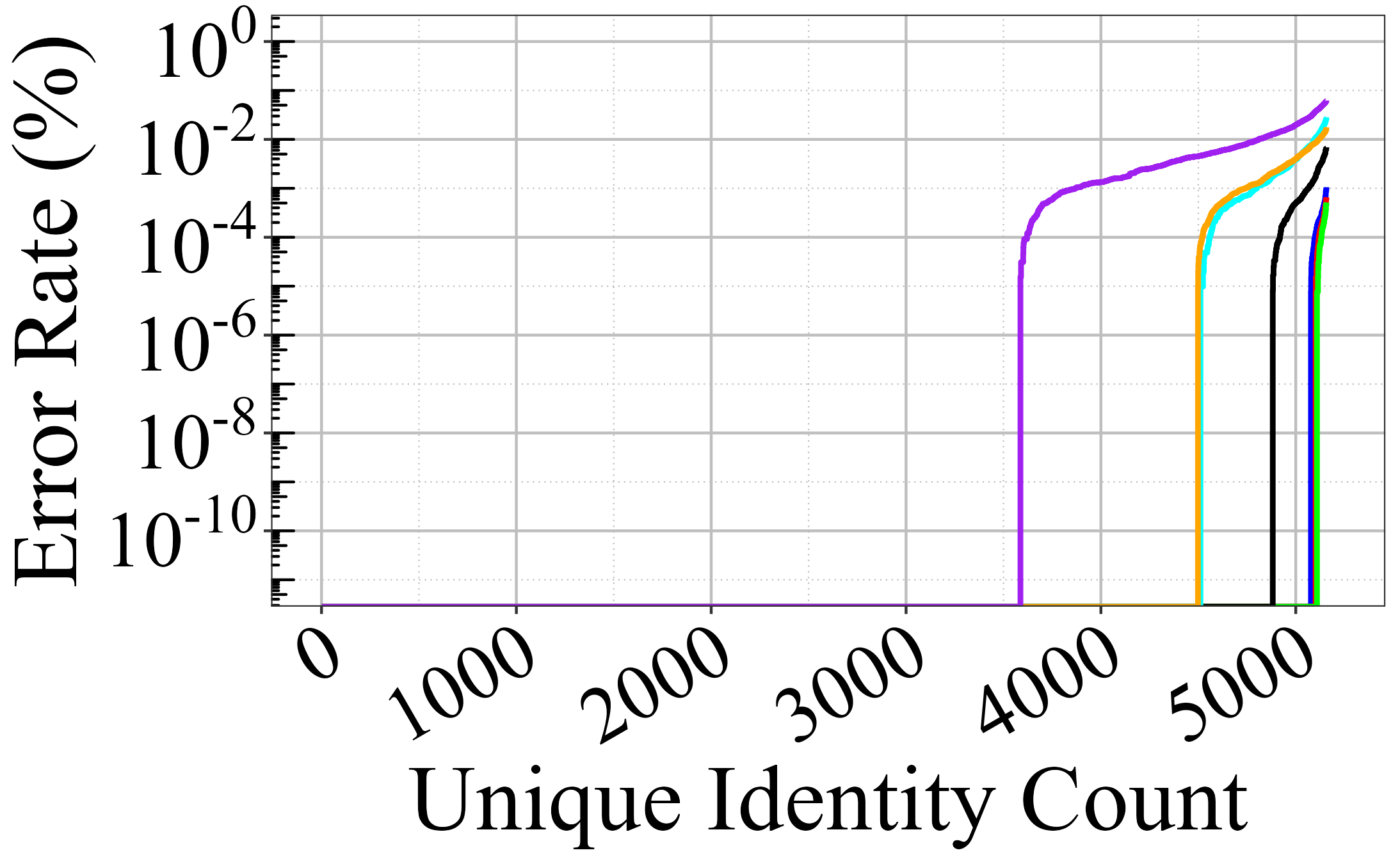}}
\hfill
\par\medskip
\line(1,0){700} 
\par\medskip
\subcaptionbox{\footnotesize D1: ISOQ\label{fig:D1_AR_ISOQ_0.1}}
    {\includegraphics[width=3.5cm, height=3cm,keepaspectratio]{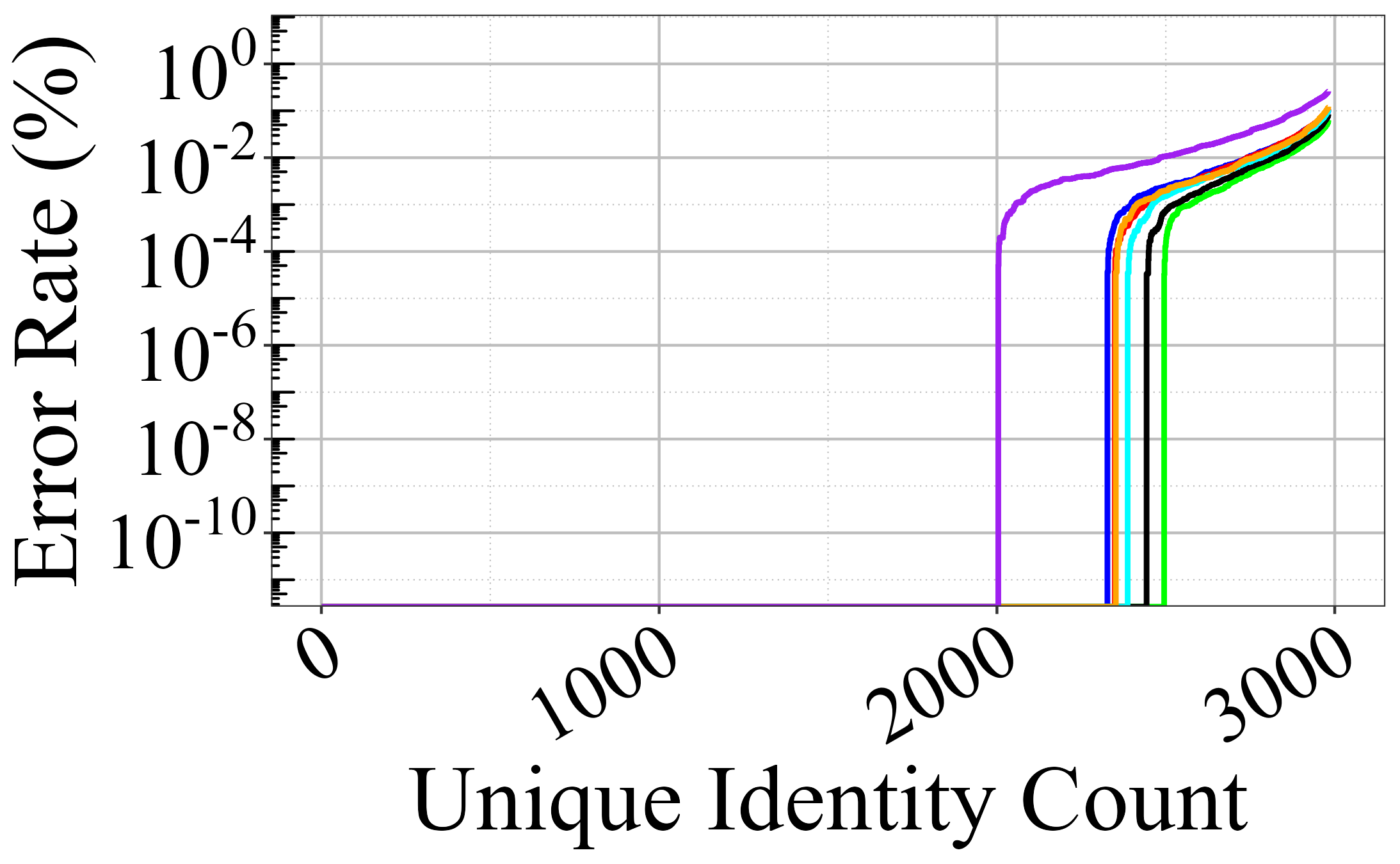}}
\hfill    
\subcaptionbox{\footnotesize D1: ISOQ\label{fig:D1_AR_ISOQ_0.01}}
    {\includegraphics[width=3.5cm, height=3cm,keepaspectratio]{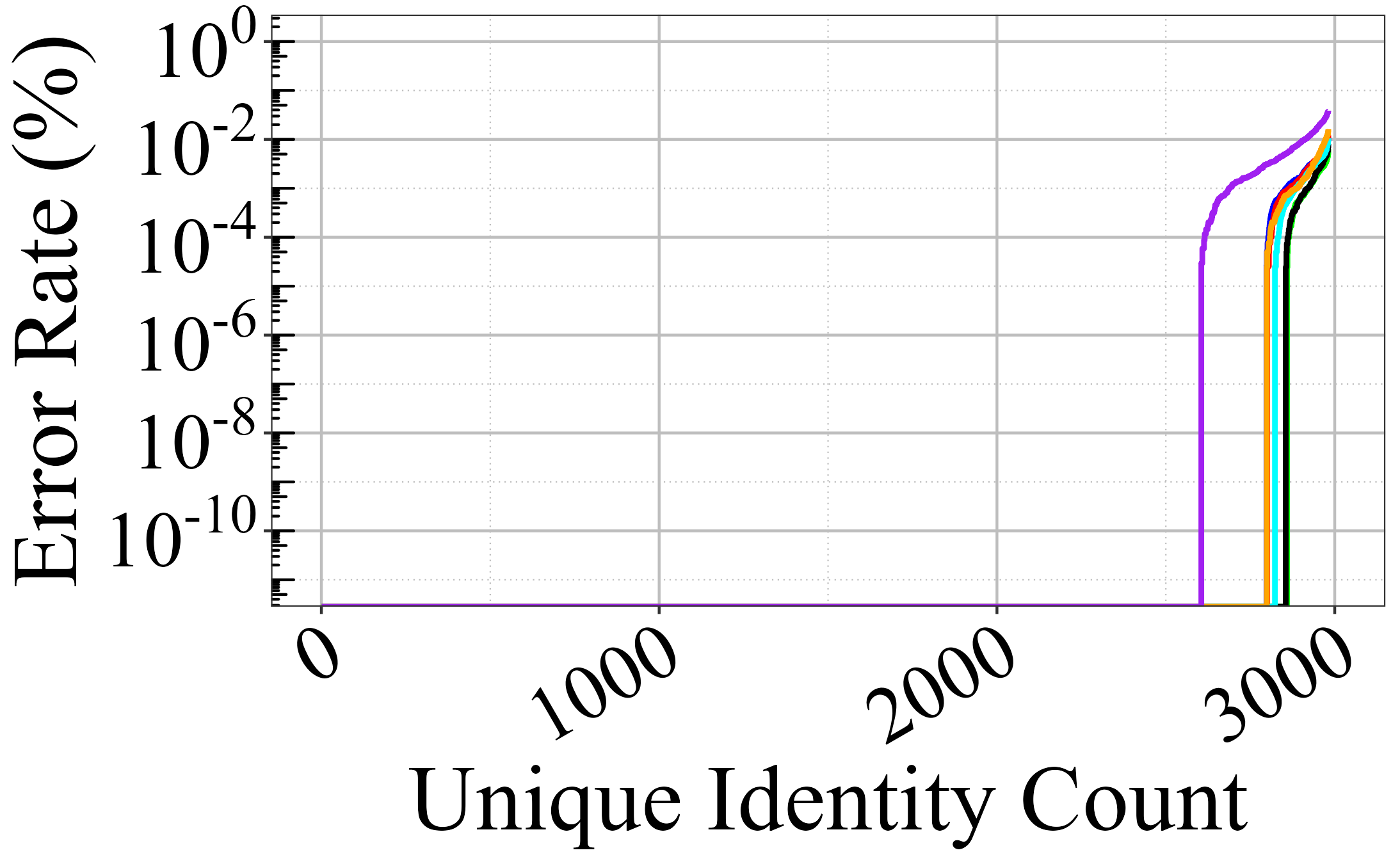}}    
\hfill      
\subcaptionbox{\footnotesize D1: ISOQ\label{fig:D1_AR_ISOQ_0.001}}
    {\includegraphics[width=3.5cm, height=3cm,keepaspectratio]{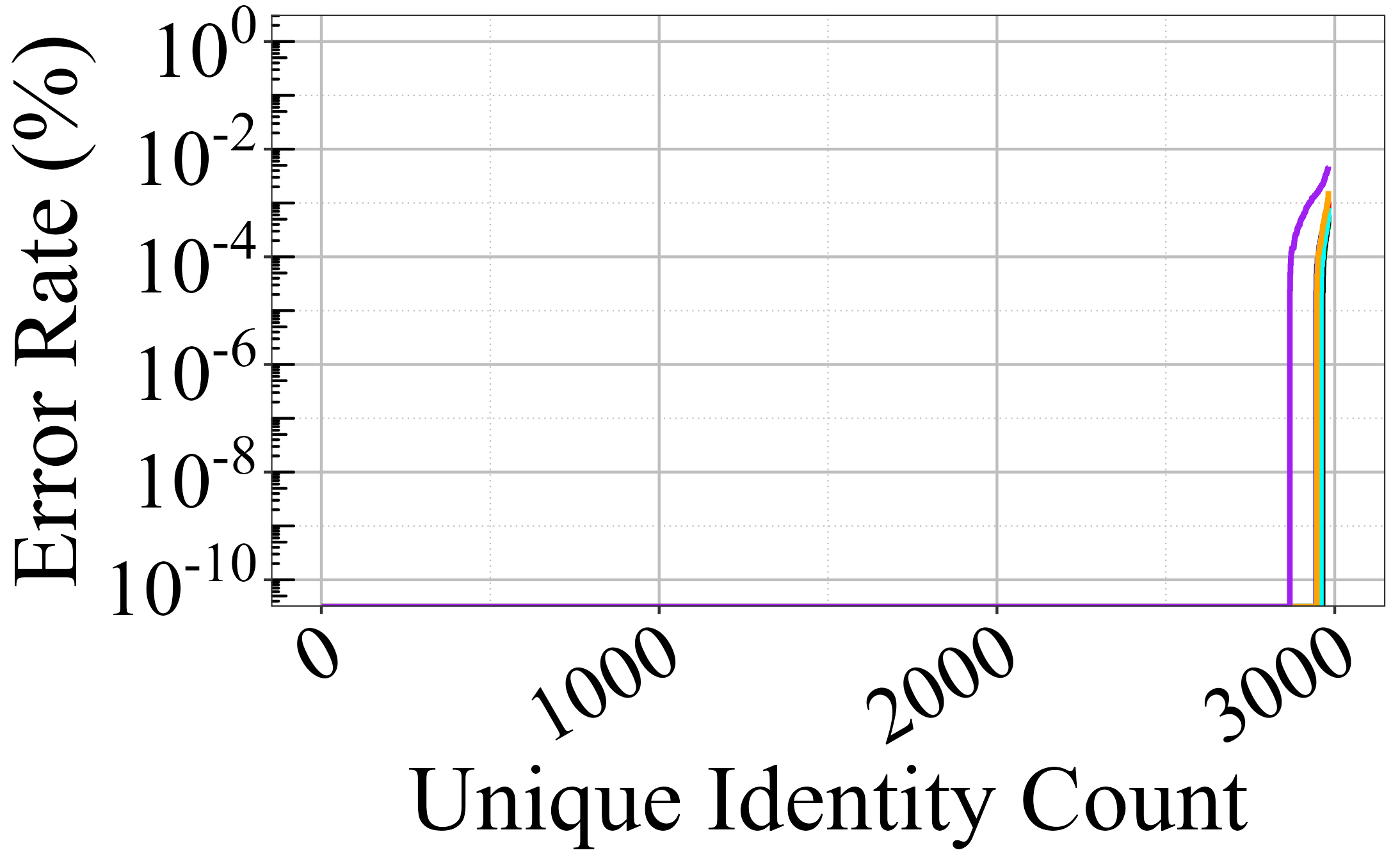}}
\hfill
\vrule
\hspace{1em}
\subcaptionbox{\footnotesize D1: ISOQ\label{fig:D1_SR_ISOQ_0.1}}
    {\includegraphics[width=3.5cm, height=3cm,keepaspectratio]{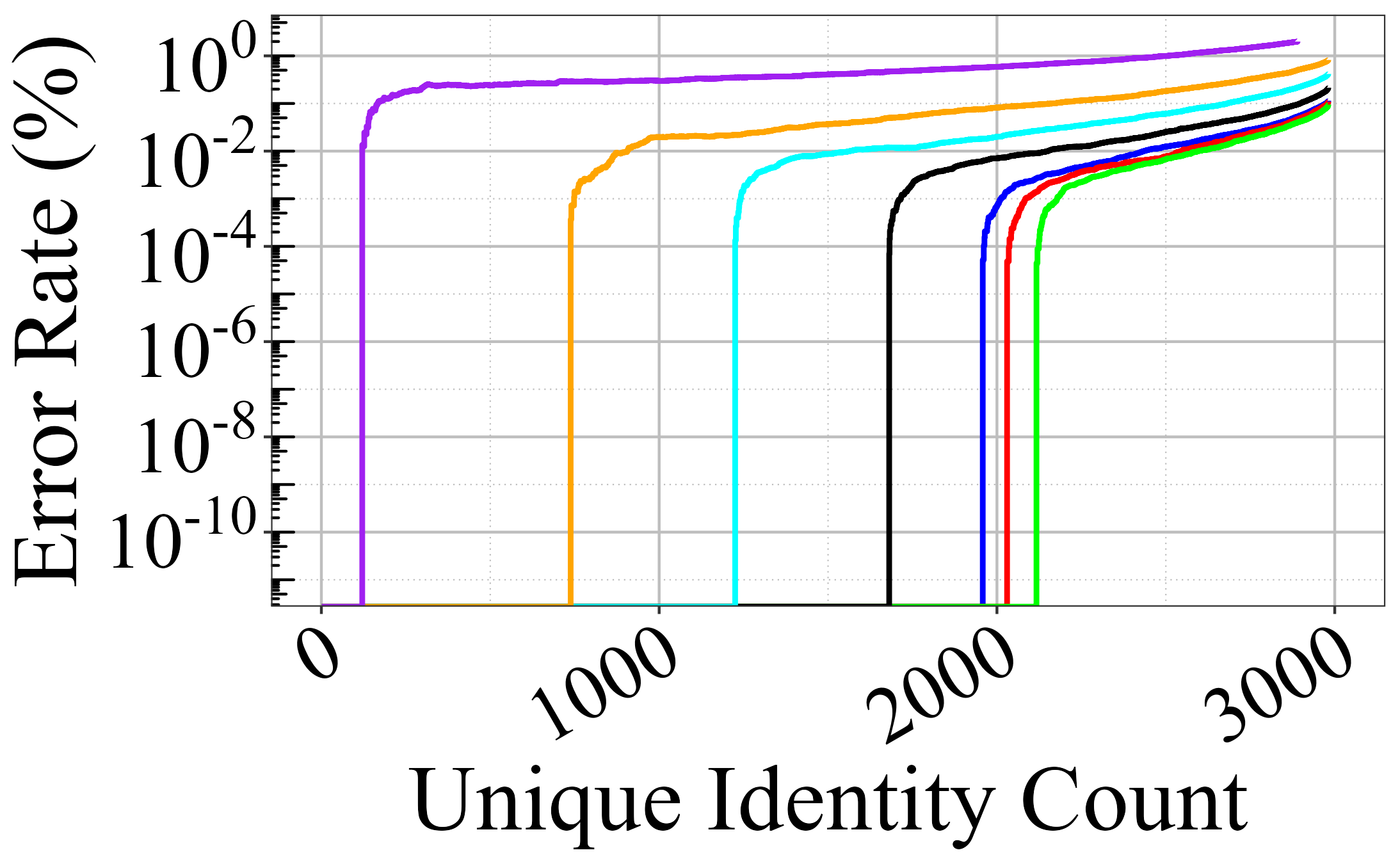}}
\hfill        
\subcaptionbox{\footnotesize D1: ISOQ\label{fig:D1_SR_ISOQ_0.01}}
    {\includegraphics[width=3.5cm, height=3cm,keepaspectratio]{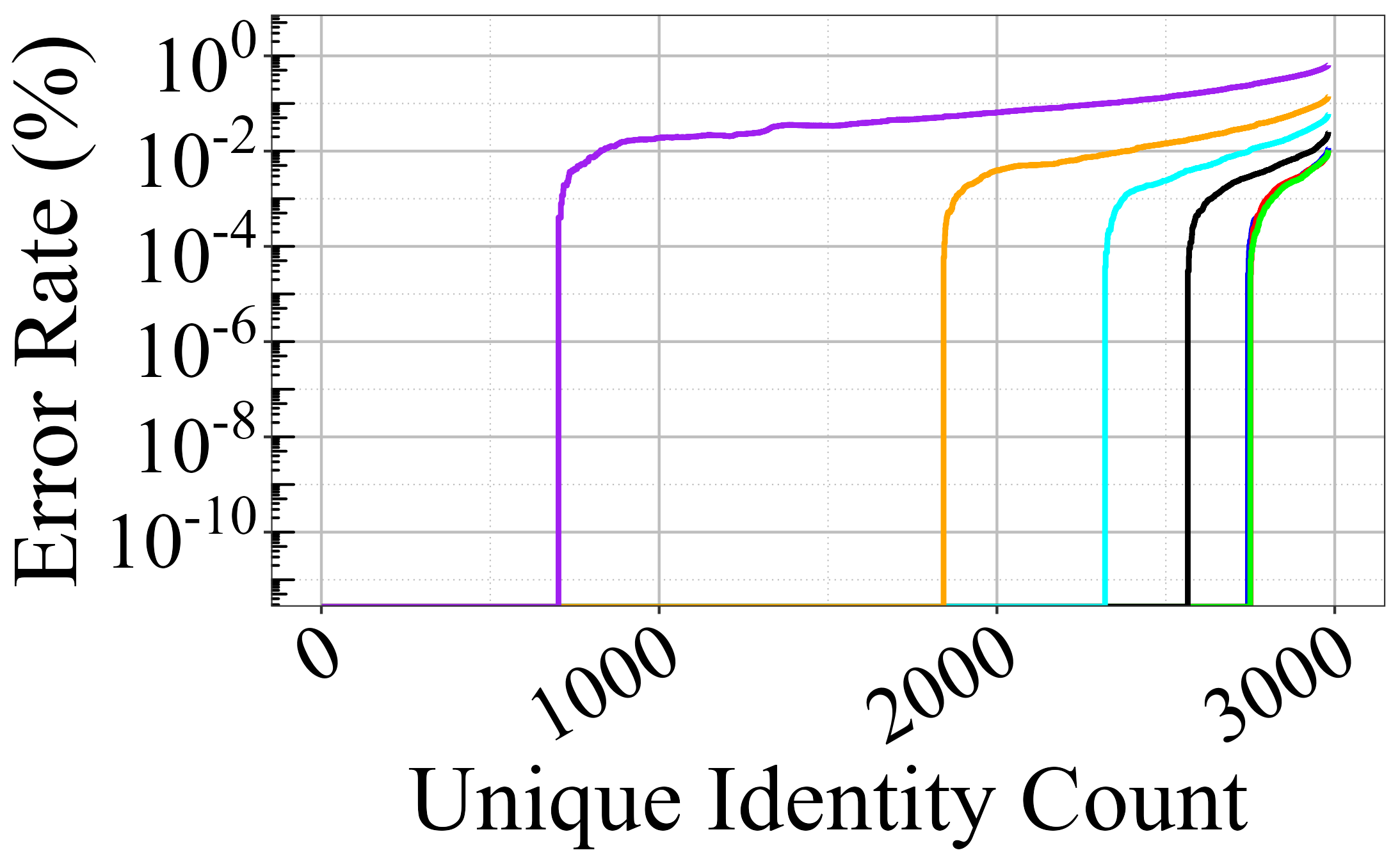}}  
\hfill      
\subcaptionbox{\footnotesize D1:ISOQ\label{fig:D1_SR_ISOQ_0.001}}
    {\includegraphics[width=3.5cm, height=3cm,keepaspectratio]{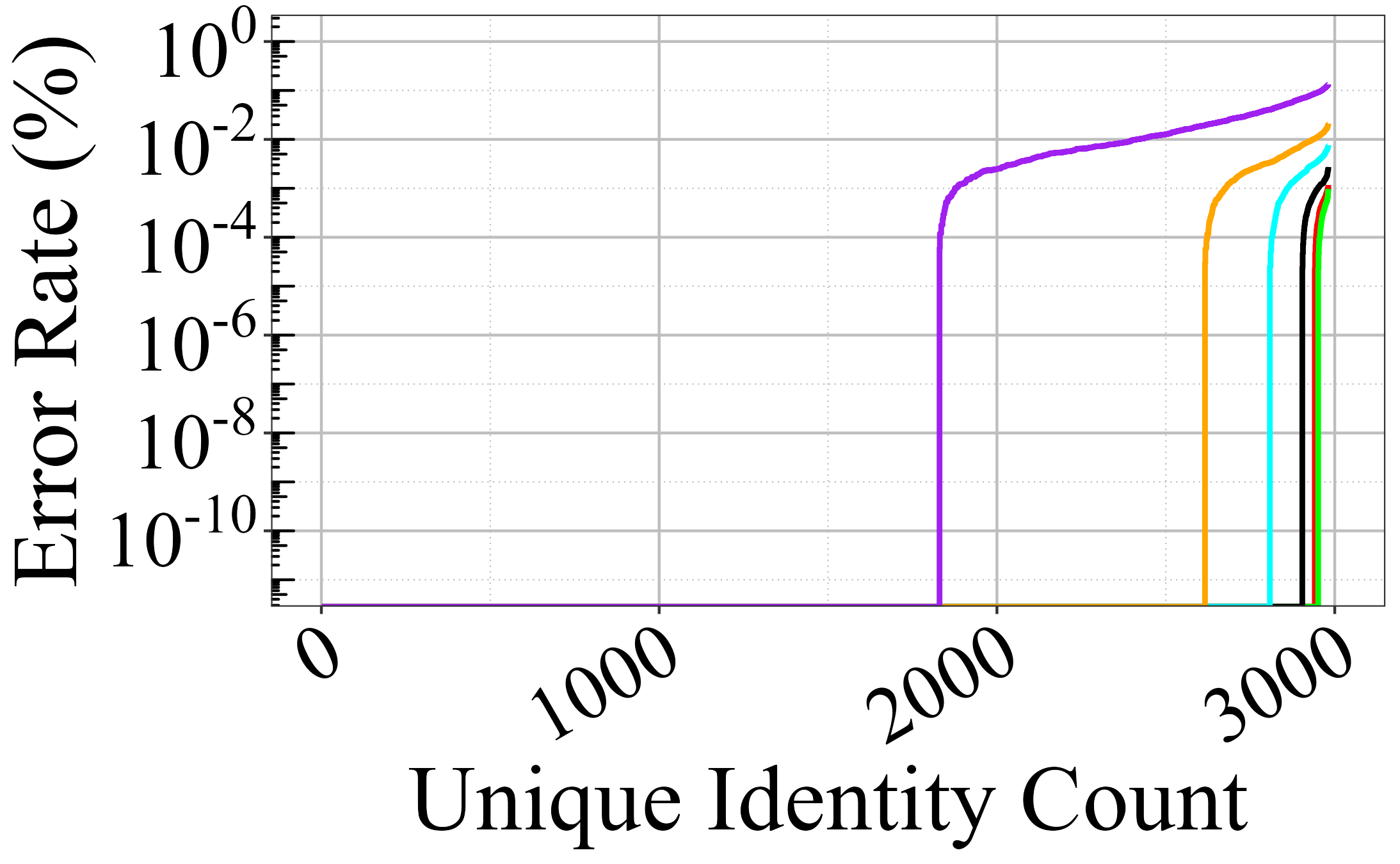}}
 \hfill  
 \par\bigskip
\subcaptionbox{\footnotesize D2: ISOQ\label{fig:D2_AR_ISOQ_0.1}}
    {\includegraphics[width=3.5cm, height=3cm,keepaspectratio]{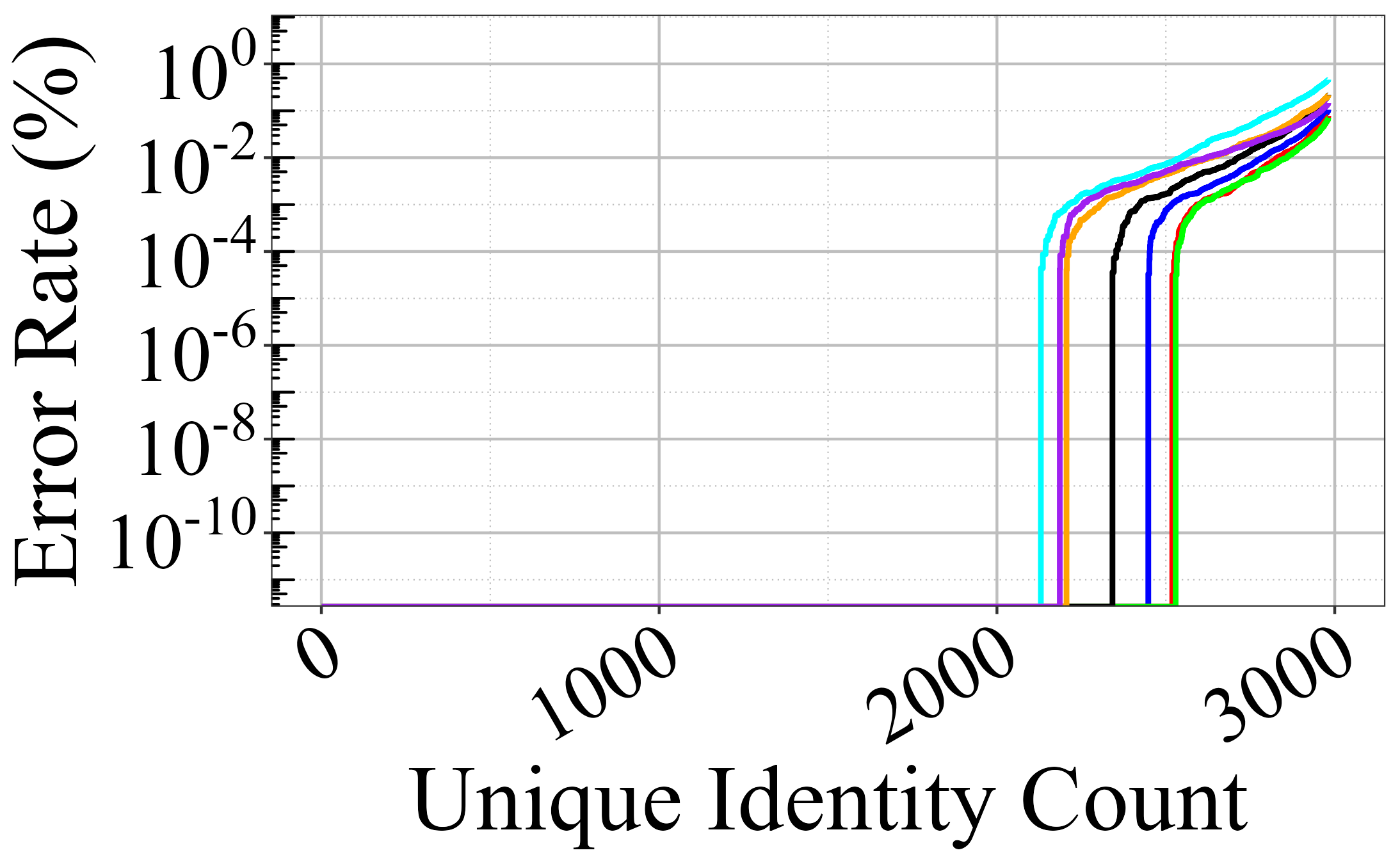}}
\hfill    
\subcaptionbox{\footnotesize D2: ISOQ\label{fig:D2_AR_ISOQ_0.01}}
    {\includegraphics[width=3.5cm, height=3cm,keepaspectratio]{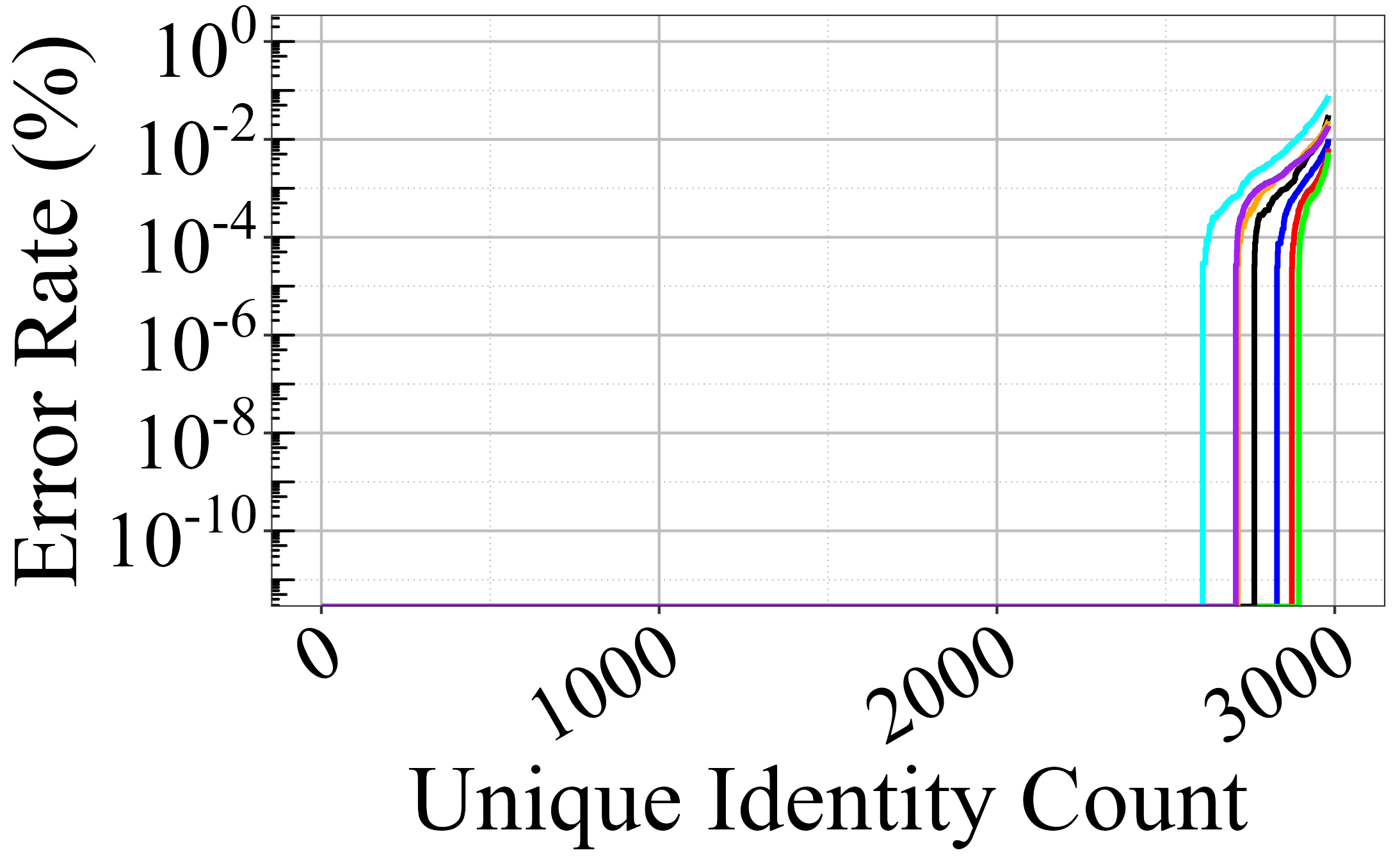}}    
\hfill      
\subcaptionbox{\footnotesize D2: ISOQ\label{fig:D2_AR_ISOQ_0.001}}
    {\includegraphics[width=3.5cm, height=3cm,keepaspectratio]{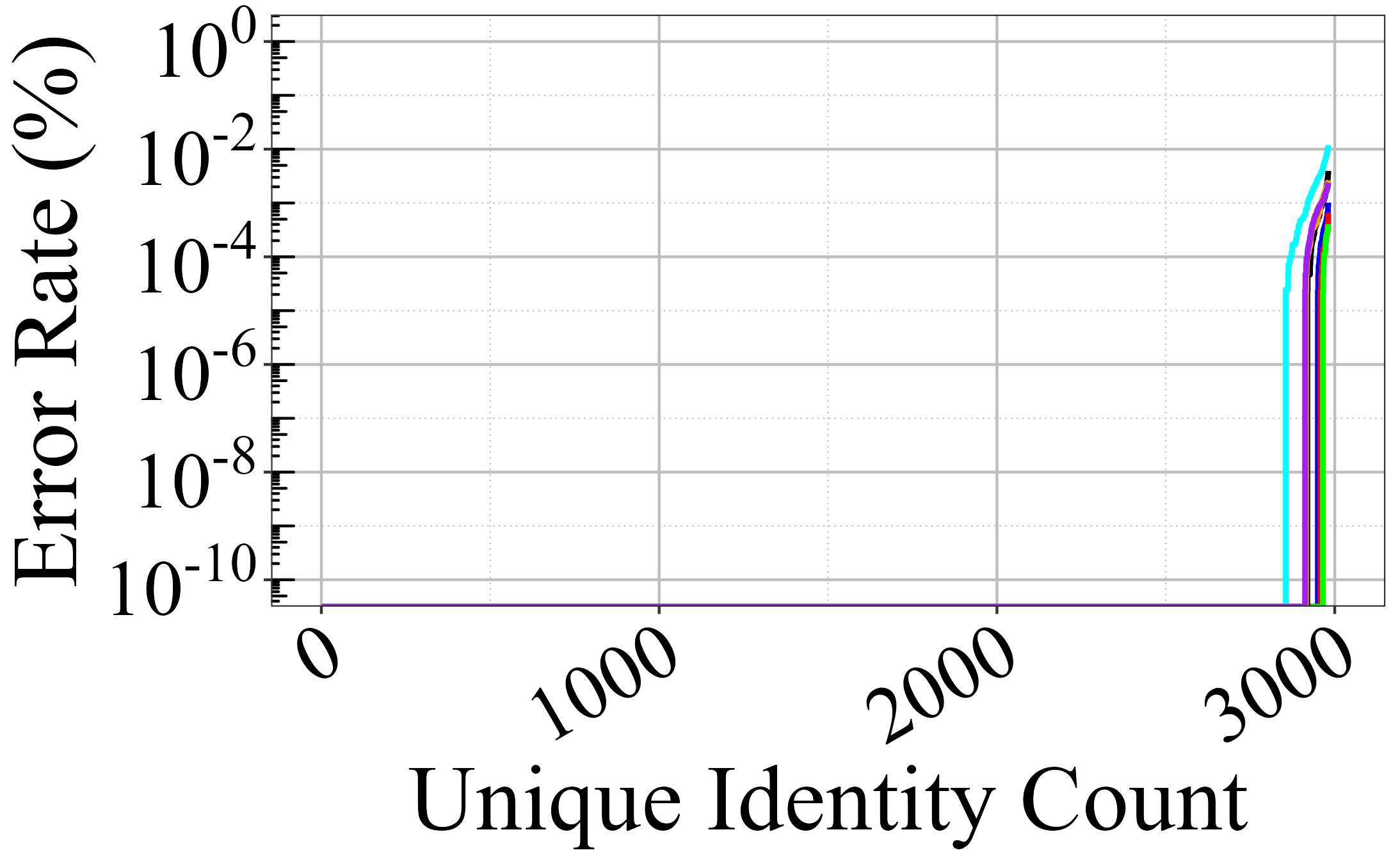}}
\hfill 
\vrule
\hspace{1em}
\subcaptionbox{\footnotesize D2: ISOQ\label{fig:D2_SR_ISOQ_0.1}}
    {\includegraphics[width=3.5cm, height=3cm,keepaspectratio]{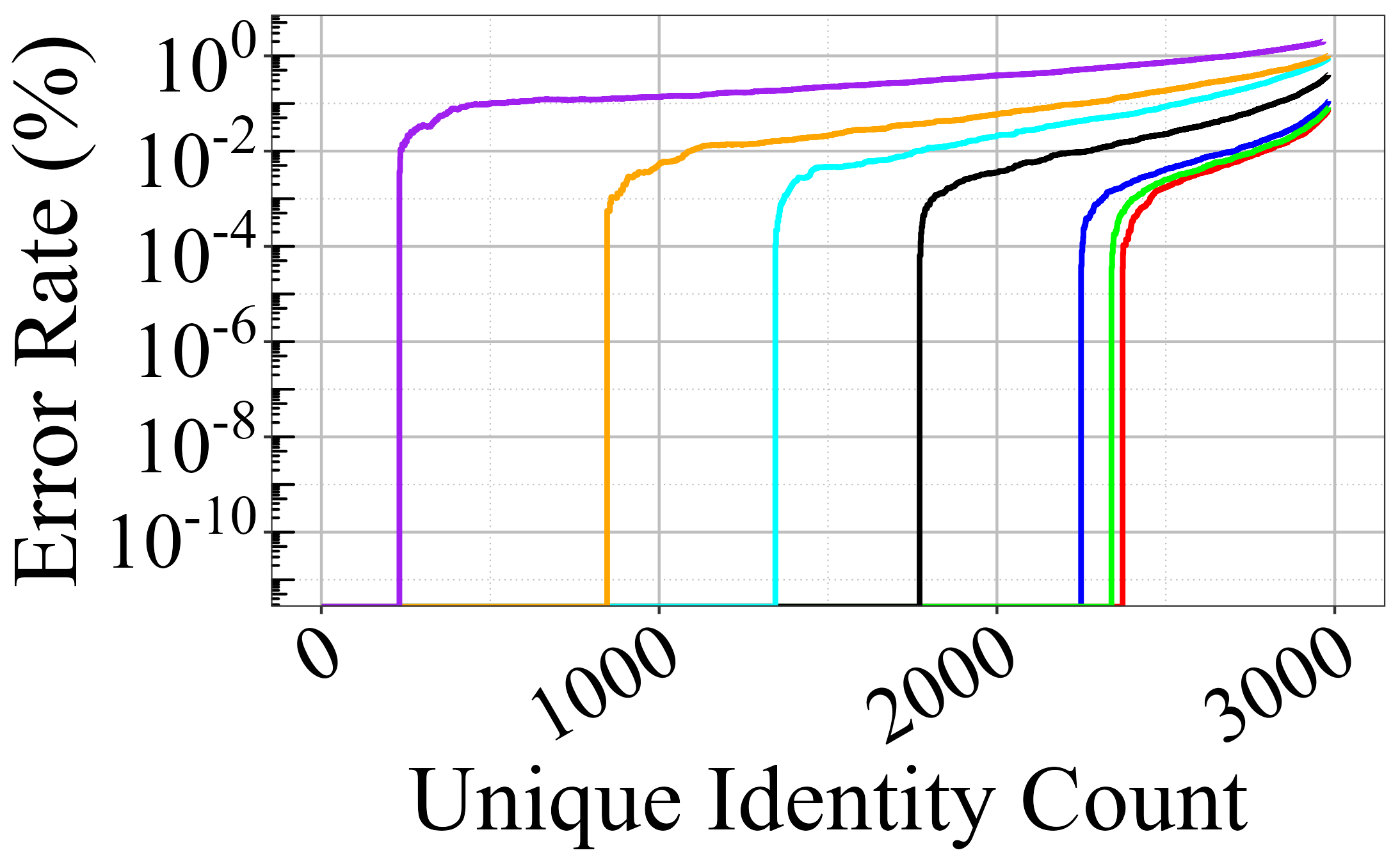}}
\hfill        
\subcaptionbox{\footnotesize D2:ISOQ\label{fig:D2_SR_ISOQ_0.01}}
    {\includegraphics[width=3.5cm, height=3cm,keepaspectratio]{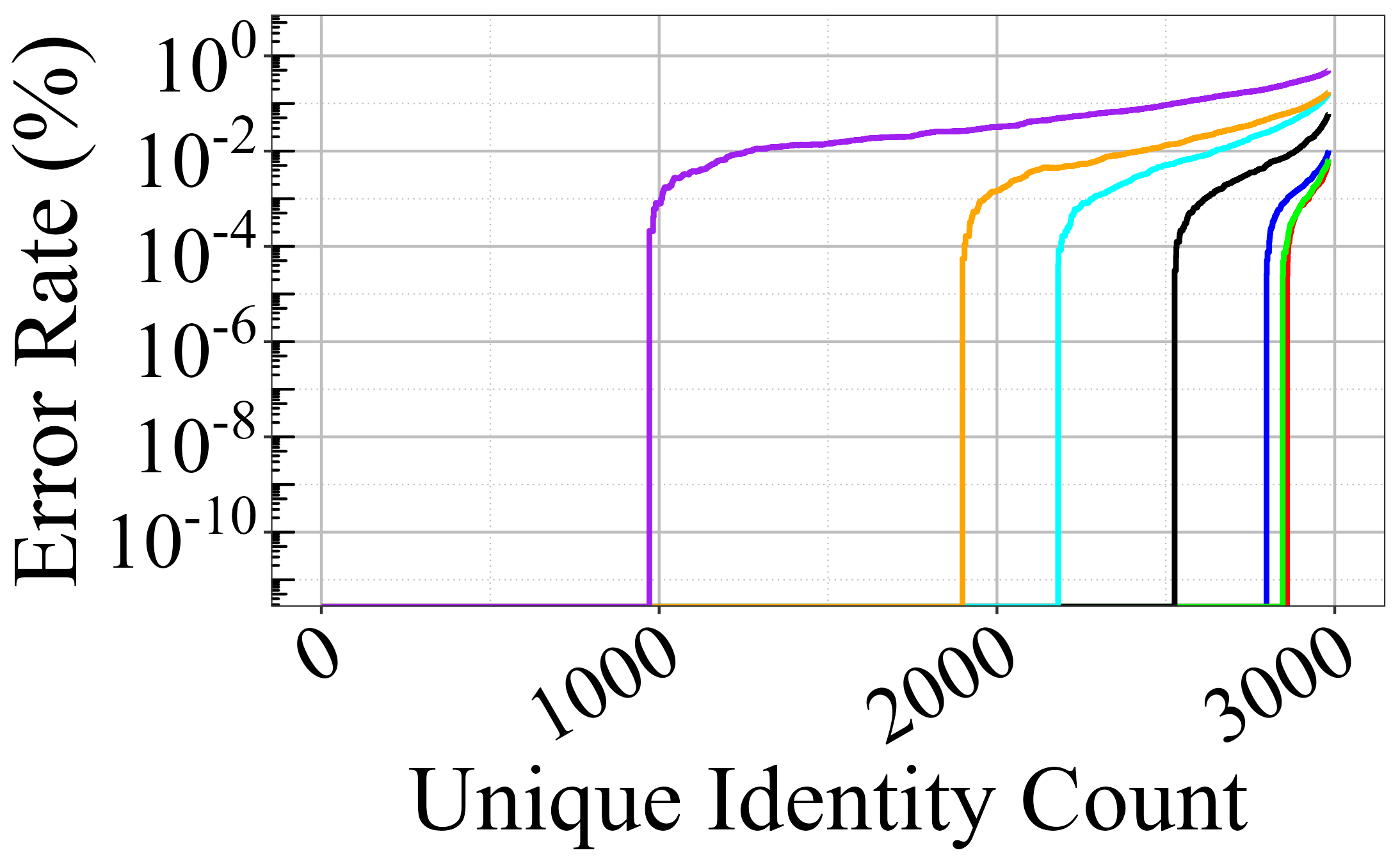}}  
\hfill      
\subcaptionbox{\footnotesize D2: ISOQ\label{fig:D2_SR_ISOQ_0.001}}
    {\includegraphics[width=3.5cm, height=3cm,keepaspectratio]{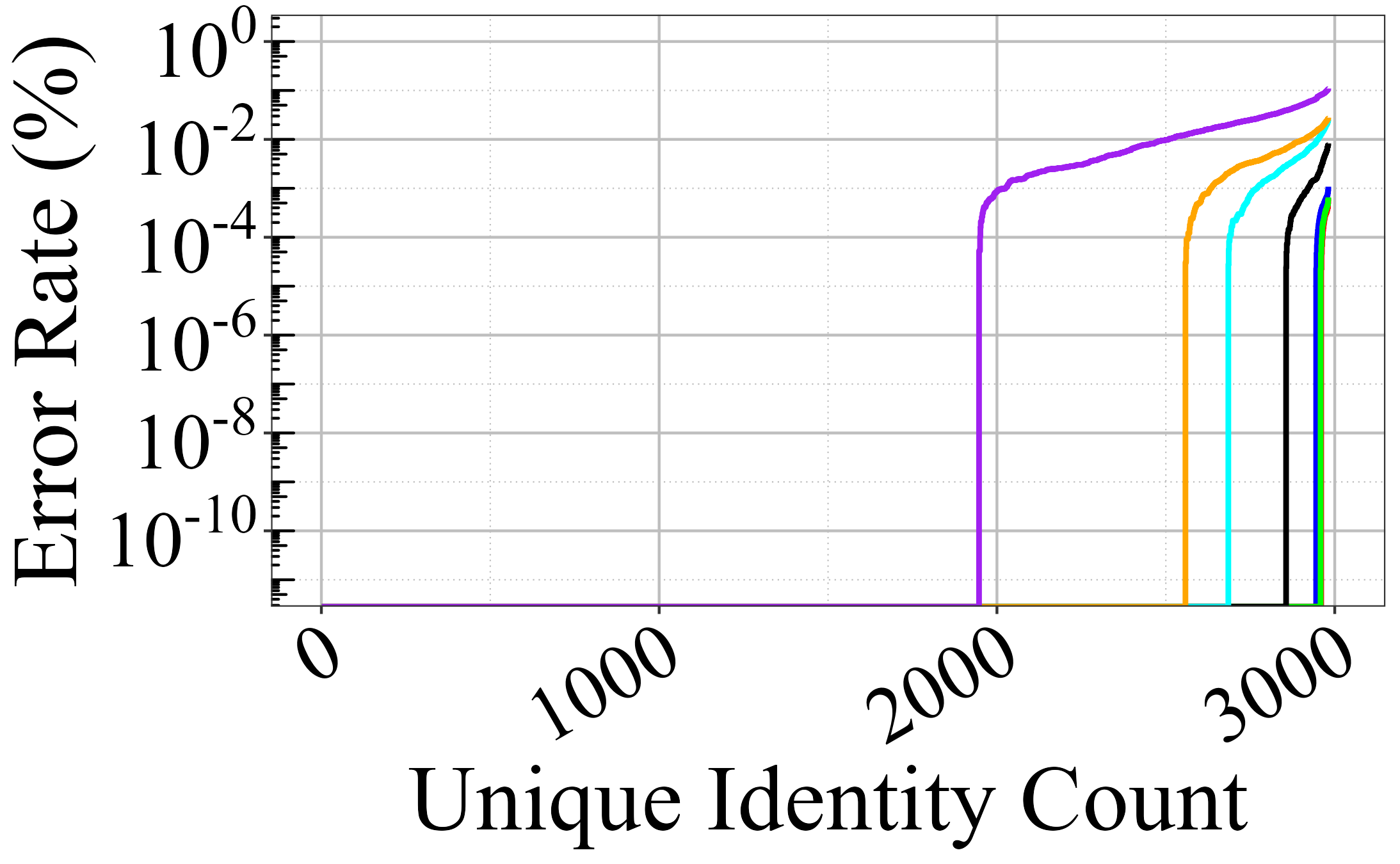}}
\hfill
\par\bigskip
\subcaptionbox*{}
[0.5\textwidth]{\includegraphics[width=10cm, height=3cm,keepaspectratio]{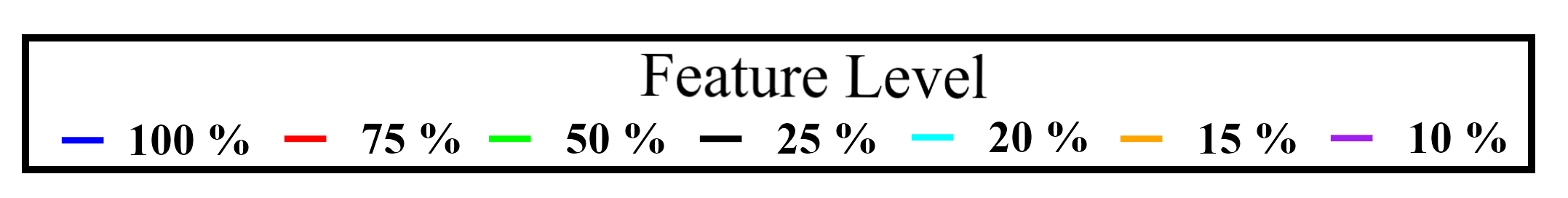}}
\hfill
      \caption{ \footnotesize Error rate as the number of identities increases in the system at different feature levels (100\% to 10\%). Plots show 24 different system configurations for different system configurations - different template dimensions (D1 and D2) generated with different filter resolutions (Single and Multi-resolution filter) functioning at three operating points (0.1\%, 0.01\% and 0.001\% FAR) for different quality datasets - ALLQ and ISOQ. ALLQ dataset consists of 5158 identities and a subset of this dataset forms the ISOQ dataset with 2982 identities. The number of identities the system resolves before encountering the first error is the \textit{constrained capacity} for that system configuration.\\ }
 \label{fig:Error_IdentityCount_FeatureLevel}
\end{sidewaysfigure*}

As the number of identities increases in a system, the probability of identity conflict increases. We study the impact of increased subject count on the system errors for the 24 different system configurations and report on the constrained system capacity. We study how the error in a system varies, for each of the 24 system configurations, as we gradually increase the number of identities, up to 5158 for the ALLQ dataset and up to 2982 for the ISOQ dataset, for 7 different feature levels- 100\% to 10\%. As the number of identities in the ALLQ and ISOQ datasets are different, these two segments are not directly comparable. We introduced \textit{percent capacity} to compare the performance, which is discussed in more detail in the following sections. The relationship between the unique identity count in a system and the corresponding error rate for the 24 different system configurations for 7 different feature levels- 100\% to 10\% is shown in Figure~\ref{fig:Error_IdentityCount_FeatureLevel}. It is important to note that the HD corresponding to 100\% feature level for each of the 24 configurations was considered for all 7 different feature levels. The constrained capacity of the 24 systems are tabulated in Table~\ref{table:ALLQ_FA_NICF} and Table~\ref{table:ISOQ_FA_NICF}. 

With ALLQ data, the best-constrained capacity of 5111 identities out of 5128 identities is obtained for the system configuration- ALLQ, Multi-resolution, D2, at OP 0.001\% FAR (refer Fig~\ref{fig:D2_AR_ALLQ_0.001}) and is achieved with 50\% random features for this system configuration, rendering percent capacity of 99.09\%. Arranging identities in the ascending order of the number of false accepts, the system encountered the first identity clash after resolving 5111 identities. Thus, 47 identities (5158 - 5111) contributed to 64 error counts. The number of identities contributing to the false accepts is referred to in the following sections as NICF.

 A detailed assessment of different variability factors on the system capacity is reported later in this section. We report the following observations from our assessment of the 24 system configurations as we vary the number of identities in the system  and provide further details in the next sections-
 
 \begin{itemize}[leftmargin=*, noitemsep]
     \item System configurations are strongly correlated with the error count as the number of identities increases in the system. We note substantial variability in the constrained system capacity (the number of identities the system can resolve before it starts encountering errors at a particular OP) for different structures.
     
     \item For 19 of the 24 system configurations, considering 50\% of the total bits forming the iris template achieves the highest system capacity. These observations could be indicative of the following phenomenon as a result of the random elimination of radial features:
     \begin{itemize}
     \item Retainment of stable bits and subsequent elimination of ``inconsistent bits" \cite{4586380},
     leading to optimized feature template.
     
     \item Elimination of radially correlated bits which do not append to the discriminating ``unique" information \cite{daugman2015information}
     
     \item A combination of both the above two phenomenon
     
     \end{itemize}
     The best performance is recorded for resisting identity conflict for 99.43\% of the dataset considering 50\% of the bits of a $26116 * 3$ bits multi-resolution template (D2) at 0.001\% FAR with ISO quality data; the best performance could be because the \textit{D2 - multi resolution} combination contains the highest number of bits. 
     
     \item Different system structures contribute to system error count from as low as 18 errors (refer Table~\ref{table:ISOQ_FA_NICF}: ISOQ, Multi-resolution, D2, 0.001\% OP with 50\% Features) to as high as 2.7 million errors out of 13.2 million comparisons (refer Table~\ref{table:ALLQ_FA_NICF}: ALLQ, Single-resolution, D1, 0.1\% OP with 10\% Features) 
     
 \end{itemize}

\subsection{Capacity Assessment: Impact of Quality}
Image quality is an instrumental factor in iris recognition performance. The impact of image quality score on false accept rate (FAR) and false reject rate (FRR) has been analyzed in IREX-I \cite{grotherirex1}. We assessed two sets of data- ALLQ as defined in section~\ref{sec:Quality} and a subset of ALLQ, which meets the minimum ISO image quality standards, ISOQ, to understand the impact of quality on constrained system capacity. Table~\ref{table:FAR_FRR} reports the FRR at the corresponding FAR for 24 different system configurations. If we compare the 12 system configurations which operate with ALLQ data with the other 12 system configurations that operate with ISOQ data, we note substantial improvement in FRR by double digits with ISOQ data. We should be mindful of the impact of the chosen operating point of our system on the FRR which heavily impacts the performance of the system. \par Table~\ref{table:ALLQ_FA_NICF} and Table~\ref{table:ISOQ_FA_NICF} report the capacity of systems operating with ALLQ and ISOQ data respectively. The thresholds for each OP are selected  with templates having 100\% features. \textit{It is important to note that our definition of capacity is in terms of subject count, i.e. the number of identities a system can correctly resolve at an acceptable error rate. However, in our study, the ISOQ dataset is a subset of the ALLQ dataset that meets the quality standards. Thus, the subject count in the two quality datasets is different and thus we cannot compare the capacity for these two segments in terms of the number of identities. For the sake of comparing the two segments, we introduced percent capacity - constrained capacity represented in terms of the percentage of the number of identities in the system.} We note an increase in percent capacity for systems operating with ISOQ data by approximately 1\% to 10\% compared to systems operating with ALLQ data. For example, with D2 single resolution template, the ALLQ dataset performs with an FRR of 5.97 \% at 0.1\% FAR and the ISOQ dataset performs with 1.76 \% FRR at 0.001\% FAR. Thus, with the ISOQ dataset, a system can operate at a stricter OP (0.001\%) without compromising FRR and achieve a higher percent capacity of 98.7\% as opposed to 73.3\% with the ALLQ dataset operating at 0.1\% OP. \par
We conclude from our analysis that using images which meet the basic criteria set by ISO\cite{iris_standard_report} positively impacts system capacity. Additionally, ISOQ data impact FRR significantly. Thus, ISOQ dataset will allow a system to operate at a stricter operating point (e.g. 0.001\% FAR) to achieve a higher system capacity without a significant trade-off in FRR.  \par

\begin{table*}[!ht]
\scriptsize
\centering
\caption{\footnotesize ALLQ: Dataset: 5158 Unique Identities; Constrained Capacity at varying resolutions (multi and single resolution), template dimension (D1, D2), feature dimension (100\% to 10\%) and operating points (0.1, 0.01 and 0.001\% verification FAR) }
\label{table:ALLQ_FA_NICF}
\begin{tabular}{|c|c|c||c|c|c|c|c||c|c|c|c|c|}
\hline

 \multirow{2}{*}{\textbf{Template Dimension}} &  \textbf{Feature}  &
 \multirow{2}{*}{\textbf{OP}}&
 \multicolumn{5}{c|}{\textbf{ Multi Resolution }} & \multicolumn{5}{c|}{\textbf{Single Resolution }} \\
\cline{4-13}
 & (\%) & (\%) &
\textbf{HD} & 
 \textbf{FA} &
 \textbf{FAR (\%)} &
 \textbf{CC} &
 \textbf{PC} &
\textbf{HD} & 
 \textbf{FA} &
 \textbf{FAR (\%)} &
 \textbf{CC} &
 \textbf{PC}\\
 
\hline
\hline

\multirow{21}{*}{\textbf{D1}} & 100	 & \multirow{7}{*}{0.1}	 & \multirow{7}{*}{0.385}	 & 13460	 & 0.101	 & 3868	 & 74.9\% & \multirow{7}{*}{0.418}	 & 14071	 & 0.106	 & 3550 & 68.8\%\\
\cline{2-2}\cline{5-8}\cline{10-13}
 & 75	 & 	 & 	 & 15766	 & 0.119	 & 3854 & 74.7\% 	 & 	 & 15461	 & 0.116 & 3525 & 68.3\%\\
\cline{2-2}\cline{5-8}\cline{10-13}
 & 50	 & 	 & 	 & 7484	 & 0.056	 & \textbf{4151} & 80.4\%	 & 	 & 10188	 & 0.077	 & \textbf{3669} &71.1\%\\
\cline{2-2}\cline{5-8}\cline{10-13}
 & 25	 & 	 & 	 & 12290	 & 0.092	 & 3984 &77.2\%	 & 	 & 25311	 & 0.19	 & 2952 & 57.2\%\\
\cline{2-2}\cline{5-8}\cline{10-13}
 & 20	 & 	 & 	 & 16946	 & 0.127	 & 3836 & 74.3\%	 & 	 & 49170	 & 0.37	 & 2250 &43.6\%\\
\cline{2-2}\cline{5-8}\cline{10-13}
 & 15	 & 	 & 	 & 20012	 & 0.15	 & 3780 & 73.2\%	 & 	 & 85241	 & 0.641	 & 1412 &27.3\%\\
\cline{2-2}\cline{5-8}\cline{10-13}
 & 10	 & 	 & 	 & 36807	 & 0.277	 & 3151 & 61.0\%	 & 	 & 274738	 & 2.066	 & 258 & 5\%\\
 \cline{2-13}
  \cline{2-13}
 & 100	 & \multirow{7}{*}{0.01}	 &  \multirow{7}{*}{0.368}	 & 1458	 & 0.011	 & 4748 & 92.0\%	 &   \multirow{7}{*}{0.404}	 & 1483	 & 0.011	 & 4702 & 91.1\%\\
\cline{2-2}\cline{5-8}\cline{10-13}
 & 75	 & 	 & 	 & 1737	 & 0.013	 & 4740 & 91.9\%	 & 	 & 1623	 & 0.012	 & 4687 & 90.8\%\\
\cline{2-2}\cline{5-8}\cline{10-13}
 & 50	 & 	 & 	 & 756	 & 0.006	 & \textbf{4871} & 94.4\%	 & 	 & 947	 & 0.007	 & \textbf{4793} & 92.9\%\\
\cline{2-2}\cline{5-8}\cline{10-13}
 & 25	 & 	 & 	 & 1388	 & 0.01	 & 4798 & 93.0\%	 & 	 & 2796	 & 0.021	 & 4496 & 87.1\%\\
\cline{2-2}\cline{5-8}\cline{10-13}
 & 20	 & 	 & 	 & 2056	 & 0.015	 & 4690 & 90.9\%	 & 	 & 5982	 & 0.045	 & 4075 & 79.0\%\\
\cline{2-2}\cline{5-8}\cline{10-13}
 & 15	 & 	 & 	 & 2675	 & 0.02	 & 4661 & 90.3\%	 & 	 & 11970	 & 0.09	 & 3481 & 67.4\%\\
\cline{2-2}\cline{5-8}\cline{10-13}
 & 10	 & 	 & 	 & 5452	 & 0.041	 & 4277 & 82.9\%	 & 	 & 52485	 & 0.395	 & 1724 & 33.4\%\\
 \cline{2-13}
  \cline{2-13} 
 & 100	 & \multirow{7}{*}{0.001}	 & \multirow{7}{*}{0.351}	 & 138	 & 0.001	 & 5075 & 98.3\%	 & \multirow{7}{*}{0.388}	 & 135	 & 0.001	 & 5069 & 98.2\%\\
\cline{2-2}\cline{5-8}\cline{10-13}
 & 75	 & 	 & 	 & 164	 & 0.001	 & 5068 & 98.2\%	 & 	 & 140	 & 0.001	 & 5065 & 98.2\%\\
\cline{2-2}\cline{5-8}\cline{10-13}
 & 50	 & 	 & 	 & 76	 & 0.001	 & \textbf{5105} & 98.9\%	 & 	 & 79	 & 0.001	 & \textbf{5099} & 98.8\%\\
\cline{2-2}\cline{5-8}\cline{10-13}
 & 25	 & 	 & 	 & 164	 & 0.001	 & 5069 & 98.2\%	 & 	 & 236	 & 0.002	 & 5031 & 97.5\%\\
\cline{2-2}\cline{5-8}\cline{10-13}
 & 20	 & 	 & 	 & 236	 & 0.002	 & 5032 & 97.5\%	 & 	 & 544	 & 0.004	 & 4904 & 95.0\%\\
\cline{2-2}\cline{5-8}\cline{10-13}
 & 15	 & 	 & 	 & 330	 & 0.002	 & 5017 & 92.2\%	 & 	 & 1197	 & 0.009	 & 4723 & 91.5\%\\
\cline{2-2}\cline{5-8}\cline{10-13}
 & 10	 & 	 & 	 & 770	 & 0.006	 & 4889 & 94.7\%	 & 	 & 7593	 & 0.057	 & 3756 & 72.8\%\\
 \hline
\hline

 \multirow{21}{*}{\textbf{D2}} & 100	 & \multirow{7}{*}{0.1}	 & \multirow{7}{*}{0.331}	 & 13502	 & 0.102	 & 3938 & 74.9\%	 & \multirow{7}{*}{0.382}	 & 14664	 & 0.11	 & 3782 & 73.3\%\\
\cline{2-2}\cline{5-8}\cline{10-13}
 & 75	 & 	 & 	 & 9057	 & 0.068	 & 4117 & 79.8\%	 & 	 & 9563	 & 0.072	 & 4008 & 77.7\%\\
\cline{2-2}\cline{5-8}\cline{10-13}
 & 50	 & 	 & 	 & 6968	 & 0.052	 & \textbf{4234} & 82.0\%	 & 	 & 8265	 & 0.062	 & \textbf{3924} & 76.0\%\\
\cline{2-2}\cline{5-8}\cline{10-13}
 & 25	 & 	 & 	 & 33236	 & 0.25	 & 3712 & 71.9\%	 & 	 & 55789	 & 0.419	 & 2909 & 56.4\%\\
\cline{2-2}\cline{5-8}\cline{10-13}
 & 20	 & 	 & 	 & 80554	 & 0.606	 & 3228 & 62.5\%	 & 	 & 136204	 & 1.024	 & 2170 & 42.0\%\\
\cline{2-2}\cline{5-8}\cline{10-13}
 & 15	 & 	 & 	 & 30399	 & 0.229	 & 3438 & 66.6\%	 & 	 & 106346	 & 0.8	 & 1545 & 29.9\%\\
\cline{2-2}\cline{5-8}\cline{10-13}
 & 10	 & 	 & 	 & 14803	 & 0.111	 & 3502 & 67.8\%	 & 	 & 194393	 & 1.462	 & 485 & 9.4\%\\
\cline{2-13}
 \cline{2-13}
 & 100	 & \multirow{7}{*}{0.01}	 & \multirow{7}{*}{0.31}	 & 1421	 & 0.011	 & 4779 & 92.6\%	 & \multirow{7}{*}{0.363}	 & 1482	 & 0.011	 & 4714 & 91.3\%\\
\cline{2-2}\cline{5-8}\cline{10-13}
 & 75	 & 	 & 	 & 866	 & 0.007	 & 4863 & 94.2\%	 & 	 & 843	 & 0.006	 & \textbf{4860} & 94.2\%\\
\cline{2-2}\cline{5-8}\cline{10-13}
 & 50	 & 	 & 	 & 654	 & 0.005	 & \textbf{4926} & 95.5\%	 & 	 & 730	 & 0.005	 & 4851 & 94.0\%\\
\cline{2-2}\cline{5-8}\cline{10-13}
 & 25	 & 	 & 	 & 5130	 & 0.039	 & 4512 & 87.4\%	 & 	 & 8342	 & 0.063	 & 4175 & 80.9\%\\
\cline{2-2}\cline{5-8}\cline{10-13}
 & 20	 & 	 & 	 & 15827	 & 0.119	 & 4151 & 80.4\%	 & 	 & 26399	 & 0.198	 & 3549 & 68.8\%\\
\cline{2-2}\cline{5-8}\cline{10-13}
 & 15	 & 	 & 	 & 4556	 & 0.034	 & 4432 & 85.9\%	 & 	 & 17680	 & 0.133	 & 3549 & 63.0\%\\
\cline{2-2}\cline{5-8}\cline{10-13}
 & 10	 & 	 & 	 & 2148		 & 0.016	 & 4537 & 87.9\%	 & 	 & 42261	 & 0.318	 & 1880 & 36.4\%\\
 \cline{2-13}
  \cline{2-13}
 & 100	 & \multirow{7}{*}{0.001}	 & \multirow{7}{*}{0.291}	 & 134	 & 0.001	 & 5074 & 98.3\%	 & \multirow{7}{*}{0.343}	 & 140	 & 0.001	 & 5077 & 98.4\%\\
\cline{2-2}\cline{5-8}\cline{10-13}
 & 75	 & 	 & 	 & 85	 & 0.001	 & 5098 & 98.8\%	 & 	 & 87	 & 0.001	 & 5101 & 98.8\%\\
\cline{2-2}\cline{5-8}\cline{10-13}
 & 50	 & 	 & 	 & 64	 & 0	 & \textbf{5111} & 99.0\%	 & 	 & 68	 & 0.001	 & \textbf{5108} & 99.0\%\\
\cline{2-2}\cline{5-8}\cline{10-13}
 & 25	 & 	 & 	 & 702	 & 0.005	 & 4950 & 95.9\%	 & 	 & 916	 & 0.007	 & 4881 & 94.6\%\\
\cline{2-2}\cline{5-8}\cline{10-13}
 & 20	 & 	 & 	 & 2599	 & 0.02	 & 4728 & 91.6\%	 & 	 & 3743	 & 0.028	 & 4510 & 87.4\%\\
\cline{2-2}\cline{5-8}\cline{10-13}
 & 15	 & 	 & 	 & 648	 & 0.005	 & 4922 & 95.4\%	 & 	 & 2344	 & 0.018	 & 4493 & 87.1\%\\
\cline{2-2}\cline{5-8}\cline{10-13}
 & 10	 & 	 & 	 & 309	 & 0.002	 & 4990 & 96.7\%	 & 	 & 8292	 & 0.062	 & 3581 & 69.4\%\\
 
\hline

\multicolumn{13}{c}{CC =Constrained Capacity; PC = Percent Capacity; OP**= Operating Point of 0.1 is where the threshold is selected for 0.1\% FAR with 100\% features}\\

\end{tabular}
\vspace{-4mm}
\end{table*}

\begin{table*}[!ht]
\scriptsize
\centering

\caption{\footnotesize ISOQ: Dataset: 2982 Unique Identities; Constrained Capacity at varying resolutions (multi and single resolution), template dimension (D1, D2), feature dimension (100\% to 10\%) and operating points (0.1, 0.01 and 0.001\% verification FAR) }
\label{table:ISOQ_FA_NICF}
\begin{tabular}{|c|c|c||c|c|c|c|c||c|c|c|c|c|}
\hline

 \multirow{2}{*}{\textbf{Template Dimension}} &  \textbf{Feature} &
 \textbf{OP}&
 \multicolumn{5}{c|}{\textbf{ Multi Resolution }} & \multicolumn{5}{c|}{\textbf{Single Resolution }} \\

\cline{4-13}
 & ( \% ) & ( \% ) &
\textbf{HD} & 
 \textbf{FA} &
 \textbf{FAR (\%)} &
 \textbf{CC} &
 \textbf{PC} &
\textbf{HD} & 
 \textbf{FA} &
 \textbf{FAR (\%)} &
 \textbf{CC} &
 \textbf{PC}\\
 
\hline
\hline

 \multirow{21}{*}{\textbf{D1}}  &  100	  &  \multirow{7}{*}{0.1}	  &  \multirow{7}{*}{0.397}	  &  4520	  &  0.102	  &  2324 & 77.9\%	  &  \multirow{7}{*}{0.427}	  &  5075	  &  0.114	  &  1957 & 65.6\%\\
 \cline{2-2}\cline{5-8}\cline{10-13}
  &  75	  &  	  &  	  &  4867	  &  0.11	  &  2347 & 78.7\%	  &  	  &  4857	  &  0.109	  &  2029 & 68.0\%\\
\cline{2-2}\cline{5-8}\cline{10-13}
  &  50	  &  	  &  	  &  2794	  &  0.063	  & \textbf{2489} & 83.5\%	  &  	  &  4219	  &  0.095	  &  \textbf{2115} & 70.9\%\\  
\cline{2-2}\cline{5-8}\cline{10-13}
  &  25	  &  	  &  	  &  3724	  &  0.084	  &  2441 & 81.8\%	  &  	  &  9526	  &  0.214	  &  1679 & 56.3\%\\
\cline{2-2}\cline{5-8}\cline{10-13}
  &  20	  &  	  &  	  &  4677	  &  0.105	  &  2380 & 79.8\%	  &  	  &  18551	  &  0.417	  &  1224 & 41.0\%\\
\cline{2-2}\cline{5-8}\cline{10-13}
  &  15	  &  	  &  	  &  5439	  &  0.122	  &  2347 & 78.7\%	  &  	  &  36463	  &  0.82	  &  731 & 24.5\%\\
\cline{2-2}\cline{5-8}\cline{10-13}
  &  10	  &  	  &  	  &  11541	  &  0.26	  &  1998 & 67.0\%	  &  	  &  122072	  &  2.746	  &  119 & 3.9\%\\
  \cline{2-13}
      \cline{2-13}
  &  100	  &  \multirow{7}{*}{0.01}	  &  \multirow{7}{*}{0.382}	  &  479	  &  0.011	  &  2796 & 93.7\%	  &  \multirow{7}{*}{0.416}	  &  516	  &  0.012	  &  2739 & 91.8\%\\
\cline{2-2}\cline{5-8}\cline{10-13}
  &  75	  &  	  &  	  &  539	  &  0.012	  &  2799 & 93.8\%	  &  	  &  451	  &  0.01	  &  2749 & 92.2\%\\
\cline{2-2}\cline{5-8}\cline{10-13}
  &  50	  &  	  &  	  &  283	  &  0.006	  &  \textbf{2858} & 95.8\%	  &  	  &  463	  &  0.01	  &  \textbf{2749} & 92.2\%\\
\cline{2-2}\cline{5-8}\cline{10-13}
  &  25	  &  	  &  	  &  357	  &  0.008	  &  2854 & 95.7\%	  &  	  &  1120	  &  0.025	  &  2562 & 85.9\%\\
\cline{2-2}\cline{5-8}\cline{10-13}
  &  20	  &  	  &  	  &  468	  &  0.011	  &  2820 & 94.6\%	  &  	  &  2678	  &  0.06	  &  2310 & 77.4\%\\
\cline{2-2}\cline{5-8}\cline{10-13}
  &  15	  &  	  &  	  &  716	  &  0.016	  &  2798 & 93.8\%	  &  	  &  6259	  &  0.141	  &  1834 & 61.5\%\\
\cline{2-2}\cline{5-8}\cline{10-13}
  &  10	  &  	  &  	  &  1715	  &  0.039	  &  2603 & 87.3\%	  &  	  &  28264	  &  0.636	  &  701 & 23.5\%\\
    \cline{2-13}
    \cline{2-13}
  &  100	  &  \multirow{7}{*}{0.001}	  &  \multirow{7}{*}{0.367}	  &  46	  &  0.001	  &  2943 & 98.7\%  &  \multirow{7}{*}{0.405}	  &  46	  &  0.001	  &  2945 & 98.7\%\\
\cline{2-2}\cline{5-8}\cline{10-13}
  &  75	  &  	  &  	  &  48	  &  0.001	  &  2950 & 98.9\%	  &  	  &  52	  &  0.001	  &  2940 & 98.6\%\\
\cline{2-2}\cline{5-8}\cline{10-13}
  &  50	  &  	  &  	  &  27	  &  0.001	  &  2960 & 99.3\%  &  	  &  43	  &  0.001	  &  \textbf{2950} & 98.9\%\\
\cline{2-2}\cline{5-8}\cline{10-13}
  &  25	  &  	  &  	  &  27	  &  0.001	  &  2962 & 99.3\%	  &  	  &  121	  &  0.003	  &  2903 & 97.3\%\\
\cline{2-2}\cline{5-8}\cline{10-13}
  &  20	  &  	  &  	  &  36	  &  0.001	  & 2958 & 99.2\%	  &  	  &  339	  &  0.008	  &  2807 & 94.1\%\\
\cline{2-2}\cline{5-8}\cline{10-13}
  &  15	  &  	  &  	  &  74	  &  0.002	  &  2945 & 98.7\% 	  &  	  &  933	  &  0.021	  &  2614 & 87.6\%\\
\cline{2-2}\cline{5-8}\cline{10-13}
  &  10	  &  	  &  	  &  210	  &  0.005	  &  2864 & 96.0\%	  &  	  &  5930	  &  0.133	  &  1824 & 61.1\%\\

\hline
\hline

 \multirow{21}{*}{\textbf{D2}}   &  100	  &  \multirow{7}{*}{0.1}	  &  \multirow{7}{*}{0.343}	  &  4680	  &  0.105	  &  2441 & 81.8\%	  &  \multirow{7}{*}{0.395}	  &  4969	  &  0.112	  &  2242 & 75.2\%\\
\cline{2-2}\cline{5-8}\cline{10-13}
  &  75	  &  	  &  	  &  3458	  &  0.078	  &  2517 & 84.4\%	  &  	  &  3347	  &  0.075	  &  2368 & 79.4\%\\
\cline{2-2}\cline{5-8}\cline{10-13}
  &  50	  &  	  &  	  &  3154	  &  0.071	  &  \textbf{2526} & 84.7\%	  &  	  &  3701	  &  0.083	  &  2337 & 78.4\%\\
\cline{2-2}\cline{5-8}\cline{10-13}
  &  25	  &  	  &  	  &  9723	  &  0.219	  &  2340 & 78.5\%	  &  	  &  17769	  &  0.4	  &  1767 & 59.2\%\\
\cline{2-2}\cline{5-8}\cline{10-13}
  &  20	  &  	  &  	  &  20490	  &  0.461	  & 2120 & 71.1\%	  &  	  &  39414	  &  0.887	  &  1340 & 44.9\%\\
\cline{2-2}\cline{5-8}\cline{10-13}
  &  15	  &  	  &  	  &  9524	  &  0.214	  &  2204 & 73.9\%	  &  	  &  44903	  &  1.01	  &  842 & 28.2\%\\
\cline{2-2}\cline{5-8}\cline{10-13}
  &  10	  &  	  &  	  &  6502	  &  0.146	  &  2180 & 73.1\%	  &  	  &  95873	  &  2.157	  &  226 & 7.5\%\\
 \cline{2-13}
  \cline{2-13}
  &  100	  &  \multirow{7}{*}{0.01}	  &  \multirow{7}{*}{0.322}	  &  450	  &  0.01	  &  2828 & 94.8\%	  &  \multirow{7}{*}{0.378}	  &  455	  &  0.01	  &  2797 & 93.8\%\\
\cline{2-2}\cline{5-8}\cline{10-13}
  &  75	  &  	  &  	  &  290	  &  0.007	  &  2870 & 96.2\%	  &  	  &  285	  &  0.006	  &  2856 & 95.8\%\\
\cline{2-2}\cline{5-8}\cline{10-13}
      &  50	  &  	  &  	  &  240	  &  0.005	  &  \textbf{2892} & 97.0\%  &  	  &  297	  &  0.007	  &  2845 & 95.4\%\\
\cline{2-2}\cline{5-8}\cline{10-13}
  &  25	  &  	  &  	  &  1380	  &  0.031	  &  2760 & 92.5\%	  &  	  &  2689	  &  0.06	  &  2523 & 84.6\%\\
\cline{2-2}\cline{5-8}\cline{10-13}
  &  20	  &  	  &  	  &  3450	  &  0.078	  &  2607 & 87.4\%	  &  	  &  7150	  &  0.161	  &  2179 & 73.1\%\\
\cline{2-2}\cline{5-8}\cline{10-13}
  &  15	  &  	  &  	  &  1052	  &  0.024	  &  2712 & 90.9\%	  &  	  &  7658	  &  0.172	  &  1897 & 63.6\%\\
\cline{2-2}\cline{5-8}\cline{10-13}
  &  10	  &  	  &  	  &  834	  &  0.019	  &  2706 & 90.7\%	  &  	  &  21709	  &  0.488	  &  962 & 32.2\%\\
  \cline{2-13}
  \cline{2-13}
  &  100	  &  \multirow{7}{*}{0.001}	  &  \multirow{7}{*}{0.303}	  &  45	  &  0.001	  &  2949 & 98.9\%	  &  \multirow{7}{*}{0.362}	  &  48	  &  0.001	  &  2944 & 98.7\%\\
\cline{2-2}\cline{5-8}\cline{10-13}
  &  75	  &  	  &  	  &  29	  &  0.001	  &  2957 & 99.1\%	  &  	  &  28	  &  0.001	  &  2959 & 99.2\%\\
\cline{2-2}\cline{5-8}\cline{10-13}
  &  50	  &  	  &  	  &  18	  &  0	  &  \textbf{2964} & 99.4\%	  &  	  &  29	  &  0.001	  &  2956 & 99.1\%\\
\cline{2-2}\cline{5-8}\cline{10-13}
  &  25	  &  	  &  	  &  175	  &  0.004	  &  2916 & 97.8\%	  &  	  &  367	  &  0.008	  &  2849 & 95.5\%\\
\cline{2-2}\cline{5-8}\cline{10-13}
  &  20	  &  	  &  	  &  526	  &  0.012	  &  2854 & 95.7\%	  &  	  &  1137	  &  0.026	  &  2683 & 90.0\%\\
\cline{2-2}\cline{5-8}\cline{10-13}
  &  15	  &  	  &  	  &  117	  &  0.003	  &  2913 & 97.7\%	  &  	  &  1236	  &  0.028	  &  2557 & 85.7\%\\
\cline{2-2}\cline{5-8}\cline{10-13}
  &  10	  &  	  &  	  &  105	  &  0.002	  &  2909 & 97.5\%  &  	  &  4865	  &  0.109	  &  1939 & 65.0\%\\

\hline

\multicolumn{13}{c}{CC =Constrained Capacity; PC = Percent Capacity; OP**= Operating Point of 0.1 is where the threshold is selected for 0.1\% FAR with 100\% features}\\
\end{tabular}
\vspace{-4mm}
\end{table*}

\subsection{Capacity Assessment: Impact of Filter Resolution} 
This section explores how increasing features in terms of filter resolution impacts system capacity. Does fusing the information content in different resolutions of the iris increase the discriminability between identities? For this study, we considered templates generated at a single filter resolution versus templates generated at 3 different filter resolutions. We analyzed 12 different system configurations for each resolution type. For each of the 12 comparisons between the false reject rate of systems using multi-resolution template versus single-resolution template, as reported in Table~\ref{table:FAR_FRR}, single resolution outperforms multi-resolution (e.g. FRR: S24 - 1.76\% vs S22 - 9.56\%). However, our analysis of constrained capacity shows where multi-resolution may have an importance which is not reflected in considering only FRR and FAR, and is further discussed below.\par
Table~\ref{table:ALLQ_FA_NICF} and Table~\ref{table:ISOQ_FA_NICF} compare the impact of resolution on system capacity. As we compare the FA and CC between multi and single-resolution, we note variable impact. Systems working with a lower OP (0.1\%) perform considerably better with multi-resolution templates. Said more simply, for operating points with less strict FAR of 0.1\% (and better FRR), fewer subjects contribute to errors for multi-resolution templates compared to a single-resolution, i.e., a multi-resolution has a higher system capacity providing a more robust defence to false accepts from different identities. For example, a D1 multi-dimensional template generated from ALLQ data, with 100\% features functioning at an OP of 0.1\%, 1290 identities (CC:3868) contribute to the 0.1\% FAR, whereas 1608 identities (CC:3550) contribute to the false accepts with similar systems configuration with single resolution templates. However, for systems working at a stricter OP (0.001\% ), the performance variation between single and multi-resolution templates is not significant. For example, a D1 multi-dimensional template generated from ALLQ data, with 100\% features functioning at an OP of 0.001\%, 83 identities (CC:5075) contribute to the 0.001\% FAR whereas 89 identities (CC: 5069) contribute to the false accepts with similar systems configuration with single resolution templates. The variation gradually fades with stricter OP (0.01\% and 0.001\%). This pattern holds true across all parameters- different quality datasets (ALLQ, ISOQ), different template dimensions (D1, D2) and different feature dimensions (100\% to 10\%).

\subsection{Capacity Assessment: Impact of Spatial Resolution }\label{sec:ITD}
This section explores the impact of spatial resolution or template dimension on system capacity. Does increased spatial resolution lead to higher discriminable information content and improve system capacity? We assess two dimensions- $\sim $ 48k  (D1) and $\sim$ 26k bits (D2) at single resolution. For multi-resolution templates, the pixel count is three times that of the single-resolution templates. D1 and D2 templates also vary in their frequency information content as detailed in Section~\ref{sec:FDR}.

We observe in Table \ref{table:FAR_FRR} that a higher dimension/resolution (D1) leads to a comparatively higher FRR by a substantial margin in the range of 1\% to 8\%. For example, S5 and S7, both having similar system configurations render an FRR of 24.9\% with multi-resolution data compared to 16.59\% FRR with single-resolution data. Similar results are observed across all configurations. Looking at the performance in terms of percent capacity (PC) in Table~\ref{table:ALLQ_FA_NICF} and Table~\ref{table:ISOQ_FA_NICF}, lower template dimension (D2) performs better consistently across different feature dimensions (100\% to 10\%), more dominantly in ISOQ data. 

The observations could be indicative of the impact of spatial resolution or/and filter design. D1 templates are generated at the default dimension of OSIRIS and with the recommended filters. D2, which is a downsampled representation of D1, is comparatively more resistant to false rejects with the same filter design extracting different sets of feature components.

In Table~\ref{table:ISOQ_FA_NICF}, with ISO Quality data at 0.1\% OP, D2 renders a PC of 81.8\% (multi-resolution) and 75.2\% (single-resolution) compared to a PC of 77.9\% (multi-resolution) and 65.6\%  (single resolution) with D1. However, the variation in performance fades at a stricter operating point. One exception is noted, where higher dimensional templates (D1) when operating with 25\% of the features outperform lower dimensional templates (D2). This observation is true across all OP, quality and different resolution templates.  At 0.001\% FAR, a multi-resolution template with 100\% features has a system capacity of 5075 with D1 and 5074 with D2, with NICF of 83 and 84 respectively. In terms of constrained capacity (CC), we note a similar pattern. High dimensional templates (D1) leads to lower capacity in comparison to the lower dimensional template (D2) for similar system configuration. Even a stricter OP does not mitigate the variation in performance between D1 and D2. It is important to note, we assessed FAR with a precision of up to three decimal places. Thus, even though the OP has been fixed ( 0.1\%, 0.01\% and 0.001\% FAR), the CC was determined at the closest experimentally computed FAR to the selected OP. The actual FAR at which the FA and CC are computed are reported in Table~\ref{table:ALLQ_FA_NICF} and Table~\ref{table:ISOQ_FA_NICF} under the column ``FAR.'' 

We conclude from our assessment that a higher template dimension does not necessarily add to higher discriminable information. Rather, we observe performance degradation with higher template dimensions. From our observation, we hypothesize and attribute this result to the infusion of more redundant features with higher dimensions or more matching bits between non-mated pairs of images. A deeper analysis at the bit level needs to be performed to understand the root cause of this observation. We also note that the optimum filter design which defines the frequency information contained in the template remains unknown to the community.  

We can only conclude with the dimensions and filters we have tested in our experimentation. The optimum template dimension and optimum filter design for optimum system capacity remain open scopes of research. As biometric applications are moving towards a national stage or global stage with applications like huge-scale identification or de-duplication of identity at a national level or global level, system capacity may need to consider optimum information content in a template.

\subsection{Capacity Assessment: Impact of Random Feature Reduction vs Structured Feature Dimension }
Does different methods of selection of template dimension impact system capacity? Does reduced feature dimension impact performance? If yes, then to what extent? Is there a trade-off between constrained capacity and structured template generation at lower dimension vs random feature reduction?

To address these research questions, we assessed two methods of feature template generation - templates generated at different dimensions as described in Section~\ref{sec:FDR} and analyzed independently in Section~\ref{sec:ITD} versus bit level feature selection by random feature elimination as detailed in Section~\ref{sec:RFR}. We assessed two dimensions for structured template generation - 48128 bits (D1) and 26116 bits (D2). To assess random feature reduction, we assessed seven feature levels at 100\%, 75\%, 50\%, 25\%, 20\%, 15\% and  10\% of the original template dimensions (D1 and D2). 50\% of the D1 template, i.e. 24064 bits selected by bit-level random elimination is comparable to the D2 template (26116 bits) at 100\%. The two templates contain a similar number of bits; however, the fundamental difference between the two templates is the information content. The templates with feature dimensions D1 and D2 contain information from the entire usable iris area, whereas the bit-level selection of template features contains partial information based on a selective percentage (100\% to 10\%) of the entire usable iris. All assessments were performed for different quality datasets at different OP.

\textbf{Bit level random feature reduction:} Analyzing the performance in terms of percent capacity in Table~\ref{table:ALLQ_FA_NICF} and Table~\ref{table:ISOQ_FA_NICF}, we note an interesting pattern. Randomly eliminating radial features from the templates (D1 and D2) from 100\% to 50\%, gradually improves percent capacity (PC) by a substantial percentage. This observation is true across starting template dimension, filter resolution, quality and across OP. The positive impact is more dominating with system configurations operating with ISO-grade multi-resolution templates at lower OP (0.1\% / 0.01\%). For example, a system operating with ISO samples with D1 multi-resolution template at OP 0.1\% has a PC of 81.89\% with 100\% features and a PC of  84.74\% with 50\% features. We conclude that an iris image in its entirety is not an absolute requirement for the optimal performance of an iris recognition system. The improvement in performance after feature reduction by random elimination could be attributed to a reduction in redundant features. However, we also note that though 50\% features contribute to the highest resistance against identity clash, the variation in PC with 100\%, 75\% and 50\% features is in the range of 0.2\% to 5.5\%. This observation could be indicative of two conclusions: an improvement in performance with up to 50\% features or no substantial variation in performance with 100\%, 75\% and 50\% features. The improvement in performance after feature reduction by random elimination could be attributed to the reduction in redundant features. Alternatively, this could be indicative of no loss in discriminating features leading to stable performance even with the reduced features up to 50\%. The apparent improvement could be nullified if the random selection is iterated multiple times. 

We also note that multi-resolution templates are more robust in reducing feature dimensions compared to single-resolution templates. For most system configurations, reducing the feature dimension to 10\% of the original template dimension leads to a failure of the system ( as low as 3.9\% PC) if the templates are generated with single resolution; whereas with the multi-resolution template, though we note substantial degradation in performance, the worst performing system configuration obtained a system capacity of 61.09\% and the best performing configuration obtained 96.74\% with ALLQ data. Systems operating at stricter OP (0.001\%) have minimal impact on reducing features.

\textbf{Structured Template Generation vs Random Feature Elimination:}  we compared similar template dimensions generated from two different methods- (a) a structured generation of a template by varying the patch size during patch-wise translation of the raw iris image to its phasor representation; (b) Random elimination of radial features. The comparative performance of the two methods based on comparative number of bits in the template is presented in Table~\ref{table:FeatureSelection}. We compare template D2 with 100\% bits and 50\% bits with comparable number of bits in D1 at 50\% bits and 25\% bits respectively. We note that with a higher bit count ($\sim$24k and 26k) generated from 50\% of D1 and 100\% D2, there is little performance variation with variations fading with stricter OP. However, at 25\% D1 and 50\% D2, the performance variation is high, especially with single-resolution data. For example, a template with 13k bits generated from 50\% D2 performs better by a large percent capacity compared to 12k bits generated from 25\% D1. We surmise that this large degradation in performance is an impact of information loss as we eliminate larger usable iris area in 25\% of D1 compared to 50\% of D2 in spite of having comparable bit count in the template. This observation is true across OP, filter resolutions and quality; however, it is less dominant at stricter OP. 50\% D2 (13k bits) outperforms 50\% D1 (24k bits). However, multi-resolution templates hold comparable performance even with lower usable iris area (25\% D1). Multi-resolution templates are robust to performance degradation with less usable iris area; this observation reflects the contribution of unique discriminable information present in the different frequency bands of the iris. It also reflects the upper limit to the discriminable information content. With multi-resolution templates, percent capacity is not proportionally increased  with the more usable area (50\% D2) compared to templates with the less usable area (25\% D1).

\begin{table*}[t]
\scriptsize
\centering
\caption{ Comparative performance of percent capacity as a function of template structure }
\label{table:FeatureSelection}
\begin{tabular}{|c|c|c||c|c||c|c||c|c||c|c|}
\hline
 \textbf{OP} & \multicolumn{2}{c||}{\textbf{Feature (\%)}} & \multicolumn{8}{c|}{\textbf{Percent Capacity (\%)}}\\

\cline{4-11}
 
&  \multicolumn{2}{c||}{\textbf{(Bits)}} & 
 \multicolumn{4}{c||}{\textbf{ ALLQ }} & \multicolumn{4}{c|}{\textbf{ISOQ }}  \\
 
\cline{4-11}
&  \multicolumn{2}{c||}{\textbf{}}& 
 \multicolumn{2}{c||}{\textbf{ Multi Resolution }} & \multicolumn{2}{c||}{\textbf{Single Resolution }} & 
 \multicolumn{2}{c||}{\textbf{ Multi Resolution }} & \multicolumn{2}{c|}{\textbf{Single Resolution }}\\
\cline{2-11}

 & D1 & D2 & D1 & D2 & D1 & D2 &  D1 & D2  & D1 & D2 \\

\hline

 \multirow{2}{*}{0.1 }&	50\% (24064) &	100\%	(26116) &	80.48 &	76.35 & 71.13 &	73.32 &		83.5 &	81.89 &		70.95 &	75.21 \\
\cline{2-11}	
														
 &	25\%	(12032) &	50\% (13058) &	77.24 &	82.09 &	\textbf{57.23} &\textbf{76.08} &		81.89 &	84.74 &		\textbf{56.32} &	\textbf{78.4} \\
\hline
\hline

 \multirow{2}{*}{0.01} &	50\% (24064) &	100\% (26116) &	94.44 &	92.65 &	92.92 &	91.39 &		95.87 &	94.87 &		92.22 &	93.83\\
 \cline{2-11}
														
	& 25\%  (12032) &	50\% (13058) &	93.02 &	95.5 &	87.17 &	94.05 &	95.74 &	97.01 &		\textbf{85.94} &		\textbf{95.44}\\
\hline
\hline

 \multirow{2}{*}{0.001} &	50\% (24064) &		100\%	(26116) &	98.97 &	98.37 &	98.86 &	98.43 &		99.3 &	98.93 &		98.96 &	98.76\\
 \cline{2-11} 
														
& 	25\% (12032) &		50\%	(13058) &	98.27 &	99.09 &	97.54 &	99.03 &		99.36 &	99.43	 &	97.38 &	99.16\\
\hline

\end{tabular}
\vspace{-4mm}
\end{table*}

\section{Discussion, Limitation and Future Work}

Constrained capacity is a quantifiable measure of the ``uniqueness" of an iris recognition system and is a function of multiple system parameters. This study reports on the empirical assessment of constrained capacity for iris recognition systems operating at an acceptable error rate, i.e., the upper bound for the number of identities a system can resolve before encountering an identity clash. In our assessment, we studied iris templates captured under NIR illumination following Daugman-based IrisCode feature templates from 5158 identities comprising 13.2 million IrisCode comparisons for each of 12 different system configurations operating with all-quality data and from 2982 identities comprising 4.4 million IrisCode comparisons for each of 12 different system configurations operating with ISO-quality data. We studied 24 different system configurations varying six different IrisCode-based system parameters- sample quality (ALLQ vs ISOQ), filter resolution (multi-resolution template vs single resolution template), template dimension ($\sim 26k$ bits (D2) vs $\sim 48k$ bits (D1)), random feature elimination, system operating points (0.1\%, 0.01\% and 0.001\% FAR), and the number of identities in the system. 

\textit{Identity clash} is a concept related to the imposter matches impacting false match rate (FMR). However, biometric systems operate with an error trade-off between false match rate (FMR) and false non-match rate (FNMR) defined by a threshold. Though very very low FMR can be achieved in an iris recognition system by considering a highly stringent threshold, it would be impractical to select a threshold with extremely low FMR that results in an extremely high FNMR. NIST recommends iris biometric system operation at lower FMR with the caution that elevated FNMR should be an application-based decision \cite{grotherirex1}. In some applications high FNMR is untenable, a very high FNMR can be beneficial for a trade-off of near zero FMR \cite{grother2009performance}. We are of the view, that FNMR is an equally important factor as FMR irrespective of applications, as biometric applications are sensitive to security from a small scale to a matter of national security. Thus, it is seminal to advance research in the direction to improve performance specific to large-scale applications where the FMR-FNMR trade-off is minimal. In our study, we assess false matches at reasonable operating points being mindful of false rejections.

In our random feature selection from a single iteration, we recognize a possible limitation in generalizability. We observed that all 12 systems operating with single-resolution templates show patterns in the order of the first failure as a function of the ``feature level". However, this pattern is not reflected with multi-resolution templates. Additionally, 19 of the 24 systems studied render the best constrained capacity with 50\% feature level. Since the feature selection is random, this may reflect the effect of the particular set of selected features. However, though we have performed a single iteration of random feature selection, our process has induced randomness in 2 scenarios:
(a)	A single template is compared with 5157 other templates. In each case, a new random set of features are selected. This holds true for each of the 13.2 million / 4.4 million comparisons with each of the 24 systems studied. 
(b)	Additionally, though we are considering a single template for comparison, random selection/elimination of features replicates scenarios of 7 different templates, with reduced features, essentially replicating scenarios of different templates with obstructions, for each of the 24 systems studied. However, this does not address scenarios of templates with different dilation and angle of vision, which is a limitation of our study.
If resources in terms of time and processing power are not a constraint, multiple iterations of random selection would add more confidence to the conclusions. 

This study extends the state-of-the-art in the field of empirical assessment of the constrained capacity of iris recognition systems from two major contributors- Daugman\cite{daugman2006probing} and NIST\cite{grother2009performance} \cite{grother2012irex}. To the best of our knowledge, this is the \textbf{most extensive research on iris recognition system capacity in terms of six parameters covering 24 configurations and their correlation, using open-source algorithms with publicly accessible datasets.} Thus, the research reported here are \textbf{entirely reproducible and can be further extended,} using  the publicly available datasets listed in Table~\ref{table:Dataset} and the open-source software- OSIRIS. NIST reported on N:N matching on large proprietary databases for commercial software which are ``black boxes.'' Daugman's assessments are specific to the capacity of IrisCode. Conventionally, 2048 bits IrisCode are used for iris recognition. We have gone \textbf{beyond the conventions of iris recognition in the scope of IrisCode dimension and feature reduction in our study.} We explored capacity at higher dimensions ($\sim 26k - bits$ and $ \sim 48k - bits $) with different levels of feature content (100\% to 10\%) in the template with the motivation to study the relationship between template dimension, discriminable feature content and constrained capacity of the system. This study provides a framework for users to make a knowledge-based selection of parameters for their system configuration based on user requirements.

Limitations in our work open up areas for advanced research. Our study is limited by the publicly available dataset. The vision of solving challenges at the global level requires access to large-scale datasets representative of the variations in terms of identity count, variations in real-life data collections, and demographic variability, for the research community. Assessment of the impact of different system parameters on constrained capacity with a larger dataset would put to test the upper bounds of the system capacity.  Alternatively, there is a requirement to develop a methodology capable of predicting quantifiable system capacity with a smaller representative dataset. 

Additionally, our study is limited in its scope of iris system assessment. The scope of empirical assessment of iris recognition systems has multiple aspects - datasets, algorithm design, features and matcher. Each of these factors defines multiple system configurations. Each system design is unique in its parametric makeup. Parametric comparisons of all systems may not be possible. However, with our proposed framework, each system can be assessed independently. Our study is designed around assessing NIR-illuminated Daugman-style IrisCode-based system parameters. Arguably, IrisCode is the most popular iris feature used across different commercial systems. However, there are alternative NIR-illuminaiton-based non-Daugman iris recognition systems with open-source implementations. Given the extensive preparation (acquiring datasets, dataset cleaning, quality assessment, feature extraction, N:N matching), assessment time and analysis required for each system capacity evaluation, additional system assessment is not considered as part of this report. Alternative iris recognition resilience to system capacity remains an open research area.   

Our proposed framework for the assessment of iris recognition system is parametric and is therefore best suited for open-source algorithms which allow customization of  parameters. However, the framework can be generally applied to commercial software as well, depending on the limitations of the software.
In addition, our proposed framework has been developed specifically for NIR-illuminated Daugman-style IrisCode-based system. However, the framework can be adapted to non-Daugman systems as well.

VeriEye \cite{VerieyeSDK} is a commercial software; hence a black box. Neurotechnology, the company which developed VeriEye, uses neural networks for biometric recognition. Thus, it can be assumed that VeriEye is non-Daugman-based commercial iris recognition software. We use VeriEye as an example to illustrate the usage of our proposed framework for commercial software. The system capacity of VeriEye can be tested by varying three parameters: (a) the quality of image samples; (b) the number of identities in the system; (c) the operating system point. VeriEye provides feedback on the quality measure of iris images and has publicly documented the calibrated operating points of the system. Thus multiple configurations of the VeriEye algorithm with the three customizable parameters can be assessed. 

Our framework can be adapted by open-source non-Daugman systems. For example, the University of Salzburg Iris Toolkit (USIT) \cite{USIT3} for iris recognition is an open-source software that supports the extraction of iris features by using multiple solutions proposed in the literature. The options include: (a) 1D-LogGabor feature, (b) complex Gabor filter bank, (c) SIFT-based IrisCode, (d) SURF-based IrisCode, (e) local binary pattern (LBP) based features, (f) support for the algorithms proposed by Ma et al. \cite{ma2004efficient} and Ko et al. \cite{ko2007novel}. Each of these options has different parameters. For example, LBP features have several parameters including the number of neighbours, radius for selection of neighbours, indices-based neighbour selection, selection of thresholds, and other relevant parameters. The image quality, number of samples in the system, and operating point of the system remain common underlying parameters for all systems. Iris recognition system capacity for different system configurations of the LBP-based system can be assessed following our proposed framework.

The developed iris recognition technology is dependent on NIR-illumination of the iris. Though NIR-illuminated samples for iris recognition are standard and widely used for biometric applications, the scope of visible-range-illumination-based iris recognition is in-demand and has a large potential for applications upon technological maturity. Even though visible-range iris recognition is not used in the mainstream presently, multiple systems are proposed through research. Having the potential of being an applicable technology, assessment of the capacity of visible-illumination-based systems would add value to the existing and under-development technologies. While the basic framework for empirical assessment of iris recognition technology proposed in this work could be followed, the assessment of visible-illumination-based systems is an entire scope of research with different parametric designs of the systems - visible-illumination iris datasets, feature extractors and matching algorithms.

\section{Conclusion} \label{sec:DC}

 We conclude by summarizing our observations and answering the questions this work primarily focused on-

\textit{\textbf{ How does identity count in a system impact capacity?}}
Increased identity count in a system increases the probability of identity clash. However, the choice of system parameters highly influences the number of identities contributing to identity clashes. Systems working at stricter operating points (OP) achieve extremely high percent capacity, across resolution, data quality, template dimension and feature content, as the number of identities increases in the system. However, our conclusions are limited by the number of identities studied ( 5158 ). A large-scale database might provide a deeper perspective of the upper bound of the parameters.    

\textit{\textbf{ How does quality impact system capacity?}}
 Image quality is the most important criterion in system configuration determination for high system capacity. Quality directly impacts the choice of the OP for the system. ISOQ-based dataset performs with practically acceptable FRR at a stricter OP (0.001\% FAR) (refer Table~\ref{table:FAR_FRR}). A stricter OP leads to a substantial decline in identity clash in the database (refer Table~\ref{table:ISOQ_FA_NICF}), across different parameters (resolution, dimension, random feature reduction), which is desirable for all biometric security applications. 

\textit{\textbf{How does filter design impact system capacity? }}

Filter design determines the information contained in the template. Discriminable information extraction goes to the core of system capacity and thus filter design plays an important role. In our assessment, we tested one set of filters extracting different feature content for different spatial resolution templates, essentially testing different filter designs. We conclude that a lower dimensional template (D2) with information content extracted using the recommended filter for a higher dimensional template (D1) outperforms the OSIRIS recommended dimension-filter combination template in terms of false reject errors and system capacity. Thus, we conclude that the optimum filter design determination is a core parameter for optimum system capacity. 

\textit{\textbf{ How does filter resolution impact system capacity?}}
The choice of filter resolution is a trade-off between multiple parameters of the system. A single-resolution template would benefit processing time and power compared to a multi-resolution template. However, system capacity depends on the choice of OP, quality of the dataset, template dimension and feature level. A system working at a stricter OP (0.001\% FAR), performs similarly across template resolution and template dimension. However, a system working at less strict OP (0.1\%, 0.01\%) achieves higher system capacity using multi-resolution templates at better FRR (3.05\%, 4.48\%  vs 10.64\% at 0.001\% FAR). 

\textit{\textbf{Does higher template dimension increase discriminable information in terms of system capacity?}}
A higher dimensional (D1) multi-resolution template has comparable performance as that of a lower dimensional (D2) single-resolution template. For the two dimensions we have studied, angular precision (.7 degrees for D1 and 1.5 degrees for D2) while unwrapping the iris did not translate into i=higher system capacity. Across almost all 24 system configurations studied, systems achieve best-constrained capacity with 50\% of the random radial features. A higher number of bits in a template does not necessarily translate to higher information content in terms of template discriminability. However, our conclusions are restricted to the dimensions we have studied. 

\textit{\textbf{Does template generation methodology impact system capacity?}}
With single-resolution templates, bit-level selection of a higher feature level from a template of lower feature dimension (D2) achieves high capacity compared to a template of similar bit-size but generated from a template of higher feature dimension (D1) at a lower feature level (refer Table~\ref{table:FeatureSelection}). However, our analysis of multi-resolution templates remains inconclusive with no pattern in performance.

\section*{Acknowledgments}
This material is based upon work supported in part by the National Science Foundation under Grant No. \#1650503 and the Center for Identification Technology Research. Our work was largely supported by Chameleon, the hardware testbed  funded by a grant from National Science Foundation. 

\bibliographystyle{IEEEtran}
\bibliography{Journal}

\vspace{-1.7cm}

\begin{IEEEbiography}[{\includegraphics[width=1in,height=1.25in,clip,keepaspectratio]{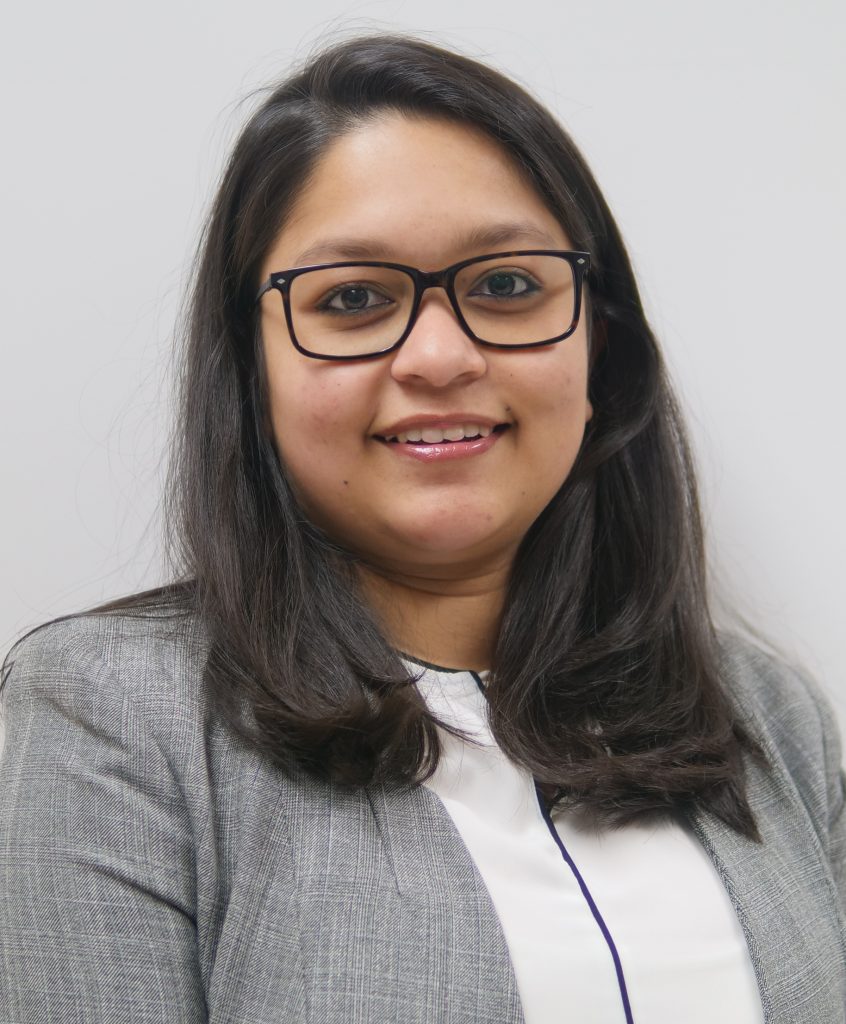}}]{Priyanka Das}

Dr. Priyanka Das is a Data Scientist with HID Global. Her research area of interest includes biometrics, computer vision, data science and human-computer interaction. She received her doctoral degree in the Department of Electrical and Computer Engineering at Clarkson University, USA with a focus on iris biometrics. She completed her ME in Bioscience and Engineering from Jadavpur University, India in 2016 and her Bachelor's in Biomedical Engineering in 2014.  
\end{IEEEbiography}

\vspace{-1.7cm}

\begin{IEEEbiography}[{\includegraphics[width=1in,height=1.25in,clip,keepaspectratio]{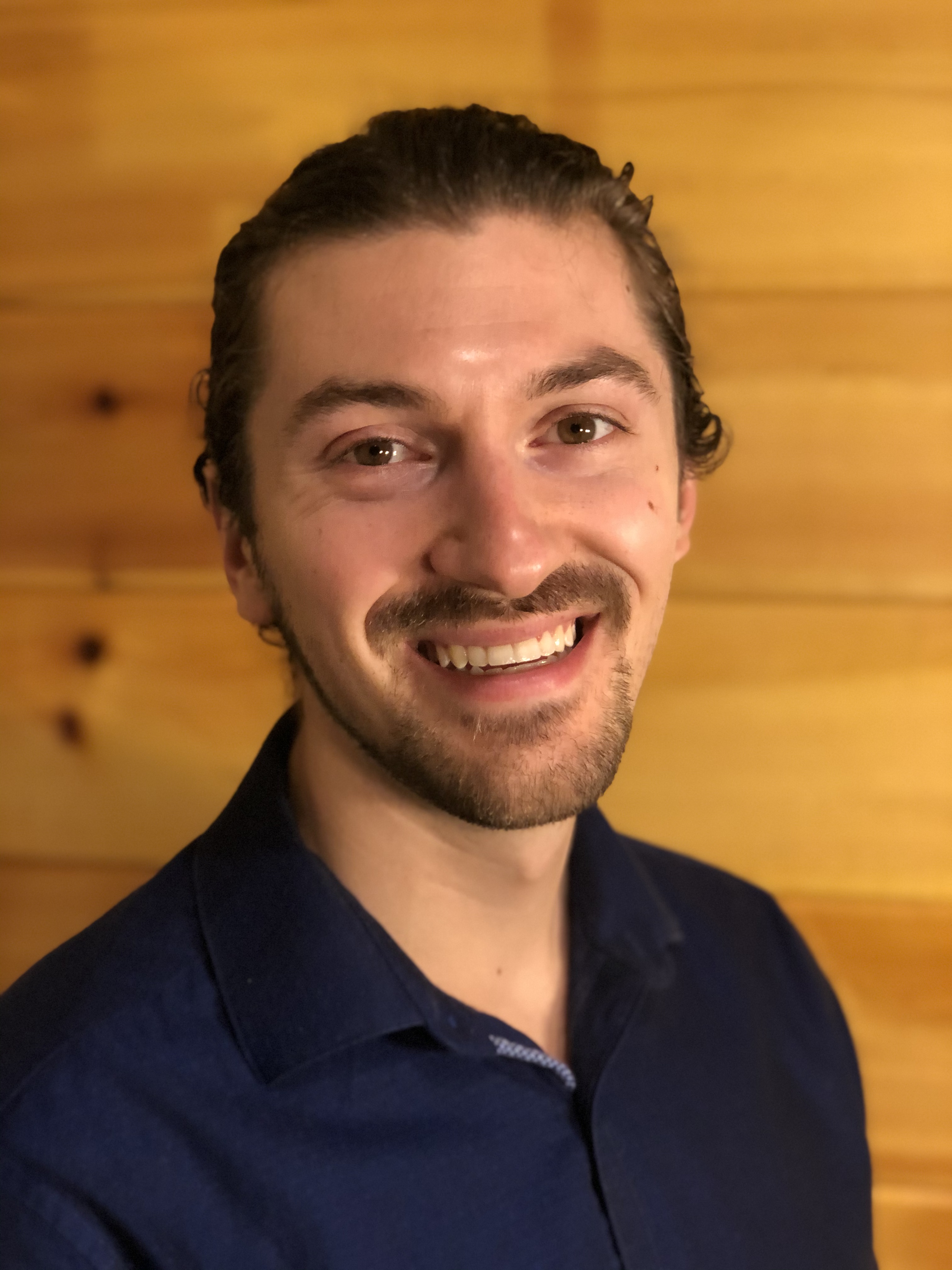}}]{Richard Plesh}

Richard Plesh received a Bachelor of Science in Electrical and Computer Engineering from Clarkson University in 2016 and a Master of Science in Data Analytics and an MBA from Clarkson University in 2019. He is currently pursuing his Ph.D. in the Department of Electrical and Computer Engineering at Clarkson University and is a 2021-2022 Fulbright Scholar at the University of Ljubljana Faculty of Electrical Engineering in Slovenia. His research centers around machine learning, data science, computer vision, and biometrics. 
\end{IEEEbiography}

\vspace{-1.7cm}

\begin{IEEEbiography}[{\includegraphics[width=1in,height=1.25in,clip,keepaspectratio]{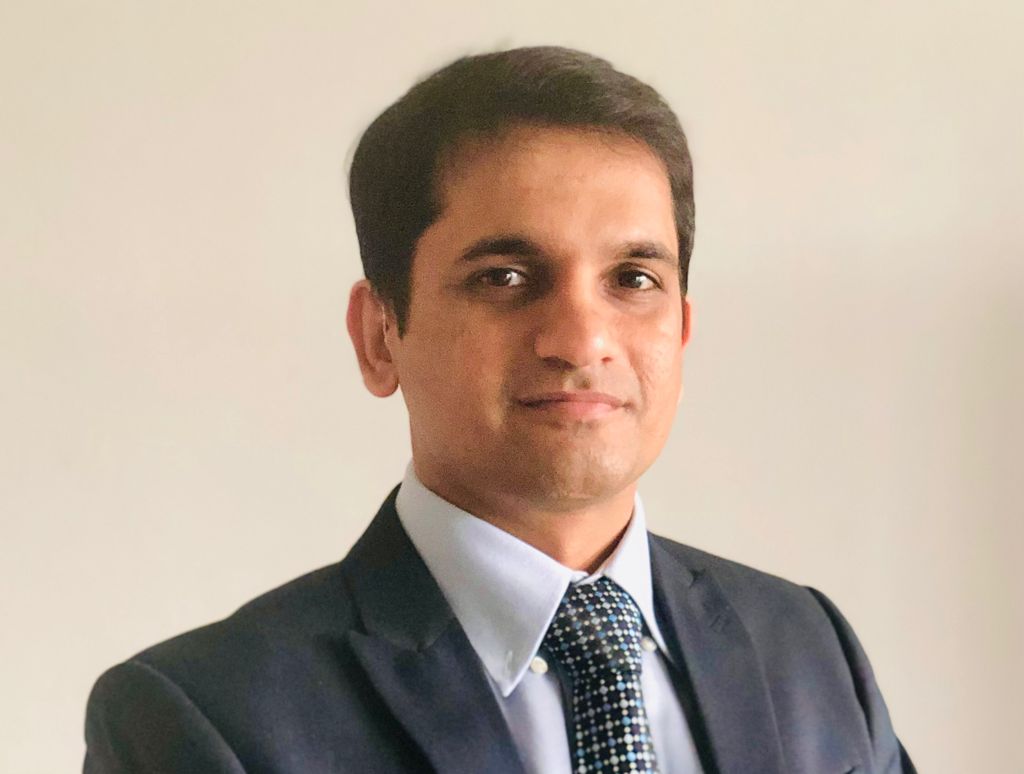}}]{Veeru Talreja}

Veeru Talreja received the M.S.E.E. and Ph.D. degrees from West Virginia
University, Morgantown, WV, USA and B.Engg. degree from Osmania University,
Hyderabad, India. From 2010 to 2013 he was a Geospatial Software Developer
with West Virginia University Research corporation. Since 2021, he has been a
Machine Learning Scientist of Product Development for VelocityEHS. His current
research interests include applied machine learning, deep learning, coding theory,
multimodal biometric recognition and security, and computer vision. 
\end{IEEEbiography}

\vspace{-1.7cm}

\begin{IEEEbiography}[{\includegraphics[width=1in,height=1.25in,clip,keepaspectratio]{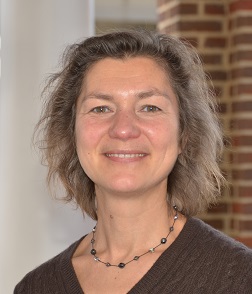}}]{Natalia A. Schmid} (M'96)

Natalia A. Schmid received her PhD in engineering from the Russian Academy of Sciences and her DSc in electrical engineering from Washington University in St. Louis, Missouri. She is currently a professor in the Department of Computer Science and Electrical Engineering at West Virginia University. Her research interests include detection and estimation, digital and statistical signal processing, information theory with applications to radio astronomy, biometrics, and digital forensics. She is a member of IEEE.
\end{IEEEbiography}
\vspace{-1.7cm}
\begin{IEEEbiography}[{\includegraphics[width=1in,height=1.25in,clip,keepaspectratio]{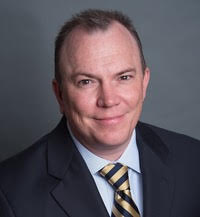}}]{Matthew Valenti} 

Matthew Valenti is a Professor in the Lane Department of Computer Science and Electrical Engineering at West Virginia University. Dr. Valenti's research and teaching interests are in the areas of communication theory, wireless networks, error control coding, cloud computing, and secure multimodal biometrics. He received B.S. and Ph.D. degrees from Virginia Tech and an M.S. from the Johns Hopkins University. He previously worked as an Electronics Engineer at the U.S. Naval Research Laboratory.  Dr. Valenti serves as Director of the West Virginia University site in the Center for Identification Technology Research (CITeR), an NSF industry/university cooperative research centre (I/UCRC). He is a recipient of the 2019 IEEE MILCOM Award for Sustained Technical Achievement and the 2021 IEEE Communications Society Communication Theory Committee Outstanding Service Award. Dr. Valenti is registered as a Professional Engineer (P.E.) in the state of West Virginia and is a Fellow of the IEEE.
\end{IEEEbiography}
\vspace{-1.7cm}
\begin{IEEEbiography}[{\includegraphics[width=1in,height=1.25in,clip,keepaspectratio]{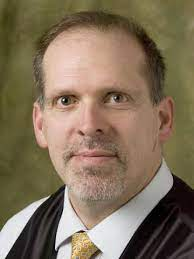}}]{Joseph Skufca}

Joseph Skufca is the Chair of Mathematics at Clarkson university.
He received a bachelor of science degree from the United States Naval Academy in 1985. He served 20 years in the Submarine Force, with sea tours aboard both fast attack and ballistic missile submarines of both the Atlantic and Pacific Fleets. He retired from active duty in 2005. He earned his master of science (2003) and Ph.D. (2005) degrees in applied mathematics from the University of Maryland, College Park. His research stretches across a broad spectrum of applied mathematics, with a particular focus on applied mathematical modelling, both with analytic and data methods.
\end{IEEEbiography}
\vspace{-1.8cm}

\begin{IEEEbiography}[{\includegraphics[width=1in,height=1.25in,clip,keepaspectratio]{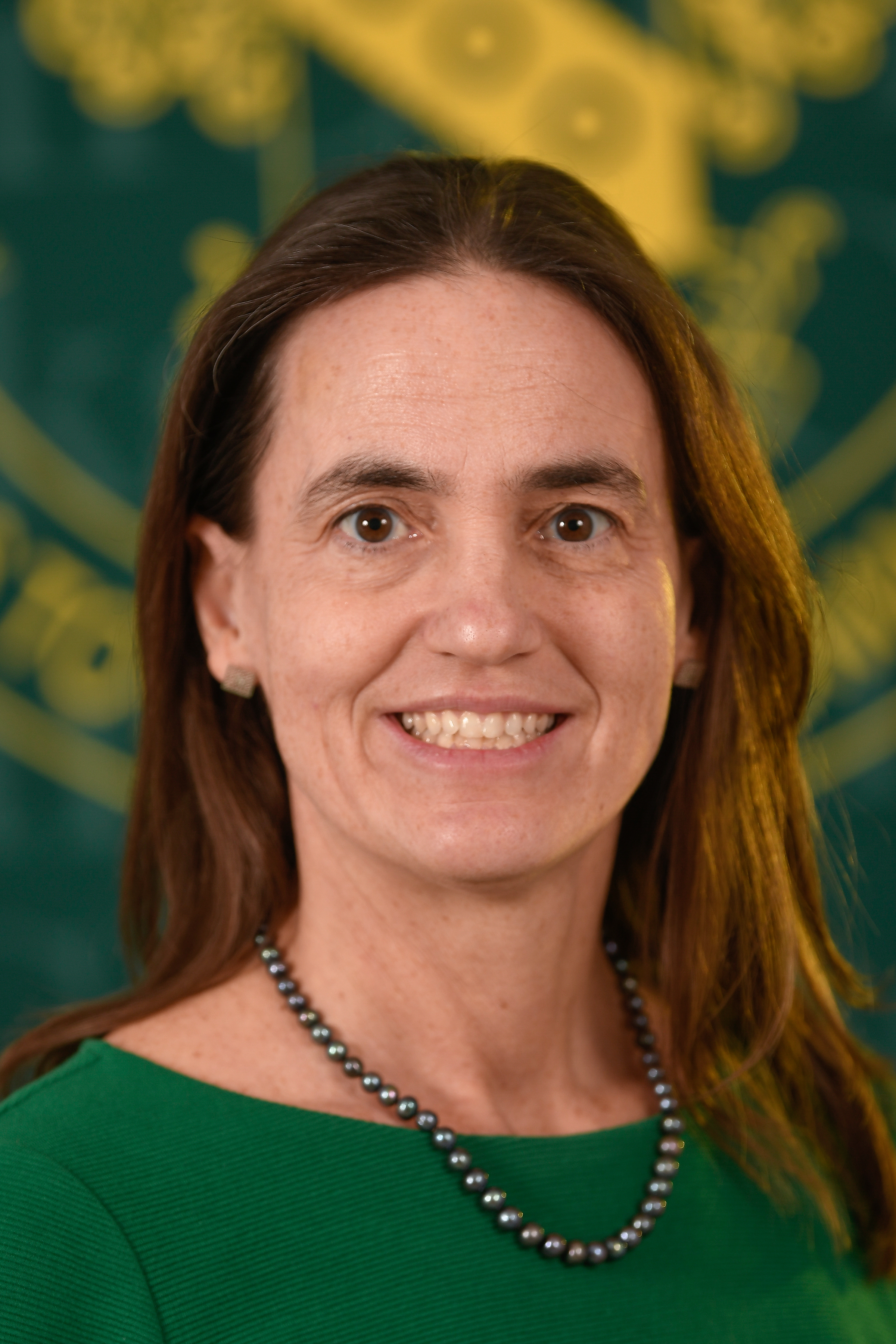}}]{Stephanie Schuckers}

Dr. Stephanie Schuckers is the Paynter-Krigman Endowed Professor in Engineering Science in the Department of Electrical and Computer Engineering at Clarkson University and serves as the Director of the Center for Identification Technology Research (CITeR), a National Science Foundation Industry/University Cooperative Research Center. She received her doctoral degree in Electrical Engineering from The University of Michigan. Professor Schuckers research focuses on processing and interpreting signals which arise from the human body. Her work is funded by various sources, including the National Science Foundation, the Department of Homeland Security, and private industry, among others.  She has started her own business, testified for the US Congress, and has over 100 other academic publications.
\end{IEEEbiography}

\end{document}